\documentclass[twoside,twocolumn,9pt]{article}
\usepackage{extsizes}
\usepackage[super,sort&compress,comma]{natbib} 
\usepackage[version=3]{mhchem}
\usepackage[left=1.5cm, right=1.5cm, top=1.785cm, bottom=2.0cm]{geometry}
\usepackage{balance}
\usepackage{mathptmx}
\usepackage{sectsty}
\usepackage{graphicx} 
\usepackage{lastpage}
\usepackage{comment}
\usepackage{soul}
\usepackage[format=plain,justification=justified,singlelinecheck=false,font={stretch=1.125,small,sf},labelfont=bf,labelsep=space]{caption}
\usepackage{float}
\usepackage{fancyhdr}
\usepackage{fnpos}
\usepackage{ulem}
\usepackage[english]{babel}
\addto{\captionsenglish}{%
  
}
\usepackage{array}
\usepackage{droidsans}
\usepackage{charter}
\usepackage[T1]{fontenc}
\usepackage[usenames,dvipsnames]{xcolor}
\usepackage{setspace}
\usepackage[compact]{titlesec}
\usepackage{hyperref}

\usepackage{epstopdf}

\definecolor{cream}{RGB}{222,217,201}

\usepackage{amssymb}
\usepackage{multirow}

\usepackage[switch]{lineno}

\begin{document}

\pagestyle{fancy}
\thispagestyle{plain}
\fancypagestyle{plain}{
\renewcommand{\headrulewidth}{0pt}
}

\newcommand*{\rood}[1]{\textcolor{red}{#1}}
\newcommand{\argmax}{\text{argmax}}
\newcommand{\argmin}{\text{argmin}}

\makeFNbottom
\makeatletter
\renewcommand\LARGE{\@setfontsize\LARGE{15pt}{17}}
\renewcommand\Large{\@setfontsize\Large{12pt}{14}}
\renewcommand\large{\@setfontsize\large{10pt}{12}}
\renewcommand\footnotesize{\@setfontsize\footnotesize{7pt}{10}}
\makeatother

\renewcommand{\thefootnote}{\fnsymbol{footnote}}
\renewcommand\footnoterule{\vspace*{1pt}%
\color{cream}\hrule width 3.5in height 0.4pt \color{black}\vspace*{5pt}} 
\setcounter{secnumdepth}{5}

\makeatletter 
\renewcommand\@biblabel[1]{#1}            
\renewcommand\@makefntext[1]%
{\noindent\makebox[0pt][r]{\@thefnmark\,}#1}
\makeatother 
\renewcommand{\figurename}{\small{Fig.}~}
\sectionfont{\sffamily\Large}
\subsectionfont{\normalsize}
\subsubsectionfont{\bf}
\setstretch{1.125} 
\setlength{\skip\footins}{0.8cm}
\setlength{\footnotesep}{0.25cm}
\setlength{\jot}{10pt}
\titlespacing*{\section}{0pt}{4pt}{4pt}
\titlespacing*{\subsection}{0pt}{15pt}{1pt}

\fancyfoot{}
\fancyfoot[LO,RE]{\vspace{-7.1pt}\includegraphics[height=9pt]{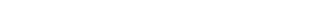}}
\fancyfoot[CO]{\vspace{-7.1pt}\hspace{13.2cm}\includegraphics{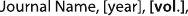}}
\fancyfoot[CE]{\vspace{-7.2pt}\hspace{-14.2cm}\includegraphics{RF}}
\fancyfoot[RO]{\footnotesize{\sffamily{1--\pageref{LastPage} ~\textbar  \hspace{2pt}\thepage}}}
\fancyfoot[LE]{\footnotesize{\sffamily{\thepage~\textbar\hspace{3.45cm} 1--\pageref{LastPage}}}}
\fancyhead{}
\renewcommand{\headrulewidth}{0pt} 
\renewcommand{\footrulewidth}{0pt}
\setlength{\arrayrulewidth}{1pt}
\setlength{\columnsep}{6.5mm}
\setlength\bibsep{1pt}

\makeatletter 
\newlength{\figrulesep} 
\setlength{\figrulesep}{0.5\textfloatsep} 

\newcommand{\topfigrule}{\vspace*{-1pt}%
\noindent{\color{cream}\rule[-\figrulesep]{\columnwidth}{1.5pt}} }

\newcommand{\botfigrule}{\vspace*{-2pt}%
\noindent{\color{cream}\rule[\figrulesep]{\columnwidth}{1.5pt}} }

\newcommand{\dblfigrule}{\vspace*{-1pt}%
\noindent{\color{cream}\rule[-\figrulesep]{\textwidth}{1.5pt}} }

\makeatother

\twocolumn[
  \begin{@twocolumnfalse}
{\includegraphics[height=30pt]{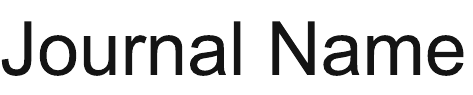}\hfill\raisebox{0pt}[0pt][0pt]{\includegraphics[height=55pt]{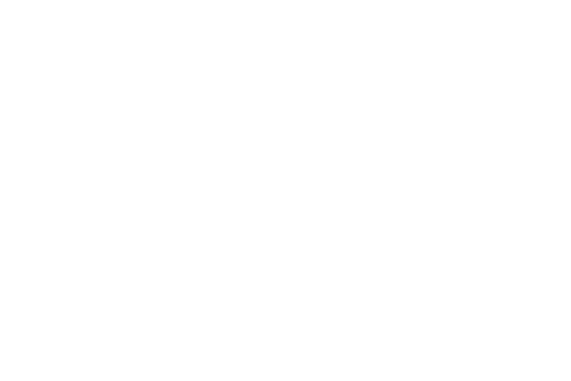}}\\[1ex]
\includegraphics[width=18.5cm]{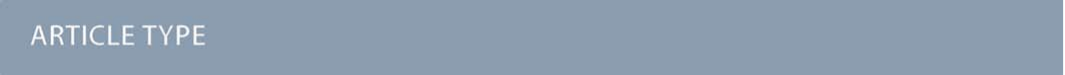}}\par
\vspace{1em}
\sffamily
\begin{tabular}{m{4.5cm} p{13.5cm} }

\includegraphics{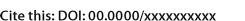} & \noindent\LARGE{\textbf{{Towards best practices in }low-dimensional semi-supervised latent Bayesian optimization for {the} design {of} antimicrobial peptides$^\dag$}} \\
\vspace{0.3cm} & \vspace{0.3cm} \\

 & \noindent\large{Jyler Menard,\textit{$^{a}$} and RA Mansbach$^{\ast}$\textit{$^{a}$}} \\

\includegraphics{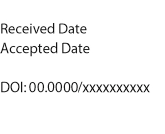} & \noindent\normalsize{{Generative deep learning techniques have demonstrated an impressive capacity for tackling biomolecular design problems in recent years. Despite their high performance, however, they still suffer from a lack of interpretability and rigorous quantification of associated search spaces, which are necessary to unlock their full potential for scientific inquiry beyond efficient design. An area in which they are of particular interest is in the design of antimicrobial peptides,} which are a promising class of therapeutics to treat bacterial infections. Discovering and designing such peptides is difficult because of the vast number of possible sequences {and comparatively small amount of experimental information. In this work, we perform a theoretical investigation of latent Bayesian optimization for searching through peptide sequence spaces, with a focus on antimicrobial peptides.} We investigate (1) whether searching through a dimensionally-reduced variant of the latent design space may facilitate optimization, (2) how organizing latent spaces {by differing amounts of more and less relevant information may improve the efficiency of arriving at an optimal peptide design,} and (3) the interpretability of the spaces. We find that employing a dimensionally-reduced version of the latent space is more interpretable and can be advantageous, {while the use of less-relevant but more easily-computable physicochemical properties is advantageous to latent space organization in certain contexts and the use of more-relevant but sparser properties associated with the latent Bayesian objective function is advantageous in others.} This work lays crucial groundwork for biophysically-motivated peptide design procedures, {with an especial focus on antimicrobial peptides.}} \\

%
%

\end{tabular}

 \end{@twocolumnfalse} \vspace{0.6cm}

  ]

\renewcommand*\rmdefault{bch}\normalfont\upshape
\rmfamily
\section*{}
\vspace{-1cm}


\footnotetext{\textit{$^{a}$~Department of Physics, Concordia University, Montr\'{e}al, QC, H4B 1R6, Canada. \hspace{5mm}E-mail: re.mansbach@concordia.ca}}

\footnotetext{\dag~Electronic Supplementary Information (ESI) available: {Further detailed analyses, tables and figures available in PDF form (4 sections, 10 tables, and 24 figures). Code for model training and analysis can be found at} \url{https://github.com/Mansbach-Lab/compare-latent-spaces-amps/tree/main}. {Trained model checkpoints and datasets are available at} \url{https://doi.org/10.5281/zenodo.17872449} See DOI: 00.0000/00000000.}



\section{Introduction}
Antimicrobial resistance is a growing global health concern. Through increasing usage in animal agriculture \cite{mulchandani_global_2023_animal_agriculture}, aquaculture \cite{schar_global_2020_aquaculture}, and in humans \cite{klein_global_2018_human_antibiotic_use}, it is expected that by 2050, $>8.2$ million deaths per year will be associated with AMR \cite{naghavi_global_2024_8.2}. With traditional pharmaceutical companies becoming less productive \cite{dimasi_innovation_2016}, fewer antibiotics being developed \cite{boucher_bad_2009, ventola_antibiotic_2015}, and of {the thirteen that have been approved since 2017, ten belong to antibiotic groups with known resistance mechanisms} \cite{who_antibiotics_2024}, there is an urgent need for novel classes of therapeutics to treat bacterial infections, and methods to quickly find them. 

One potentially fruitful area of research is the design and development of new and more effective antimicrobial peptides (AMPs). Antimicrobial peptides are short proteins of around 5 to 100 residues with broad-spectrum antibacterial activity \cite{seyfi2020antimicrobial}. Many AMPs are thought to act by disrupting the bacterial membrane,  through the formation of large pores, through inducing micellization of membrane patches (the so-called ``carpet method''), or through a combination of such mechanisms \cite{huan_antimicrobial_2020}. Because of the comparatively indiscriminate nature of the entire membrane as a target rather than a small protein binding pocket like most traditional small-molecule drugs, it was hypothesized that resistance to membrane-active AMPs will develop more slowly than to many traditional small-molecule antibiotics. Evidence supporting this hypothesis includes \textit{in vitro} experiments across four different gram-negative bacterial species showing less resistance developing to membrane-active antibiotics than to traditional antibiotics within the same time frame \cite{daruka_eskape_2025}. {In the same study, it was} demonstrated that strong resistance to traditional antibiotics appeared within 120 generations of repeated exposure \cite{daruka_eskape_2025}, compared to the 600 generations required for resistance to develop to an analogue of the peptide Magainin-2 that was observed by \citet{perron_experimental_2005}. {Despite differences in study methodology, the evidence overall supports slower resistance development to peptides.} In addition to the possibility of slowing resistance, certain AMPs can be used as part of combination drug cocktails to improve the efficacy of other antibiotics \cite{lewies_antimicrobial_2019_reviews_AMPantibioticSynergism}. Such combination therapies have also been found to have slower resistance development\cite{pizzolato-cezar_combinatory_2019}.  Due to their broad-spectrum activity, slow resistance, synergism with other antibiotics, and the need for new antibiotics, AMPs are a promising therapeutic class. 

While peptides are short compared to other proteins, the space of all possible peptide sequences is still vast. Considering only naturally-occurring amino acids, there are on the order of $\sum_{L=5}^{100} 20^L$ possible sequences. Traditional mutagenesis methods are inadequate to explore more than a tiny fraction of this sequence space to find peptides with properties of interest. { Traditional computational methods, including machine learning, can improve the speed of candidate identification by rapidly screening databases full of peptide and protein sequences. For example, some of the earliest predictive models for AMP identification were hidden markov models and shallow neural networks} \cite{fjell2007amper,fjell2009identification}{, with logistic regression} \cite{randou2013systematic} {and SVMs} \cite{lee2017can} {following after. More recently, motivated by the success of deep neural networks for (natural language) sequence representations, deep neural network architectures began to be employed} \cite{spanig_encodings_2019}. {All of these methods, however, function to identify candidates in a database and cannot suggest completely novel AMPs.} Because of the number of possible peptides compared to the speed at which they can be tested, finding peptides with desired properties is difficult. 

Deep learning generative models can partially obviate the need to pre-generate libraries of sequences by providing a method to sample arbitrarily many new sequences. Recently, such models have been adopted by the biomolecular design community and have demonstrated a powerful capacity to produce peptides with desirable properties. \citet{capecchi_machine_2021} {trained a strain-specific RNN to generate peptide sequences and filtered them using classifiers.} \citet{li2024foundation} {trained a generative model based on a transformer architecture intended to generate high-activity mutants from an input sequence.} \citet{zakharova_machine_2022} {used an RNN to generate anticancer peptides.} \citet{szymczak_discovering_2023} {designed a conditional VAE--conditioned on categorical classifications of AMPness and high/low activity levels--in an attempt to directly sample from the high-activity AMP conditional probability distribution.} \citet{das_accelerated_2021} {used a Wasserstein autoencoder together with a number of classifiers to generate peptides}. 

{While impressive work is being done in improving both predictive and generative models, tying them together in a closed-loop optimization scheme is (currently) less common. Furthermore,} the focus of these models has tended to be on design of peptides for specific applications rather than an understanding of the ways in which the specific model and procedure affect the underlying design space. Recently, we developed quantitative benchmarks and investigated the properties of the design space itself in the context of our own generative model for antimicrobial peptide sequence design based on a variational autoencoder (VAE) \cite{renaud_latent_2023}. VAE-based models possess associated latent spaces that can be thought of as explicit continuous design space representations of the discrete peptide sequences. This formulation of the problem facilitates a focus on the quality of the design space itself. In this article, we expand on this work to investigate how the organization of the design space can affect optimization procedures in that space.

While generative deep learning models on their own do not resolve the fundamental needle-in-haystack issue of \emph{finding} sequences with significant antimicrobial potency and/or other properties of interest, they can produce candidates or design spaces to be used in conjunction with optimization techniques. Broadly speaking, finding a sequence with high antimicrobial activity can be modelled as an optimization problem, with a corresponding objective function that maps from peptide sequences to antimicrobial activity. In a scenario where there is a straightforward function mapping sequences to antimicrobial activity, a litany of optimization algorithms would allow us to find the sequence (or sequences) corresponding to maximal antimicrobial activity. However, how antimicrobial activity relates to the peptide sequence is not sufficiently understood to write down a straightforward function, instead leaving us in the situation of having to treat the relationship as a black box we can probe through expensive experimental assays, machine learning models limited by their data sets, and/or long, resource-intensive MD simulations depending on our available resources. 

Bayesian optimization ( BayesOpt ) is an iterative method of finding the optimum of a black-box function. BayesOpt fits a data-driven, simple, and flexible \textit{surrogate model} to approximate the relationship between input points (such as peptide sequences) and the objective function (such as antimicrobial potency). To perform BayesOpt one must also define an \textit{acquisition function} that scores to what extent a point in the design space is predicted to improve on the current best observed value. At each iteration, the acquisition function combined with the surrogate model identifies a point of interest for the next round predicted to most improve upon the optimal point found so far. 

One can employ BayesOpt to search through the latent space of an associated generative model; this is referred to as Latent Bayesian Optimization (LBO) \cite{gomez-bombarelli_automatic_2018, grosnit_high-dimensional_2021, lee_advancing_2023, tripp_sample-efficient_2020}. LBO has been previously successfully applied to the design of small molecules \cite{gomez-bombarelli_automatic_2018} and synthetic optoelectronic peptides \cite{shmilovich_discovery_2020}. Although screening peptides sampled from a generative model to design AMPs \textit{without} using Bayesian optimization also has precedent \cite{szymczak_discovering_2023, das_accelerated_2021}, we focus on LBO because {of its data efficiency and potential for application with higher fidelity, more costly oracles.} 

Other works have explored how the organization of the latent space can impact the efficiency of LBO. A number of ways to organize the latent space have been investigated, with particular focus on small molecule discovery where large amounts of data are readily available. In \citet{gomez-bombarelli_automatic_2018}, the authors investigated whether jointly training a property predictor and a decoder can improve the efficiency of LBO for the task of finding optimal small molecules, finding compelling evidence that it can.  \citet{lee_advancing_2023} examined how LBO performs in a latent space induced to be correlated with the objective they are trying to optimize, finding promising results. In \citet{grosnit_high-dimensional_2021} the authors compare a number of contrastive learning methods, finding that a triplet loss can be effective. Given these works, it seems the efficiency of LBO can be improved by organizing the underlying latent space being searched through, but this has not been previously investigated in the context of antimicrobial peptide design. {One further consideration for LBO is that often the associated neural network has a relatively high-dimensional latent space. BayesOpt can have difficulty in high-dimensional spaces. On the other hand, reducing the dimensionality of the latent space may corrupt the ability of the associated generative model to generate diverse, novel, and realistic sequences. This leaves us with a dilemma where we may expect either worsened BayesOpt performance due to a high-dimensional latent space, or worsened generative model performance due to too small of a latent space.}

Unlike in the small molecule design space, where large amounts of data relevant to the performance of such molecules as antibiotics are readily available, the corresponding data on antimicrobial peptides is relatively sparse. There are hundreds of thousands or even millions of examples of small-molecule data as experimental drug-target binding affinities and well-established computed properties related to drug-likeness and solubility readily available. Conversely, in the context of peptides, antimicrobial activity in the form of experimentally-measured minimum inhibitory concentration (MIC) values is scarce; specifying particular bacterial species as targets restricts the available data even further. For example, for \textit{E. coli} there are just 11k entries with corresponding MIC measurements, and if we restrict ourselves to those sequences with more than one reported MIC measurement, then we are left with 4k entries \cite{pirtskhalava_dbaasp_2021}. Experimentally-measured peptide hemolysis, which is important for the assessment of host toxicity, is available for even fewer examples, with DBAASP containing 8k sequences with reported hemolysis measurements that falls to 1.6k when restricting ourselves to more than one measurement per sequence \cite{pirtskhalava_dbaasp_2021}. While this may be a sufficient amount for traditional machine learning methods, it is unclear how this amount may impact the effectiveness of generative deep learning models and the organization of their latent space. Moreover, it is unclear whether using easily-computed physicochemical properties as proxies may offer a benefit over the more direct but limited amount of activity data. 

{In this work, we determine best practices for performing latent Bayesian optimization in the context of peptide sequence spaces, with a focus on the antimicrobial peptide context in particular.} We introduce and assess a modification of {latent Bayesian Optimization} as it has been previously applied in drug discovery applications by performing Bayesian Optimization over a reduced space representation of {a peptide sequence-based} {the} latent space {for AMP design.} Rather than performing Bayesian Optimization over the high-dimensional latent space representation, we explore the possibility of performing Bayesian Optimization over a much lower-dimensional projection of that space {constructed with Principal Components Analysis.} The purpose of this approach is twofold: (i) we expect an increase in efficiency by using BayesOpt in a lower-dimensional space {without modifying the size of the generative model's latent space}, and (ii) we expect an increase in interpretability by performing optimization in a more similar space to that which we employ for visualization compared to using reduced projections solely to visualize (a proxy of) the design space. {We also note that we employ a proxy objective function (an “oracle”) related to predicted AMP activity for better control of algorithmic design prior to deployment on more realistic, expensive objectives, such as experimental measurements of peptide activity.}

We further investigate three important questions related to {incorporation of prior knowledge to latent Bayesian optimization in peptide sequence-based design spaces.} We first ask whether organizing the latent space with easily-computable physicochemical properties that are not directly related to {predicted} AMP activity is more informative or higher-performing than organizing the latent space with {a} sparse amount of {predicted} AMP activity data. We secondly ask whether organizing by more properties leads to a more structured, easier to optimize in, latent space. We thirdly ask whether increasing the percentage of labelled data available has a direct relationship to LBO efficiency and latent space organization, or if there are diminishing returns. By providing an analysis of LBO in both a semi-supervised and latent space organization context, we hope to further the discussions of how best to perform Bayesian optimization in latent spaces in a manner that informs us about the design space that is being traversed by the optimization algorithm, and introduce the technique to the therapeutic peptide design problem {in the context of AMPs}.

\section{Methods}

In this section, we describe the design of our AMP optimization process. We begin with an overview of our Bayesian optimization procedure and parameters, then define the associated objectives and the manner in which they are assessed. Then we describe the generative models for which the associated latent spaces become the design spaces of possible peptide sequences and present the methods we chose to organize the latent spaces. Lastly, we describe how we implemented Bayesian optimization in the latent space of this generative model.

\begin{figure*}
    \centering
    \includegraphics[width=\linewidth]{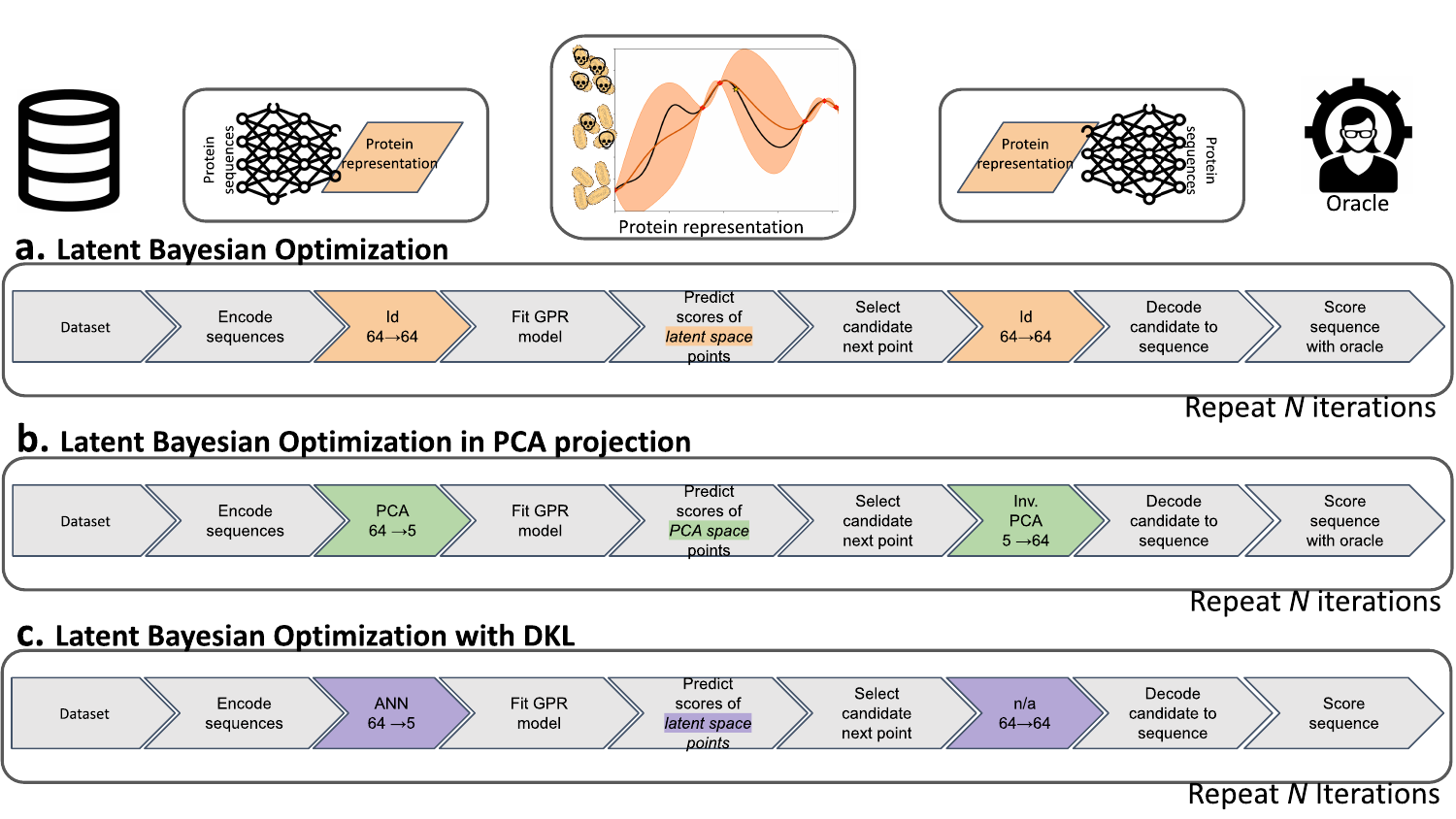}
    \caption{{Schematic overview of three approaches to Latent Bayesian Optimization investigated herein (differences highlighted). LBO begins assuming a pre-trained generative model with an encoder and a decoder module. Each method follows a similar procedure: encode the optimization dataset into the latent space of a transVAE model; fit a Gaussian Process Regressor; maximize the Expected Improvement to select the next candidate to score using the oracle; decode the candidate latent point to a amino acid sequence; score the sequence using the oracle; add the now-labelled sequence to the optimization dataset, and repeat $N$ times. The differences lie in whether (a) we use the full latent space representation directly in fitting the GPR, (b) we use a PCA projection of the latent space when fitting the GPR, or (c) whether a small neural network that projects the latent points into a smaller-dimensional space is fit simultaneously with the GPR. Following (b) for one BayesOpt iteration: encode optimization dataset into the 64-dimensional latent space of a transVAE; project those points into a five-dimensional space constructed using PCA; fit a GPR and select a candidate five-dimensional point; inverse project back to the 64-dimensional latent space; decode the candidate latent point to a candidate sequence; score the sequence using the oracle.}}
    \label{fig:schematic-overview}
\end{figure*}

\subsection{Bayesian Optimization}
To find highly active antimicrobial peptides, we use Bayesian Optimization ( BayesOpt ) with a single objective function, as implemented in the BoTorch Python package using the SingleTaskGP class \cite{balandat2020botorch}. BayesOpt is an iterative method of efficiently identifying design optima by using a surrogate (usually a quicker, less accurate assessment) of a desired objective function in combination with an acquisition function that selects the next potential design to be evaluated according to the true objective function (usually slower and more accurate assessment)\cite{shahriari_taking_2016}. A good acquisition function balances exploration of the design space with exploitation of observed data points\cite{shahriari_taking_2016}; in other words, it balances assessing new regions in design space with searching near known high-performing designs. 

We use a Gaussian Process Regression (GPR) model for the surrogate objective function. A GPR is a probabilistic model where, for any point $\vec{x}$ in our design space, we assume we have a Gaussian distribution (our prior distribution), fully determined by a mean $m$ and covariance matrix $\sigma = (\sigma_{ij})$. The mean and covariance are parameterized by a mean function $m(\vec{x})$ and kernel function $k(\vec{x}_i,\;\vec{x}_j)$ each of which depend on data points assessed according to the true objective function. For a data set $\mathcal{D}$ of $n$ points in an $F$-dimensional space with associated objective values $\mathcal{D} = \{(\vec{x}_j, y_j)\}_{j=1}^n$, where $\vec{x} \in \mathbb{R}^F$, at iteration $l$, we fit the mean and covariance 
\begin{align}
    m_l(\vec{x}) = m_l &= \frac{1}{|\mathcal{D}|}\sum_{j=1}^{|\mathcal{D}|} y_j \label{eqn:gpr_mean_prior} \\
    k_l(\vec{x}_i,\;\vec{x}_j) &= \exp\left(-\frac{1}{2}(\vec{x}_i - \vec{x}_j)^T\Theta_l^{-2} (\vec{x}_i - \vec{x}_j)\right) 
    \label{eqn:gpr_covar_prior}
\end{align}
where $\Theta_l \in \mathbb{R}^{F\times F}$ is a diagonal matrix of fitted length-scales, which are fitted using the exact marginal log-likelihood. Here we use the BoTorch default choice of a constant mean function--specifically the sample mean of the observed objective function values--so it does not depend on the input vector. Together these give a \textit{prior} distribution
\begin{equation}
\label{eqn:gpr_prior}
\mathcal{G}(m_l,\;\sigma_l).
\end{equation}
with which we compute a point-dependent \textit{posterior} distribution  using Bayes' rule. Using this set of \textit{posterior} distributions, we make predictions on unobserved points that have not been assessed according to the true objective function, and to compute an uncertainty in those predictions. The posterior distribution's mean $M_l$ and variance $\Sigma_l$ are determined to be:
\begin{align}
M_l(\vec{x}) &= m_l(\vec{x}) + \sigma_l(\vec{x},x_D)^T\sigma_l(x_D,x_D)^{-1}\cdot ({f}(x_D) - {m_l}(x_D) )  \label{eqn:gpr_mean_posterior} \\
\Sigma_l(\vec{x}) &= \sigma_l(\vec{x},\vec{x}) - \sigma_l(\vec{x}, x_D)^T\sigma_l(x_D, x_D)^{-1} \sigma_l(\vec{x},x_D) \label{eqn:gpr_covar_posterior},
\end{align}
where ${f}$ is the true objective function, $x_D$ is short-hand for the dataset of $n$ previously-assessed input points $\{\vec{x}_i\}$, $\sigma_l(x_D, x_D)$ is an $n\times n$ matrix with entries $\big[\sigma_l(x_D, x_D)\big]_{i,j} = k(\vec{x}_i, \vec{x}_j)$, and $\sigma_l(\vec{x},x_D)$ is a vector of length $n$ with entries $\big[\sigma_l(\vec{x},x_D)\big]_i = k(\vec{x},\vec{x}_i)$. We use the radial basis function kernel (default when using BoTorch's SingleTaskGP \cite{balandat2020botorch}), but other choices exist. Because the Radial Basis Function Kernel is distance-dependent, there is an implicit assumption that points near each other have similar objective values. Through the distance-dependence, points near each other in the design space will be predicted to have similar surrogate objective values. With the posterior distribution in hand, predictions are made either by sampling from the posterior distribution 
\begin{equation}
    \label{eqn:sample_posterior_distribution}
    \hat{f}(\vec{x}) \sim \mathcal{G}\big(M_l(\vec{x}),\; \Sigma_l(\vec{x})\big)
\end{equation}
or, more simply, by using the posterior mean (Eqn. \ref{eqn:gpr_mean_posterior}) as a point estimate, and the posterior covariance to compute the uncertainty range (Eqn. \ref{eqn:gpr_covar_posterior})

As mentioned, in addition to the Gaussian Process surrogate model, we must select an acquisition function, which identifies the next point to be evaluated with the true objective. A common choice is the Expected Improvement, defined as,

\begin{equation*}
    \label{eqn:expected_improvement}
    \text{EI}(\vec{x}) \equiv \mathbb{E}_{\hat{f}(\vec{x})}\left[\left(\hat{f}(\vec{x}) - f_l^*\right)^+\right],
\end{equation*}

\noindent where $f_l^*$ is the best value we have observed up to iteration $l$, $\hat{f}(x)$ is the predicted value of the surrogate function at point $\vec{x}$, and $\left(\hat{f}(\vec{x}) - f_l^*\right)^+ = \max\left(0,\hat{f}(\vec{x}) - f_l^*\right)$ is the improvement. $\text{EI}(\vec{x})$ encourages exploitation by searching for a point that is likely to improve on the current best point, and it encourages exploration through the standard deviation and cumulative distribution of the GPR function.

For our case, rather than using the Expected Improvement, we use the \textit{Log Expected Improvement}, which has been found to be nearly identical but is easier to numerically optimize\cite{ament_unexpected_2023-logei}. It is calculated as the logarithm of the expectation value of the improvement over the previous best value according to the surrogate model,
\begin{equation*}
    \label{eqn:log_expected_improvement}
    \text{LogEI}(x) = \log{\mathbb{E}_{\hat{f}(\vec{x})}[(\hat{f}(\vec{x}) - f_l^*)^+]}
\end{equation*}
The next point to be evaluated with the objective function is therefore 
\begin{equation*}
    \label{eqn:EI_next_point}
    \vec{x}_{l+1} = \argmax_{\vec{x}}\left(\;\text{LogEI}(\vec{x})\; \right)
\end{equation*}
After assessing the $\vec{x}_{l+1}$-th point using the true objective function, we add the pair $(\vec{x}_{l+1},\; f(\vec{x}_{l+1}))$ to the dataset $\mathcal{D} = \mathcal{D} \;\cup \{(\vec{x}_{l+1},f(\vec{x}_{l+1}))\}$ and start the next iteration. 
\subsection{Latent Bayesian Optimization}
Latent Bayesian Optimization attempts to resolve the issue of optimizing over large, discrete design spaces, such as, in our case, amino acid sequences describing peptide candidates. Rather than attempting to search for optimal peptide sequences in the discrete combinatorial space, we instead perform the optimization over a continuous-valued latent space, where we have access to both an encoder $E: X\rightarrow Z$ encoding sequences into a latent space, and a decoder $D:Z\rightarrow X$ that reconstructs the sequence corresponding to a latent point. Here $Z$ is typically the latent space of a generative model. The conventional choice is to use the latent space of a VAE, where the encoded mean $\vec{\mu}_\theta (\vec{x})$ of a trained VAE with fitted model weights $\theta$ is used in the optimization \cite{gomez-bombarelli_automatic_2018}. More formally, Latent Bayesian Optimization is the process of solving 
\[
\mathop{\argmax}_{\vec{x} \in X}f(\vec{x})
\]
by instead approximating the problem as 
\[
\mathop{\argmax}_{\vec{z} \in Z}f(D(\vec{z}))
\]
\textit{i.e.} Bayesian Optimization over the latent space of a generative model or other invertible mapping. The surrogate model $\hat{f}: Z \rightarrow \mathbb{R}$ approximates the relationship between latent points and their corresponding objective value; for a peptide sequence $\vec{x}$ and its encoded representation $\vec{\mu}_{\theta}(\vec{x})$, the surrogate model approximates $\vec{\mu}_{\theta}(\vec{x}) \rightarrow f(\vec{x}) $. At iteration $l$, the acquisition function determines the subsequent point in the latent space $\vec{\mu}_{l+1}$ to decode into a peptide sequence $\vec{x}_{l+1}$ we can evaluate with the objective function $f(\vec{x}_{l+1})$. In other words, rather than performing the Bayesian Optimization on what we want to design (peptide sequences) we perform it over a (learned) representation of those sequences. 

\subsection{``True'' Objective Function for Assessment of AMP Design Spaces}
In this work we are interested in finding peptide sequences with strong antimicrobial activity. We choose to use minimum inhibitory concentration (MIC) as a measure for antimicrobial activity. The MIC refers to the smallest concentration of peptide that inhibits the growth of bacteria in a standardized assay\cite{kaderabkova_antibiotic_2024-MIC}. We make this choice because it is a theoretically continuous, standard, and meaningful measure of antimicrobial activity, and standardized data is available for measurements on it \cite{witten_deep_2019}. Because a lower MIC corresponds to stronger antimicrobial activity, we maximize the negative of MIC: $M= -\log_{10}(MIC)$, thereby minimizing MIC.

In this pilot study, we wished to assess the impact of algorithmic choices and ensure our understanding of the design space prior to engaging in time-consuming experimental work. Therefore, we use a proxy model (also called an ``oracle'') based on traditional machine learning techniques (see next section) as our ``ground truth'' to estimate the MIC values of sequences for which no experimentally-measured MIC is available.   This is analogous to proof-of-concept small-molecule discovery projects that optimize for {computed estimates of drug-likeness and solubility}; a discipline-relevant proxy model is used to motivate an algorithmic exploration. In future work, we plan to leverage the procedure here to inform peptide design in more realistic contexts.

\subsection{Variational Autoencoder}
For the generative model whose associated latent space becomes our continuous design space of peptide sequences, we employ a variational autoencoder (VAE) based neural network. A VAE is a probabilistic deep learning architecture comprising two neural networks: the encoder $E_{\phi}:X\rightarrow Z$, which takes an input sequence and projects it to a continuous-valued latent space $z = E_{\phi}(x)$; and the decoder $D_{\theta}:Z\rightarrow X$ modules, which maps a point in the latent space back to the original input sequence space $\hat{x} = D_\theta(z)$ with $\phi$ and $\theta$ representing learned model weights \cite{kingma_introduction_2019}. Once the model has been adequately trained, we can use the latent space and the decoder to generate new peptides, sampling from a distribution that is similar to the distribution found in the training set. 

Training a VAE entails minimizing the Evidence Lower Bound (ELBO), a combination of a reconstruction loss, which measures the difference between the input sequence and the corresponding generated output sequence, and the Kullback-Leibler Divergence which functions to push the {probability distribution of the latent points} to be close to a Gaussian prior modeled by the latent space. We define the reconstruction loss as is typically done for sequence-based tasks \cite{zhang_dive_2023} as the cross-entropy of the probability of predicting the next token $x_{i+1}$ conditioned on a context $\vec{c}$, usually the previous tokens in the sequence: 
$$
\mathcal{L}_{CE}= -\sum_{i=1}^{N_{\text{batch}}}\sum_{c=1}^A p_c \log{(q(x_{i+1,c}| \vec{c}))}
$$
\noindent where $p_c$ is the probability distribution over the target token values, and $q(x_{i+1,c}| \vec{c})$ is the predicted probability distribution, and $A$ is the number of different tokens in the alphabet. Because the target token values come from the training dataset, and because we are using the maximum likelihood principle, we say that $p_c = 1$ when $c$ is the correct class and $=0$ when c is not, leading to the simpler $= -\sum_{i=1}^{N_{\text{batch}}}\log{(q(x_{i+1,c}| \vec{c}))}$. 

We use a VAE with a transformer architecture as described previously \cite{dollar_attention-based_2021,renaud_latent_2023}, known therein as a TransVAE. The transformer architecture is largely similar to the architecture originally proposed in \citet{vaswani_attention_2023}. {Input sequences were tokenized into integers then fed through a learned embedding layer. During training, sequences were predicted autoregressively, with the $(n+1)$-th token predicted from a context of the previous $n$ tokens , where the previous $n$ tokens are given directly from the training set and the reparameterized latent point representation $\vec{z}$. The} embedding layer and a positional encoding layer then together feed into the encoder module consisting of self-attention blocks followed by convolution blocks. Each self-attention block is a sequence of layers: layer normalization, multi-head self-attention, dropout, a residual connection, then layer normalization, feed-forward, dropout, and another residual connection. After the self-attention blocks, there is a convolution block consisting of three layers of convolution, pooling, and leaky ReLU. The convolution block outputs the mean and variance for the VAE latent space. The decoder module consists of largely the same architecture in reverse: convolution blocks followed by masked self-attention blocks. In our previous results \cite{renaud_latent_2023} we found evidence suggesting the TransVAE architecture may limit latent space interpretability compared to other architectures, however we did observe high performance at the assigned tasks. Here, we chose to use a transformer architecture because of their effectiveness at sequence-to-sequence tasks, and because traversing their latent space can correlate well with a predicted property \cite{renaud_latent_2023}, an important aspect of this work. Additionally, to limit the scope of this work, we focused on only the TransVAE architecture, but in future work will expand to other architectures and representation learning choices.

As previous work has shown, jointly training a property predictor can induce latent space organization \cite{gomez-bombarelli_automatic_2018, renaud_latent_2023}. In this work, we investigate the impact of the choice of predictive task and the availability of data associated with that task on the performance of the LBO algorithm. To do so, we train multiple TransVAE models jointly with a property predictor consisting of two feed-forward layers (see next section). The property predictor loss term $\mathcal{L}_{\text{PropPred}}$ is a simple mean-squared error between the property predictor's outputs $y_{\text{pred}}$, and the computed property values $y_{\text{true}}$, 
\begin{equation}
    \mathcal{L}_{\text{PropPred}} = \sum_i^{N_{\text{properties}}}\left(\frac{1}{N_\text{batch}}\sum_j^{N_\text{batch}}\left(y_{\text{pred},i,j} - y_{\text{true},i,j} \right)^2 \right)
\end{equation}
for a batch of size $N_{\text{batch}}$ and for $N_{\text{properties}}$ number of different properties we are predicting.

\subsection{TransVAE Models and Training}

To investigate the importance of additional information on the LBO algorithm and on the latent space that it traverses, we trained TransVAE models in conjunction with property predictors of different combinations of {the physico-chemical characteristics Boman Index, hydrophobicity, and charge}. Jointly training property predictors requires target values of these characteristics for corresponding (latent representations of) sequences. We computed Boman Index, net charge at a pH of 7.2, and hydrophobicity of every sequence in both the training set and test set. Boman Index is defined by summing the solubility values of individual amino acids in the peptide, normalized by length \cite{boman2003antibacterial}. It functions as an estimate of whether a peptide will bind to membranes or other proteins. For hydrophobicity, we use the Kyte-Doolittle scale \cite{kyte1982simple} and average the hydrophobicity values of amino acids in a given peptide for an overall estimate of peptide affinity with water. Charge at a pH of 7.2 is simply computed by summing the predicted charge of each amino acid in the peptide at a pH of 7.2. These physicochemical properties are of interest in AMP design because {of} they relate to whether an AMP will interact with or penetrate a cell membrane.   

To compute these physicochemical properties, we used the python package \textit{peptides} \cite{peptides_pypi}. The properties were then normalized to have zero mean and unit standard deviation using their respective training set means and standard deviations; the test set property values was normalized using the training set means and standard deviations. 

Overall, we studied the following models: (i) TransVAE trained jointly with a regressor directly predicting oracle values, {\emph{i.e.} $\log_{10}MIC$ values estimated by a predictor model}; (ii) TransVAE trained jointly with a regressor predicting Boman index values; (iii) TransVAE trained jointly with a regressor predicting charge values; (iv) TransVAE trained jointly with a regressor predicting hydrophobicity values; (v) TransVAE trained jointly with a regressor predicting both Boman index and charge; and (vi), TransVAE trained jointly with a regressor predicting Boman Index, charge, and hydrophobicity values.

To assess the effect of data sparsity, we trained each model with different numbers of available labels for the associated regressor.
For (i), we trained models with 100\% and 2\% (corresponding to 10,538 points in the training set) of the labels in the training set. For (ii)-(vi), we trained models with 100\%, 75\% (corresponding to 392,586 points in the training set), 50\% (corresponding to 261,647 points in the training set), 25\% (corresponding to 130,746 points in the training set), and 2\% (corresponding to 10,538 points in the training set) of the labels. The loss function for jointly training the TransVAE with a property predictor was the sum of the ELBO loss and the mean-squared error (predictor's loss), the latter of which was simply set to 0 for any data-point without an associated value in the training dataset. Thus, training proceeded in a semi-supervised or fully-supervised manner for the predictors, depending on the extent to which labels were made available for the training dataset.

All $2+5\times5 = 27$ models were trained on the associated loss function for 100 epochs using PyTorch.

\subsection{Dataset}
To train our TransVAE models we used a peptide sequence dataset consisting of 665,772 sequences. We sourced sequences from from AMP-specific databases DRAMP V3.0 \cite{shi_dramp3_2022}, APD3 \cite{wang_apd3_2016}, GRAMPA \cite{witten_deep_2019}, dbAMP 2.0 \cite{jhong_dbamp2_2022}, StarPep \cite{aguilera-mendoza_starpep_2019}, and general protein databases SwissProt and TREMBL \cite{the_uniprot_consortium_uniprot_2023}. Using peptide sequences from databases of known antimicrobial peptides ensures that the data distribution — and therefore the distribution being learned by the TransVAE — contains examples of sequences that are AMPs. Additionally, to ensure the training distribution also contained non-AMPs, we excluded those peptides from SwissProt annotated with the keywords antifungal, antimicrobial, antibiotic, antiviral defense, antiviral protein, secreted; although the impact of keyword combinations on predictive model performance has been discussed elsewhere\cite{sidorczuk_benchmarks_2022-negativeAMPsSelection}. Lastly, we clustered TREMBL sequences at 90\% identity using CD-HIT \cite{cd-hit_1,cd-hit_2}, then randomly sampled 7500 peptides at each length, yielding a subset of 668,343 peptides (the smallest length in the TREMBL set after clustering was 15). Once we removed duplicates,  the final dataset of unique peptides from every database was 665,772 sequences. This final amount was split into a training set of 523,848 sequences and a test set of 141,924 sequences where we ensured that the training set and test set had the same proportion of verified AMP and verified non-AMP peptides (SI Fig. \ref{sifig:histogram_of_dataset_sequence_lengths}).

\subsection{Principal Components Analysis for Dimensionality Reduction}
We leverage dimensionality reduction for two reasons. The first is to obtain low-dimensional visualizations of the 64-dimensional latent spaces associated with the different trained TransVAE models, enabling us to see whether the space has been organized by the property-of-interest. The second reason is to examine whether performing Bayesian Optimization over a reduced representation of the latent space impacts its performance 

We employ Principal Component Analysis (PCA) for dimensionality reduction, a commonly employed linear dimensionality reduction method \cite{jolliffe_principal_2016-PCA}. In PCA, the centered covariance matrix of the data is decomposed into its eigenvectors, which are then ordered by their corresponding eigenvalues. The top eigenvectors — the top \textit{principal components} — capture the most variance in the data points. PCA has the advantage of being a mathematically-invertible function and of being highly interpretable due to its linear nature; however, it suffers from an inability to accurately project nonlinear surfaces.

\subsection{Latent Bayesian Optimization Procedure}

To perform latent BayesOpt, we use the BoTorch Python package \cite{balandat2020botorch}, leveraging its SingleTaskGP Gaussian Process object with input unit hypercube and output standard normalization, as suggested by the codebase. As noted above, we use the Log Expected Improvement (LogEI) acquisition function, which offers improved numerical stability compared to conventional Expected Improvement \cite{ament_unexpected_2023-logei}. For each BayesOpt experiment, we perform five different optimization runs of 500 iterations ('oracle' calls). Each run is initialized with 100 points randomly sampled from predicted-$\log_{10}$(MIC) values (see Supplementary Information for additional details). Every BayesOpt experiment uses the same 5x100 initialization points, unless otherwise specified. 

For each of the models we trained, we compare searching through their 64-dimensional latent spaces with searching through different dimensional PCA projections of their latent space. To obtain the projected space, for a given VAE, we encoded the peptide training data set into the VAE's latent space, then performed a PCA decomposition of those points to obtain the components capturing most variation in the data. We used the top-$n$ components to make the BayesOpt search space, with $n$ varying across $2$, $5$, $10$, $20$, $32$. During a given BayesOpt iteration $l$, optimization of the acquisition function yields a candidate point $\vec{\mu}_{l+1}$ in the 64-dimensional latent space; alternatively if we are searching through a PCA reduced space, then the candidate point is mapped back to its corresponding 64-dimensional latent space through a PCA inverse projection, giving $\vec{\mu}_{i+1}$. We then decode $\vec{\mu}_{l+1}$ to get a peptide sequence $\vec{x}_{l+1}$ that is given to the oracle to get a predicted $\log_{10}(\text{MIC})$ value. We take the negative of that prediction to obtain the objective value for that point in the latent space (or in the PCA projection of the latent space). 

\section{Results}

\subsection{Property prediction leads to latent space organization, even in data-sparse conditions}
\begin{figure*}[h!]
    \centering
    \includegraphics[scale=0.39]{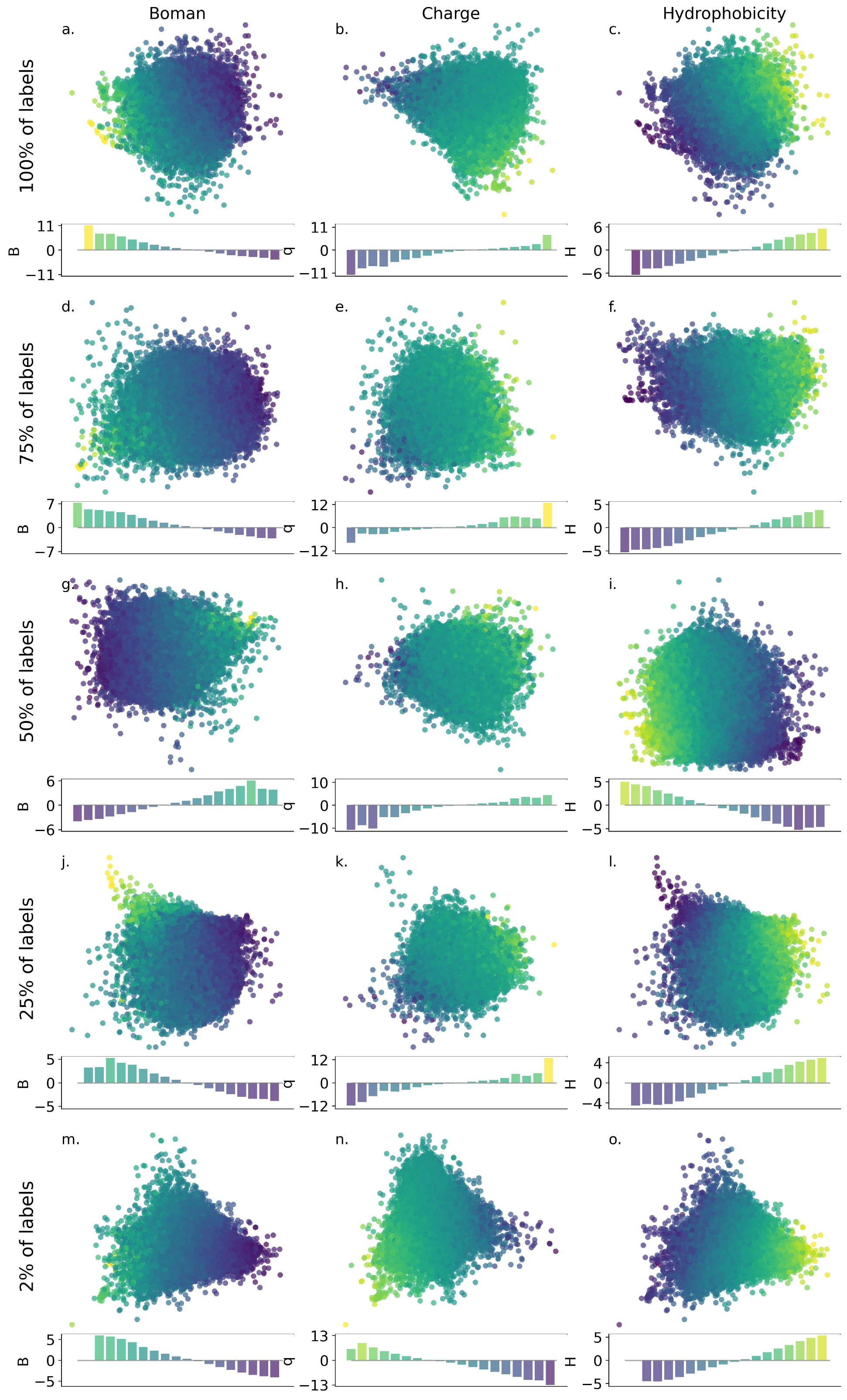}
    \caption{PCA-projected transVAE latent spaces jointly-trained by Boman index, hydrophobicity, and charge, at varying percentages of property values. The left column is coloured by Boman index. The middle column is coloured by charge. The right column is coloured by hydrophobicity. The visualized principal components are the two PCs  most correlated with the property value used for coloring the plot, as measured by a Pearson r value (see SI Table \ref{tbl:si:PCs_in5x3} for each PC pair). Barplots share $x$-axes with their corresponding scatterplots; bar heights are the average value of the column's property averaged over the points lying in a bar's $x$-axis interval width; monotonic trends suggest the (average in) property changes monotonically along the shared x-axis. Both the scatterplot and its shared barplot have the same colourmap. We observe property organization at each percentage of masking, and for each property, although with changes in relevant PC and direction.}
    \label{fig:pca_bch_train}
\end{figure*}

To assess the latent spaces associated with the different models, we quantified  the extent to which the top five principal components of a PCA projection correlated with different physicochemical properties used in training. We then employed these PCs to visualize relevant two-dimensional slices of the latent space. To do this we first computed each property over the training set of sequences. Next, we embedded those sequences in the full latent space of each model. Then we decomposed the space into its top five principal components. For each principal component, we computed the Pearson correlation coefficient ($r$) between it and the property values. {}

In Fig.~\ref{fig:pca_bch_train}, we visualize the PC components most highly correlated with Boman Index, Charge, and Hydrophobicity for the models trained with all three. In line with previous results \cite{gomez-bombarelli_automatic_2018,renaud_latent_2023}, we observe latent space organization when jointly-training a decoder and a property predictor. Additionally, training a property predictor predicting multiple different properties results in an organized latent space where different properties correlate most with different pairs of dimensions (Fig. \ref{fig:pca_bch_train} rows; SI Table \ref{tbl:si:PCs_in5x3}, particularly Boman index and charge). Furthermore, varying percentage of property labels — thereby limiting the amount of labels the model has access to in training — still results in latent space organization, as indicated by the persistence of an orderly color gradient between the first and last rows of Fig. \ref{fig:pca_bch_train}. Overall, increasing the number of properties and masking even a large proportion of the information for the predictor did not prevent the latent space from being organized. {(For a more detailed discussion of correlations between PC components and organizing variables, we refer the interested reader to Sect.~} \ref{sect:relationship-organizing-props-oracle} {in the SI.)}

Given that for 64-dimensional latent spaces we are visualizing just two-dimensional slices through PCA projections, it is possible that the organization is being induced by the projection. To check the extent to which PCA is distorting the original, high-dimensional manifold, and thus the extent to which it may be providing an illusion of organization when the high-dimensional cloud of points is not actually organized, we compute four manifold distortion metrics {as we have done previously} \cite{renaud_latent_2023}: trustworthiness and continuity \cite{venna_local_2006_trustworthinessContinuity}, steadiness and cohesiveness \cite{jeon_measuring_2022_steadinessCohesiveness}. 

These distortion metrics quantify different ways in which arrangements of points can be disrupted when transforming between the high-dimensional manifold and the low-dimensional manifold ("reduced space"). All run from zero, indicating high distortion (low faithfulness), to one, indicating low distortion (high faithfulness). Trustworthiness measures how frequently points that are not clustered together in the high-dimensional space become clustered together in the reduced space. Conversely, continuity measures the extent to which clustered points in the high-dimensional space become no longer clustered together in the reduced space. Unlike trustworthiness and continuity, which measure the number of nearest neighbours being lost or gained, steadiness and cohesiveness estimate the amount of stretching and compressing occurring when transforming between spaces. Steadiness quantifies the extent to which groups of points that are separated in the high-dimensional space remain separated in the reduced space. Cohesiveness quantifies the extent to which groups of points near each other in the high-dimensional space remain near each other in the reduced space. A more detailed description, including algorithmic details, are available in \citet{jeon_measuring_2022_steadinessCohesiveness}. 

\begin{figure*}[h!]
\centering
\includegraphics[width=\textwidth]{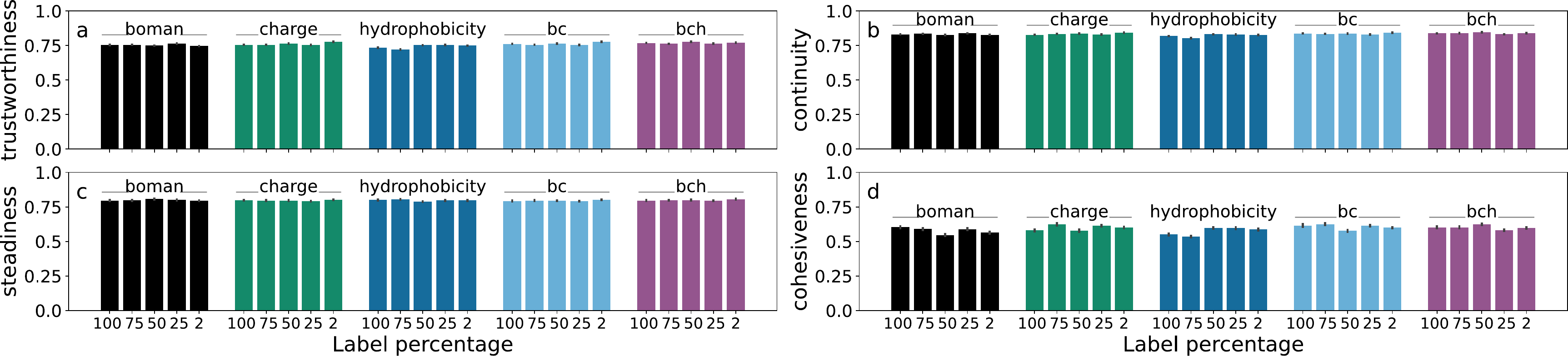}
\caption{Quantifying the distortion induced by using PCA on the latent spaces. For each quadrant, from left to right we have the property (or properties) used to organize the latent space; starting on the left, Boman index, charge (pH=7), hydrophobicity, Boman and charge (pH=7) (bc), and all three (bch). Variability in each quantity is estimated with 35 equally-sized subsamples of 42,000 different points in the latent space. (a) The estimated trustworthiness of the PCA space compared to the original 64-dimensional latent space. (b) The estimated continuity of the PCA space compared to the latent space. (c) The estimated steadiness of the PCA space compared to the latent space. (d) The estimated cohesiveness of the PCA space compared to the latent space. }
\label{fig:manifold_distortions_pca}
\end{figure*}

{We compute the extent of manifold distortion between the full latent space and the 5-dimensional PCA projections using a held out test set of sequences.} We observe similar levels of distortion between high-dimensional latent spaces and their PCA projections for models trained with different organizing properties and different percentages of labels (Fig. \ref{fig:manifold_distortions_pca}). Trustworthiness, continuity, and steadiness are all relatively high, with each organizing method exhibiting values $>0.75$ (Fig \ref{fig:manifold_distortions_pca}abc), suggesting high-quality projections with few artificial clusters (trustworthiness), minimal loss of clusters (continuity), and little stretching of distances between points (steadiness). Cohesiveness, lying near $0.6$ for all, is moderate, suggesting that compression is occurring (Fig \ref{fig:manifold_distortions_pca}d), as it necessarily must when moving from high to low dimensions. The similarity of the results and the comparatively low-to-moderate distortions observed provide confidence that the visualizations reflect extant trends between, and patterns in, the high-dimensional latent spaces.

We note in particular similar levels of distortion at various label percentages. Comparing within subpanels of Figure \ref{fig:manifold_distortions_pca}, the barplots are relatively stable in value. From these results, we infer the original high-dimensional latent spaces are being organized even when label percentage is low, and that we are not merely seeing an illusion of organization induced from the PCA projection. In conjunction with Fig.~\ref{fig:pca_bch_train}, this confirms that the number of labels can be relatively sparse and still result in the latent space being organized; perhaps most surprisingly in the case of having access to only 2\% of the property labels.  

For a final note, each model was trained to the same stopping point (100 epochs). At the end of training, the models all had similar training and validation loss values (SI Fig. \ref{fig:si:learning_curves}). We interpret this to mean that the peptide representations learned by the models are the essential difference between them, not their reconstruction accuracy.


\subsection{Bayesian optimization in a linearly-projected space can outperform optimization in the latent space directly}

Because BayesOpt can be more efficient in lower-dimensional search spaces, and because PCA provides a search space where each dimension is ordered by amount of explained variance, we hypothesized that reducing the dimensionality of the search space prior to BayesOpt via a simple linear projection may improve performance. To test this hypothesis, we compare the performance of BayesOpt in the full latent space associated with the 64-dimensional VAE with the performance of BayesOpt in lower-dimensional projections of this space and compare the results. To obtain the projected space, for a given VAE, we encode the peptide training data set into the VAE's latent space, then performed a PCA decomposition of those points to obtain the components capturing most variation in the data. We used the top-$n$ components to make the BayesOpt search space, with $n$ varying across $2$, $5$, $10$, $20$, $32$. 

\begin{figure*}
    \centering
    \includegraphics[width=\linewidth]{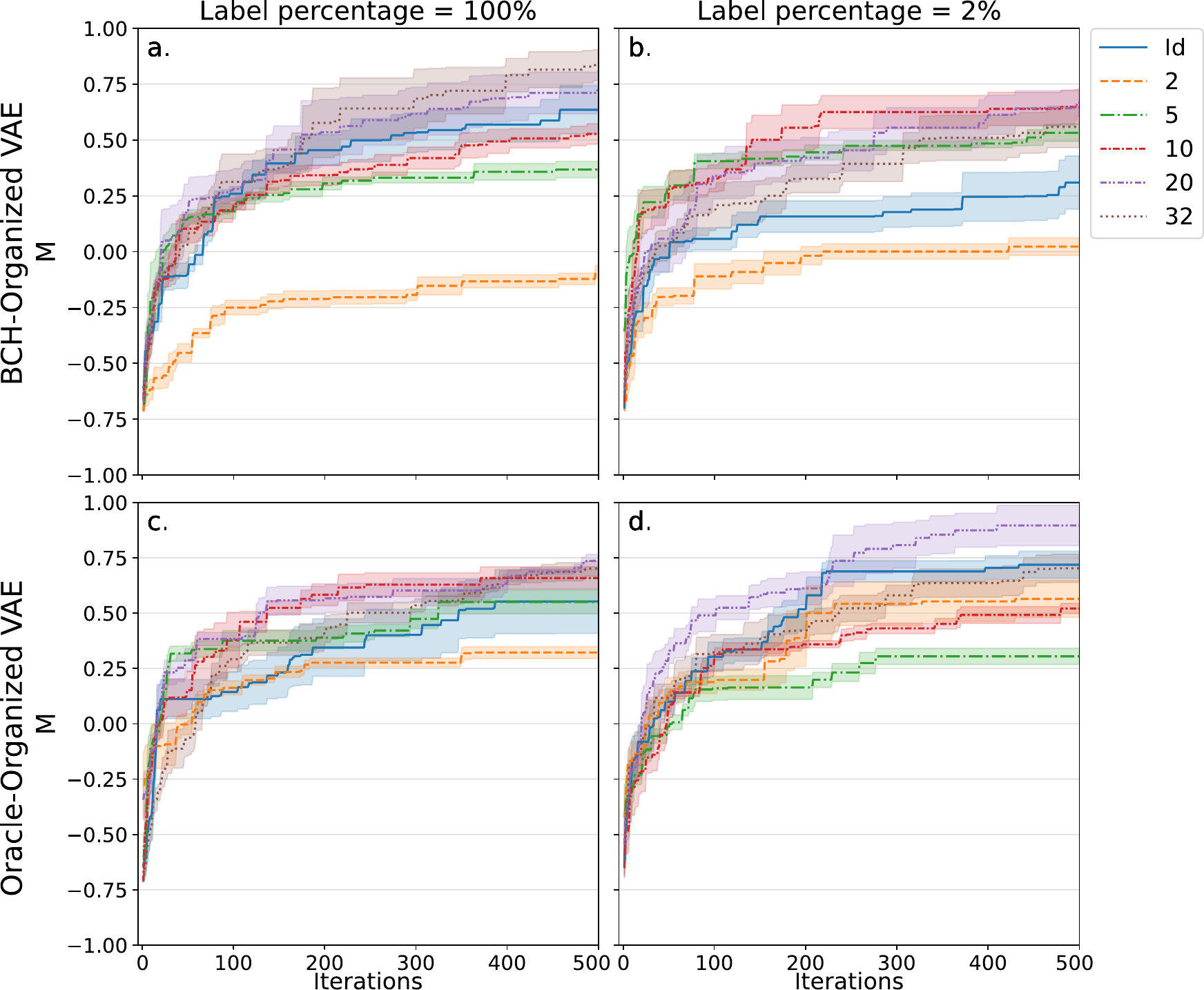}
    \caption{\textbf{Best scores identified through Bayesian optimization runs while varying the search space dimensionality.} Bayesian optimization results for searching through the 64-dimensional latent space (blue line, "Id" for "Identity projection"), and for searching through the projected latent space with varying numbers of PCA components, of BCH organized VAE (a,b) and of oracle-organized VAE (c,d). Label percentage corresponds to the percentage of property labels available during VAE training. Objective scores $\langle \mathcal{M}\rangle$ averaged over five BayesOpt runs with randomly initialized starting points. Error bars represent standard errors across the five runs.}
    \label{fig:boloop-varyPCAdims}
\end{figure*}

Somewhat to our surprise, we observe that BayesOpt in higher-dimensional linearly-projected spaces tends to result in a higher average final objective score than in lower dimensional ones, $\langle \mathcal{M}\rangle$ (Fig. \ref{fig:boloop-varyPCAdims}).  Indeed, for the BCH-organized space when 100\% of the property labels were available at VAE training time (Fig.~\ref{fig:boloop-varyPCAdims}a), we observe a monotonic increase with dimensionality in the final objective value achieved, although scores are comparatively similar for all linearly-projected spaces for about the first 100-200 iterations. For the Oracle-organized space when 100\% of the property labels were available at VAE training time and the BCH-organized space when the 2\% of the property labels were available at training time, we observe similar behavior for most of the searches in linearly-projected spaces but 10, 20, and 32 dimensions slightly outperform 5 dimensions by the final iteration, and all outperform $2$ dimensions, which flat-lines rather early (Fig.~\ref{fig:boloop-varyPCAdims}b,c). The trends are less clear for the Oracle-organized space at 2\% available label percentage; however, we do note that 20 and 32 dimensions outperform 2, 5, and 10 dimensions (Fig.~\ref{fig:boloop-varyPCAdims}d).

Despite the fact that we observe a correlation between available search space dimensionality and performance in the linearly-projected spaces, better performance is available in all cases in at least one of the lower-dimensional linearly-projected spaces than in the higher-dimensional full 64-dimensional space. When just 2\% of the labels were available at training time, BayesOpt in a linearly-projected space of the BCH-organized VAE outperformed BayesOpt in the original latent space for all but the 2-dimensional projection (Fig.~\ref{fig:boloop-varyPCAdims}b). In the case when 100\% of property labels were available, we observe better performance when optimizing in a 20 or 32 dimensional PCA reduced space than when optimizing in the latent space directly (Fig \ref{fig:boloop-varyPCAdims}a) for the BCH-organized space, while we observe better performance when optimizing in a 10, 20, or 32-dimensional PCA reduced space than when optimizing in the latent space directly for the Oracle-organize space (Fig. \ref{fig:boloop-varyPCAdims}c). Perhaps most interesting is the behavior we observe in the case when few labels are available to organize the VAE latent space, which is closely-related to real scenarios of finding optimally antimicrobial peptides with relevant experimental labels. In this case (Fig. \ref{fig:boloop-varyPCAdims}d), optimizing in 20 PCA components clearly outperforms optimizing in the corresponding latent space directly, with scores rising more rapidly to a higher final average score after 500 iterations ($\langle \mathcal{M_{\textrm{final}}}\rangle = $0.896 $\pm$ 0.092 compared to 0.719 $\pm$ 0.061). Optimizing in 32 PCA components performs about the same as optimizing directly in the latent space, reaching a final average score of $\langle \mathcal{M_{\textrm{final}}}\rangle = $0.702$\pm$ 0.063. Surprisingly optimizing in just 2 PCA components performs nearly as well: although the scores plateau at low values between $\sim$100$-$175 iterations, they rise rapidly thereafter, attaining a final average score of $\langle \mathcal{M_{\textrm{final}}}\rangle = $0.564$\pm$ 0.084, and mostly outperforming optimization in 10 components, and quite substantially outperforming optimization in 5 components.

When we expand to investigate all trained models, we also observe weak evidence of early higher performance in PCA-projected spaces (Fig.~\ref{sifig:final_values_barplot_it50}. Models are labeled using the following convention: S-prop-l\%, where S is 'id' for searching through the full latent space and 'PCAX' for searching through X-dimensional PCA projections, 'prop' is the organizing property, and l\% is the label percentage.) After just 50 iterations, most models are not performing particularly well yet, though we observe more variation in performance (both positive and negative) in BayesOpt runs in PCA-projected spaces than in the full latent space. The highest-performing models at this point are PCA5-bch-25\% (average score above 0.4) and PCA5-oracle-100\%, PCA20-oracle-100\%, PCA5-bch-2\%, PCA10-bch-2\%, and PCA20-oracle-2\% (average score above 0.3). Since all highest-performing models are in PCA projections, this provides weak evidence that PCA projections, particularly if given more properties or more relevant information can provide an early advantage; however, given that, e.g., PCA2-bch-100\% has an average score of -0.4, it is clear this is not a reliable or easily-accessible advantage.

Overall, we observe that performing BayesOpt in a PCA projection of VAE latent spaces may provides certain overall benefits. While the optimal number of PCA components to use is not obvious (\textit{e.g.} 32 vs 10 in Fig \ref{fig:boloop-varyPCAdims}ab), the number is less than the original latent space and in almost all cases provides a performance advantage. Additionally, it is substantially easier to visualize and interpret the search trajectory in lower dimensions, with two dimensions allowing direct visualization of the trajectory. 

\subsection{Searching through a PCA projection with a more relevant organizing property can provide an advantage.}
\label{sect:oracle-organization-compared-to-bch}
\begin{figure*}[h!]
    \centering
    \includegraphics[width=\linewidth]{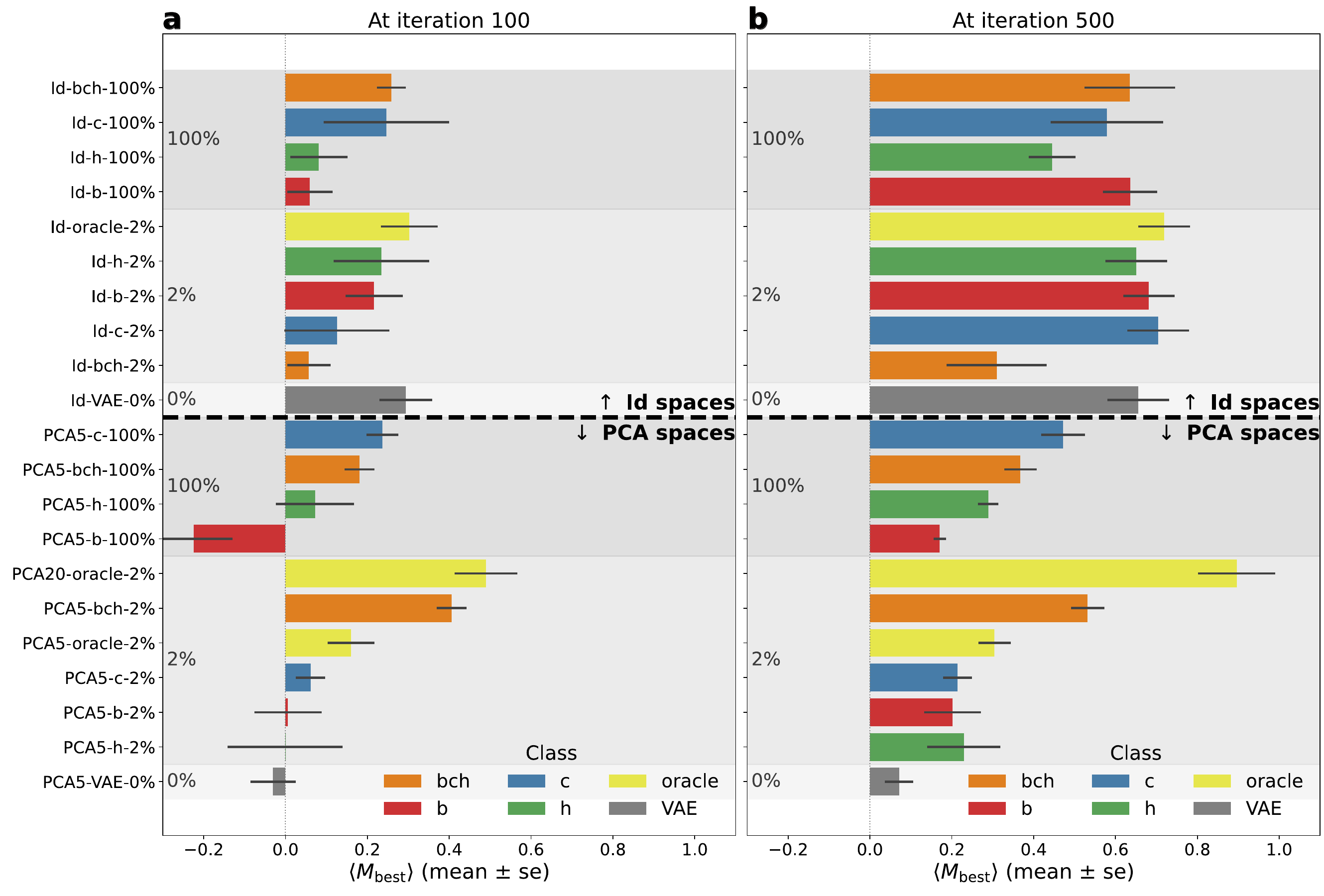}
    \caption{{Average best objective scores found after (a) $100$ iterations and after (b) $500$ iterations. Error bars correspond to the standard error of $M_{\text{best}}$ over five BayesOpt runs. Black dashed line separates those runs performed in the high-dimensional latent space (above the dashed line), and those runs performed in a PCA projection of the latent space (below the dashed line). Shaded regions denote the label percentage available at VAE training time, with darker regions corresponding to a higher label percentage. In (a) values are sorted within groups from largest to smallest, while (b) uses the same ordering as (a). Each class refers to the properties used for organizing the latent space, with "VAE" referring to an unorganized latent space. The full BayesOpt trajectories are depicted in SI Fig.} \ref{fig:boloops:exploitation:foreach-perc-vary-prop}.}
    \label{fig:final_values_barplot_it100and500}
\end{figure*}

Given the number of different physicochemical properties by which a latent space may be organized, we perform an ablation study testing different numbers and combinations. We hypothesized that organizing the latent space with more properties would lead to peptide representations more amenable to BayesOpt. We summarize the performance at different iterations in Figs \ref{sifig:final_values_barplot_it50} and {Figs}~\ref{fig:final_values_barplot_it100and500}{a}, \ref{fig:final_values_barplot_it100and500}{b}, where we investigate average best scores found under different conditions.  We primarily focus on PCA5 models. We choose the top five components to balance capturing a larger amount of explained variance (near 20\%, SI Fig. \ref{fig::si:explained_var_all_PCs}) with losing interpretability, and because three of the four scenarios where we varied the number of PCA components demonstrated five components performing better than two components (previous section).

We consider single property organization first. There is no clear advantage of any single property when searching through the 64-dimensional latent spaces--the best-performing single-property models compared to other models with only organizing property changed are id-c-100\%, id-b-75\%, id-h-50\%, id-h-2\% at 100 iterations ({Fig.}~\ref{fig:final_values_barplot_it100and500}{a}), and id-b-100\% and id-c-100\%, id-b-75\%, id-d-50\%, id-c-25\%, and id-b/c-2\%  at 500 iterations ({Fig.}~\ref{fig:final_values_barplot_it100and500}{b}). For PCA-projected spaces, we observe that charge-organized spaces tend to demonstrate better performance: at 100 iterations, PCA5-c-100\% clearly outperforms b/h-100\%, PCA5-c-75\% clearly outperforms b/h-75\%, PCA5-c-50\% clearly outperforms b/h-50\%, and PCA5-c-2\% clearly outperforms b/h-2\%. Only for the PCA5-25\% models is charge not the clear winner, and in that case, it performs nearly as well as PCA5-h-25\%. At 500 iterations, PCA5-c-100\% clearly outperforms b/h-100\%, PCA5-c-75\% clearly outperforms b/h-75\%, and PCA5-c-50\% clearly outperforms b/h-50\%. However, PCA-5-c-25\% performs worse than b/h-25\%, and all PCA5-2\% models perform roughly the same, displaying rather poor performance at <0.3 average best scores. Thus, charge is a better organizing property than Boman index or hydrophobicity, particularly when more labels are available. This trend matches with a similar trend in relevancy: charge correlates more strongly and has higher mutual information with the oracle values when looking at the full peptide training set (SI Section \ref{sect:relationship-organizing-props-oracle}). 

Considering multi-property organization, we still do not see clear trends in when searching through the full 64-dimensional latent space. After 100 iterations ({Fig.}~\ref{fig:final_values_barplot_it100and500}{a}), id-bch-100\% is one of the best performers (with performance about equivalent to id-c-100\%), and id-bch-75\% outperforms other models, but the highest-performing model for lower label percentages is one or more of the single-property models. After 500 iterations ({Fig.}~\ref{fig:final_values_barplot_it100and500}{b}), we see similar model performance across the board, though id-bch-100\% is one of the highest-performing models (tied with id-b-100\%) and id-bc-75\% is one of the highest-performing models (tied with id-b-75\%) but for lower label percentages single-property models are the best, and the performance is not too dissimilar within error bars. 

When searching through a PCA projection, there is not a clear advantage to using multi-property organization. After 100 iterations ({Fig.}~\ref{fig:final_values_barplot_it100and500}{a}), at 100\% and 75\% label percentage, charge clearly outperforms multi-property organization. At 50\% label percentage, PCA5-bc-50\% outperforms PCA5-c-50\%, at 25\% label percentage, PCA-bch-25\% by far outperforms any other models, and at 2\% the highest performing model is PCA5-bch-2\%.  After 500 iterations ({Fig.}~\ref{fig:final_values_barplot_it100and500}{b}), at 100\% and 75\% label percentage, charge still outperforms multi-property organization. For lower label percentage, multi-property organization does outperform single-property organization (PCA5-bc-50\% > PCA5-bch-50\% > PCA5-c-50\%, PCA5-bch-25\% is the clear best performer, PCA5-bch-2\% outperforms PCA5-bc-2\% and both clearly outperform the single-property organizers), but neither bc nor bch is consistently better than the other. This suggests that multi-property organization may be better at lower label percentage, an observation which is supported by the fact that the best-performing model we observed in the study was PCA-20-oracle-2\%, with a final performance of $\langle \mathcal{M_{\textrm{final}}}\rangle = $0.896 $\pm$ 0.092, where the oracle is itself a multi-property organizer, although it was not part of the ablation study. 

Overall, we find that performance depends more heavily on latent space organization when performing BayesOpt in PCA-projected latent spaces than when performing it in the full latent space. We find that for PCA-projected latent spaces, more relevant variables result in improved performance, with more types of information in the form of multiple properties or highly-relevant oracle values providing an advantage particularly in the very low percent label regime. 
Moreover, at iteration 500, using a linear projection of the oracle-organized latent space leads to the highest $\langle M_{\text{best}}\rangle$ over all experiments we tried. This suggests that the case of having very relevant data but very low label percentage (the scenario we find ourselves in with AMP datasets associated with experimentally-measured antimicrobial activity), could be a good use case for using a linear projection of the latent space as the BayesOpt search space.

\subsection{Bayesian optimization in a linearly-projected space may outperform that in a nonlinearly projected space}
{We additionally compared performing BayesOpt in a linearly-projected space to performing BayesOpt in a nonlinearly-project lower dimensional space on the fly by employing Gaussian Process Deep Kernel Learning (GP-DKL)} \cite{wilson_deep_2016_dkl}. {GP-DKL is a variant of a Gaussian Process that uses a neural network to augment the kernel function. By learning a projection $g_{\theta}(\vec{x})$ mapping design points into a feature space that the kernel function operates on, the conventional BayesOpt kernel, $k(\vec{x}_1,\vec{x}_2)$, is replaced by $k( g_{\theta}(\vec{x}_1),\;g_{\theta}(\vec{x}_2) )$, where $g_{\theta}$ is a neural network. The weights $\theta$ of $g_{\theta}$ are fit simultaneously with the parameters of the kernel $k(,)$ at each BayesOpt iteration. GP-DKL is an a promising alternative  approach to deriving a low-dimensional space for BayesOpt to traverse, as it operates directly on the high-dimensional space and leverages a learnable non-linear projection (as opposed to the static, linear one we have used) enabling the GP to determine the projection necessary to find an optimal point.}

\begin{figure}[h!]
    \centering
    \includegraphics[width=\linewidth]{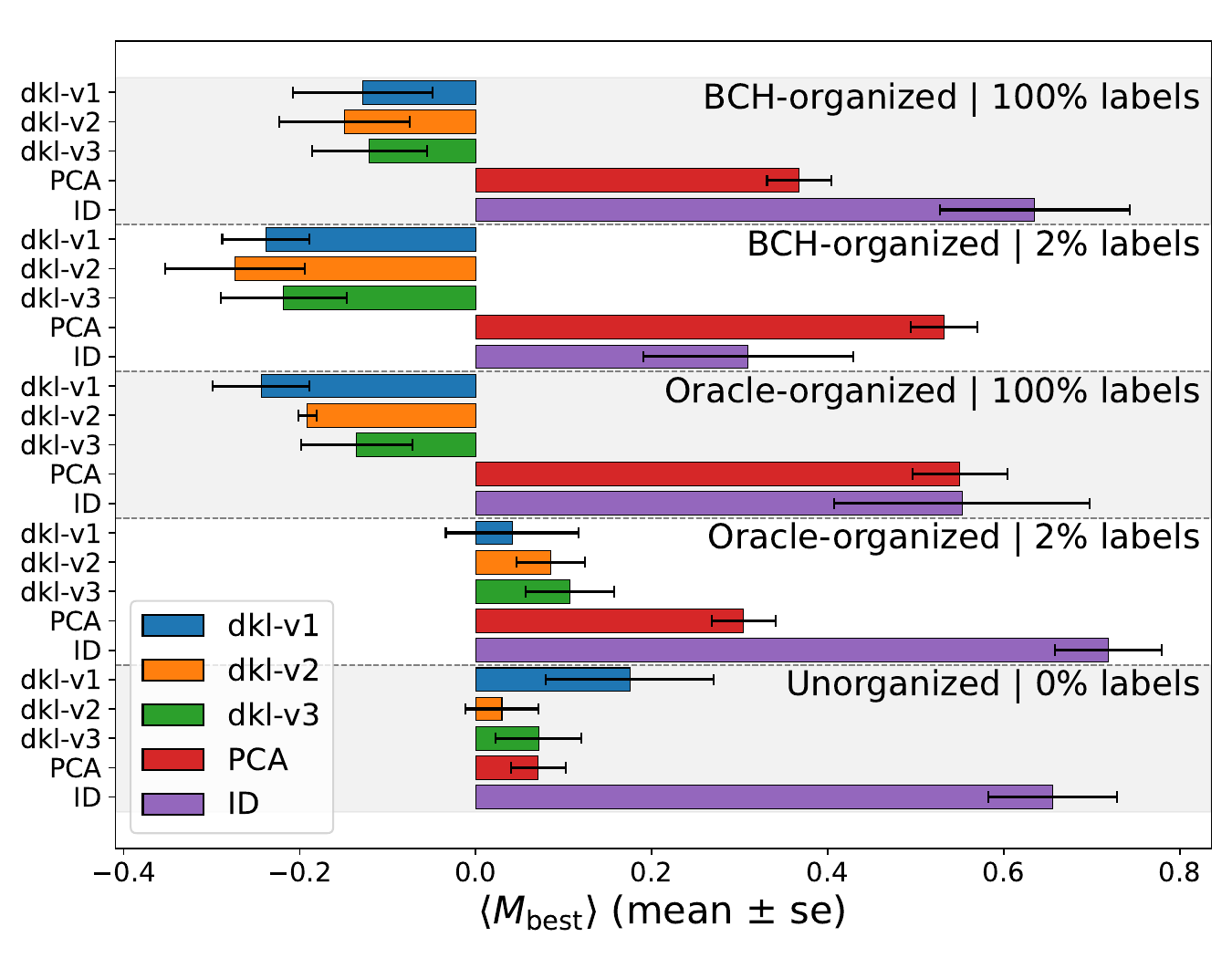}
    \caption{{Average best objective scores found after $500$ iterations. Error bars correspond to the standard error of $M_{\text{best}}$ over five BayesOpt runs. Abbreviations correspond to neural network architectures described in SI Sect.} \ref{sect:si:deep-kernel-learning}. The full BayesOpt trajectories are depicted in SI Fig. \ref{fig:boloops:exploitation:foreach-perc-vary-prop}.}
    \label{fig:final_values_DKL}
\end{figure}

{Here we implemented GP-DKL, and performed BayesOpt with five-dimensional GP-DKL in five of our TransVAE spaces: unorganized, BCH-organized with either 100\% or 2\% of the labels, and oracle-organized with either 100\% or 2\% of the labels. Across three different architectures for the GP-DKL neural network (dkl-v1, dkl-v2, dkl-v3; see SI Sect.~} \ref{sect:si:deep-kernel-learning} {for more details) we find that, in all but one case, BayesOpt in a five-dimensional PCA-projected space is able to consistently perform as well as, or better than, GP-DKL (Figs~} \ref{fig:final_values_DKL} and \ref{sifig:dkl-2x3} {)}.

{In an unorganized TransVAE model, two architectures from GP-DKL (dkl-v2, orange; dkl-v3, green) perform as well BayesOpt in a PCA-projection (red), but one GP-DKL variant (dkl-v1, blue) does slightly better than searching in the PCA-projection. However, we observe that searching directly in the high-dimensional space (purple) results in substantially higher values attained than either GP-DKL or BayesOpt in the linear-projection, thus pointing to fundamental limitations of searching in a lower-dimensional space without additional information to permit an informed choice of directions. }

{Once we move to organized spaces, there is a clearer difference. In the BCH-organized spaces, all three GP-DKL variants have average best scores below zero while the BayesOpt searching through the high-dimensional space and searching through the PCA-projection are able to by at least the 100th iteration. This trend also holds when we look at the BayesOpt runs done in the oracle-organized spaces. This may be a result of differences in data availability to the different algorithms. The PCA projector is fit to the (embedded) training set of the VAE, which is O(100k) points. This projector is then not updated throughout the BayesOpt loops. In DKL, the neural network projector is fit during the BayesOpt iterations on the fly, meaning it can search in different directions while severely restricting the amount of data available ( O(100) points in our scenario ) compared to the PCA projection. Further work could be done to optimize a nonlinear projection scenario.} 


\subsection{Searching through a PCA projection is associated with more exploration}
\label{sect:search-in-pca-more-exploration}
To further understand BayesOpt performance in our spaces, we examine how much exploration occurs; that is, how much of the space is probed by the search. We quantify the amount of exploration that occurs throughout a BayesOpt run through four lenses: (1) the amount of the total latent space (input) a BayesOpt run spans, which we estimate using the hypervolume of the sampled points; (2) the amount of the scores (output) a BayesOpt run spans, which we estimate using the variance in objective scores sampled and (3) the distance from the oracle SVR's training set; and (4) the amount of the relevant latent space (relevant input) a BayesOpt run spans, which we estimate using{, which we estimate} the total path length of the sampled points. We first characterize the exploration and then assess its relationship with exploitation as previously quantified by achieved best scores.

\begin{figure*}[h!]    
    \centering
    \includegraphics[width=\linewidth]{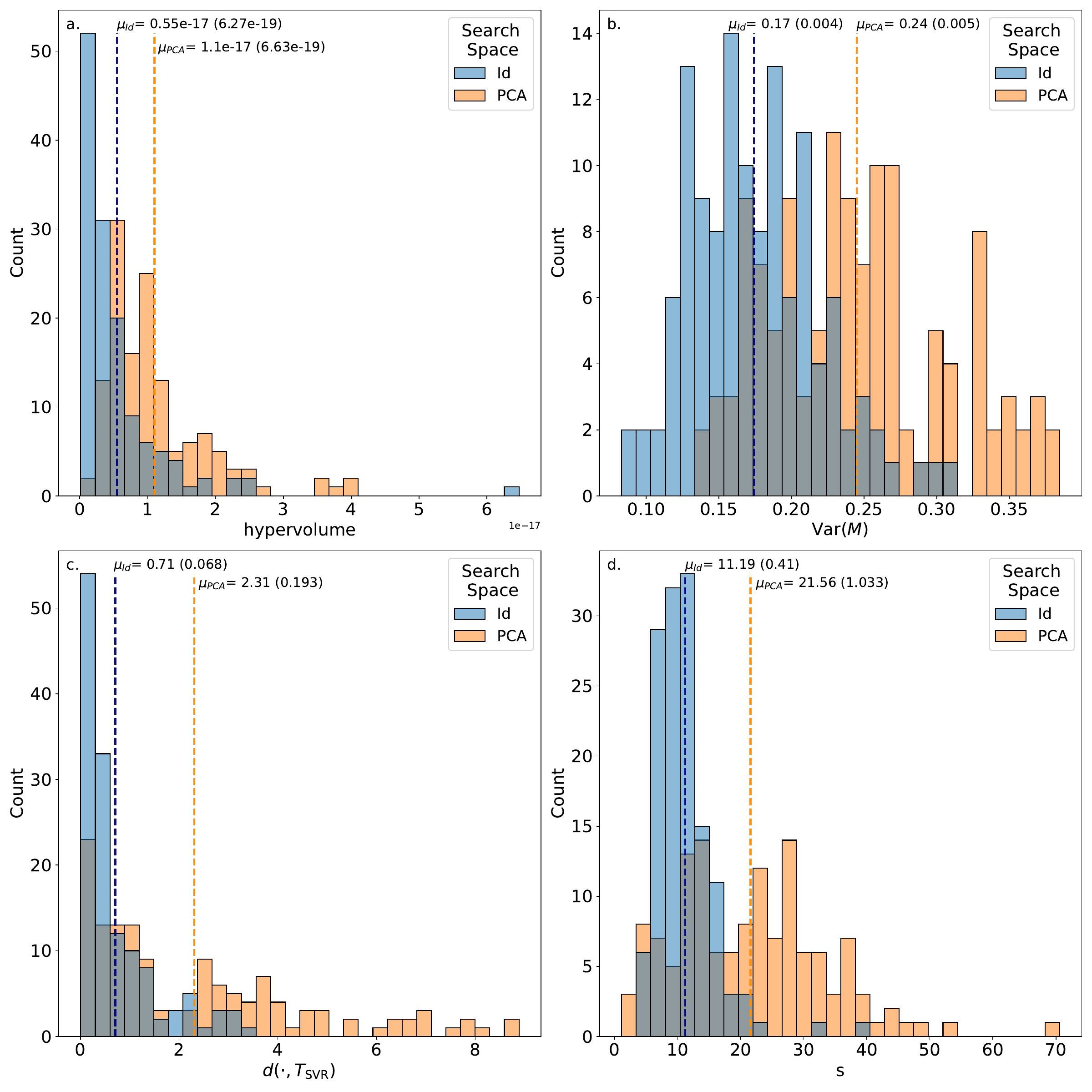}
    \caption{{Distributions of exploration quantities for optimization done in the full latent space (``Id'', blue) or the PCA projection of it (PCA, orange). Distribution of each run's: (a) computed hypervolume; (b) variance in scores sampled, $\text{Var}(M)$; (c) best sampled point's distance from the oracle's training set (Eqn.~}\ref{eqn:distance_pt_from_set}); {(d) total path length (Eqn.~}\ref{eqn:total_path_length}{). For (c) and (d) the distances are computed in the two PCs most correlated with the oracle values.}}
    \label{fig:exploration-histograms}
\end{figure*}

We compute the hypervolume of the input points sampled in each BayesOpt run, which is essentially a measure of the extent to which we explore different objective function inputs. The hypervolume was computed using the pygmo python package \cite{Biscani2020_pygmo_hypervolumes}; it computes the hypervolume of a set of points compared to a reference point; to choose our reference point, we used the corner of the search box with maximum value in each dimension ($10$ in each dimension). When looking at the distribution of resulting hypervolumes, we observe more hypervolume is explored on average when optimizing in a PCA search space compared to the high-dimensional search space; the  PCA distribution (Fig.~\ref{fig:exploration-histograms}a orange) is shifted to the right of the high-dimensional search space distribution (Fig.~\ref{fig:exploration-histograms}a, blue), with $\mu_{PCA, HV} = 1.1 \cdot 10^{-17} > 0.55 \cdot 10^{-17} = \mu_{Id, HV}$. This suggests that a broader slice (about 50\% more) of the underlying high-dimensional latent space was explored despite the PCA runs being confined to a subspace of it.

We further compute the variance in objective scores sampled throughout each BayesOpt run, which is essentially a measure of the extent to which we explore different objective function outputs. When comparing the distribution of variances in objective scores for runs searching through the PCA space (with center $\mu_{PCA, Var(M)} = 0.24\; (\pm 0.005)$) vs the high-dimensional latent space (with center $\mu_{Id, Var(M)} = 0.17\; (\pm0.004)$), we observe that the BayesOpt runs done in PCA spaces tend to have more variance in objective function scores (Fig. \ref{fig:exploration-histograms}b); $\mu_{PCA, Var(M)} = 0.24 > 0.17 = \mu_{Id, Var(M)}$. We test the hypothesis that these two means are the same using an independent samples t-test, yielding a p-value$=1.58\cdot 10^{-21}$, suggesting a statistically robust difference. This suggests that when optimizing through linear-projections of the high-dimensional space, more of the objective function is sampled (about 30\% more). 

Additionally, we compute the distance of the best point in a run from the SVR oracle's training set as an assessment of how exploration of input may be related to exploration of output and an assessment of the algorithm's ability to suggest novel optima. Because computing distances in high-dimensional spaces can be misleading, we computed these distances in a two-dimensional projection of the latent space. We construct this two-dimensional projection using the two principal components most correlated with oracle values. To do this for a run, we first encoded the oracle's training set into the high-dimensional latent space the BayesOpt run was done in. We then compute a PCA projection of the high-dimensional latent space. The oracle's training set is then projected into the PCA reduced space. Then, we compute the Euclidean distance between the best point found during the run and each point in the oracle's training set, in the PCA reduced space. The minimum distance found is used as the distance between the point and the oracle's training set $T_{\text{SVR}}$:
\begin{equation}
    \label{eqn:distance_pt_from_set}
    d(x,\; T_{\text{SVR}}) = \inf\{d(x,a):a \in T_{\text{SVR}}\}
\end{equation}
We observe that the BayesRuns searching through linear-projections of latent spaces have a distribution of distances from the oracle's training set has a mean of $\mu_{PCA}=2.31\;(\pm\;0.193)$. The distribution of runs searching through the high-dimensional latent space has a mean of $\mu_{Id}=0.71\;(\pm\;0.068)$. From the distribution of distances computed, we observe that the BayesOpt runs searching through linear-projections of the latent space tended to result in final points about 225\% farther from the oracle's training set when compared to the BayesOpt runs searching through the full high-dimensional latent space (Fig. \ref{fig:exploration-histograms}c); $\mu_{PCA}=2.31 > 0.71 = \mu_{Id}$, a statistically robust difference, with a p-value$=1.21\cdot 10^{-13}$.

Lastly, we compute the total path length of a BayesOpt run as a way of measuring the exploration of the relevant input space. For the $k$-th point sampled during a BayesOpt run, we compute the Euclidean distance from it to the $(k-1)$-th point sampled during the run, in the two PCs most correlated with the oracle values of the full VAE training set. We compute the distance between each consecutive pair of points in the PCA reduced space for the run. Then, we sum the distances to find the total path length in the PCA reduced space. That is, for a given BayesOpt run with sampled latent space points $\{\vec{\mu}_1, \vec{\mu}_2, \dots, \vec{\mu}_{500}\}$, projected into a two-dimensional reduced space $\{\vec{p}_1, \vec{p}_2, \dots, \vec{p}_{500}\}$ we compute its total path length as
\begin{equation}
    \label{eqn:total_path_length}
    s = \sum_{i} || \vec{p}_{i+1} - \vec{p_{i}} ||_2
\end{equation}
From the distribution of path lengths we observe an increase in total path length by about 93\% from searching directly in the high-dimensional compared to the linearly-projected space (Fig. \ref{fig:exploration-histograms}d); the centre of the distribution of path lengths for searching in a linearly-projected space is $\mu_{PCA} = 21.56\;(\pm\; 1.033)$; compared to the distribution of high-dimensional search space path lengths, whose centre is $\mu_{Id}=11.19\; (\pm\;0.410)$; this difference is statistically robust, with p-value$=4.18 \cdot 10^{-18}$. {We further quantified the exploration in sequence space directly and found similar results (Sect.} \ref{sisect:sequence-space-exploration}{).}

\begin{figure*}[h!]
    \centering
    \includegraphics[width=\linewidth]{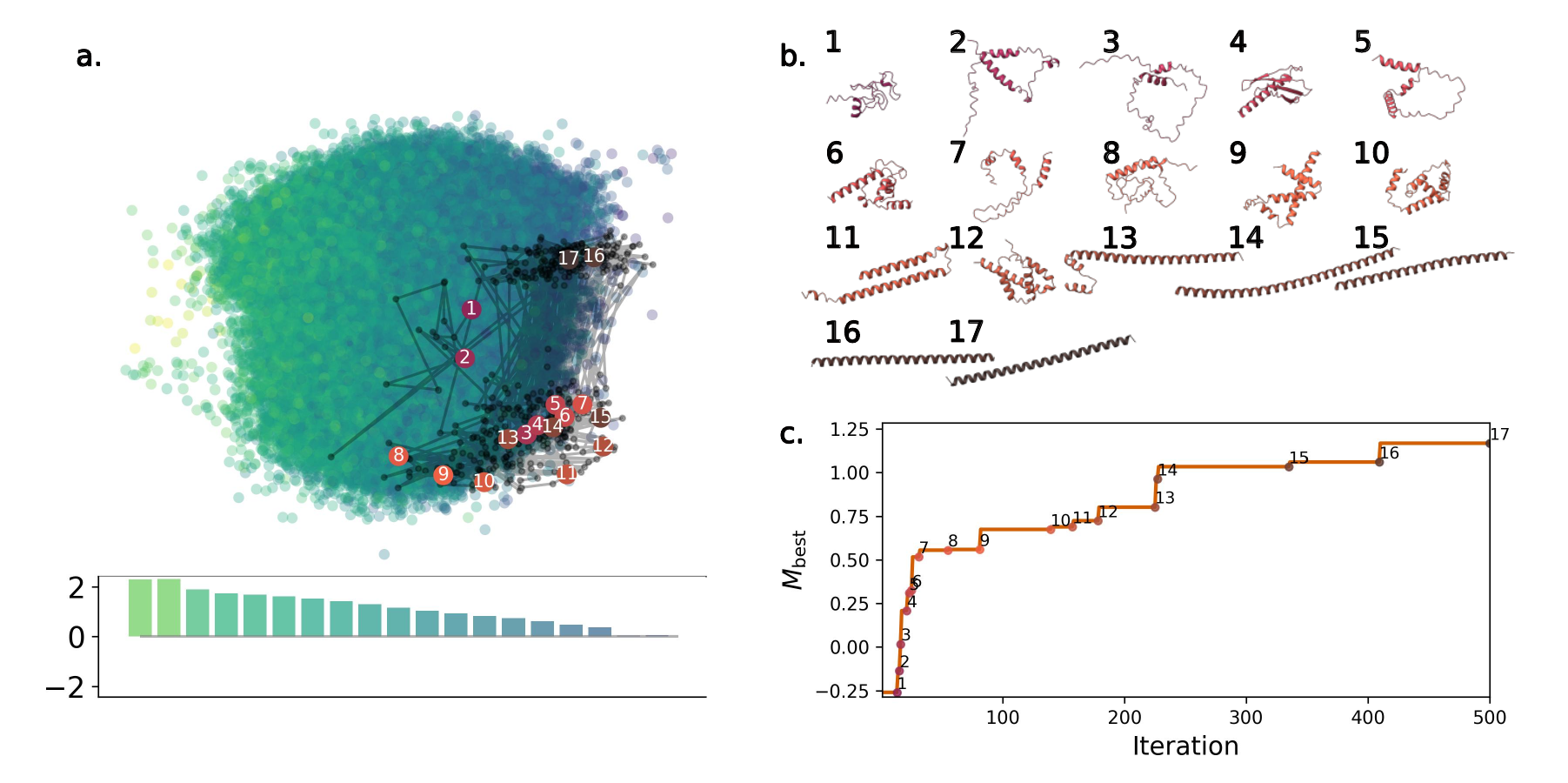}
    \caption{Best peptides identified during a BayesOpt run. (a) PCA plot of the associated oracle-organized latent space (at 2\% of labels available), with barplot depicting an average decrease in predicted-$\log_{10}(\text{MIC})$ from left-to-right across the latent space. The latent space locations of the best peptides are depicted with large circle markers; white integers are the sequence in which these peptides were identified during a BayesOpt run, but are not the iteration at which they were identified. (b) ESMFold-predicted structures \cite{lin_evolutionary-scale_2023-esmfold} of the best peptides found during the BayesOpt run, visualized using ChimeraX \cite{meng_span_2023-chimerax}.{In this case we note a strong increase in helicity with oracle value, a reasonable – but biophysically misleading – proxy that the oracle might learn for antimicrobialness, which may be ameliorated with higher-fidelity evaluations such as simulation.} (c) The corresponding best score $M_{\text{best}}$ found throughout the BayesOpt run.}
    \label{fig:sequence_of_peptides}
\end{figure*}

We further examined the BayesOpt runs in their associated latent spaces qualitatively. To do this, we plotted the peptides sampled during a BayesOpt run in a two-dimensional PCA projection of the associated latent space using the two PCs most correlated to oracle values. In Figure \ref{fig:sequence_of_peptides}, we depict a full example trajectory showing the latent space (panel a), the structures of 17 selected peptides along the run as predicted by ESMFold (panel b), and their corresponding scores (panel c). This trajectory illustrates the best-scoring run of five that were performed in a 20-dimensional projection of the oracle-organized latent space. This figure has a clear interpretation and allows us to gain a strong physical intuition for the progress of the search. We see the BayesOpt trajectory begin at a low objective value (1) and rapidly increase by moving to the right in the latent space, towards the average direction of increased objective value (lower MIC). After attaining point 7, BayesOpt performs a large explorative jump to the left to point 8, which only marginally increases the best score. It then proceeds by generally drifting to the right and eventually jumping upwards (point 16). Concomitantly, we see that as the search proceeds, the predicted structures increase strongly in helicity. {Given the over-representation of $\alpha$-helical AMPs in most datasets, helicity is a reasonable proxy for antimicrobialness that the BayesOpt is exploiting; a straightforward example of reward hacking. This confirmation of reward hacking provides another reason for using the most relevant data possible (less hackable), even if less of it is available (because e.g. it is expensive to compute).}

\begin{figure*}[h!]
    \centering
    \includegraphics[width=0.85\linewidth]{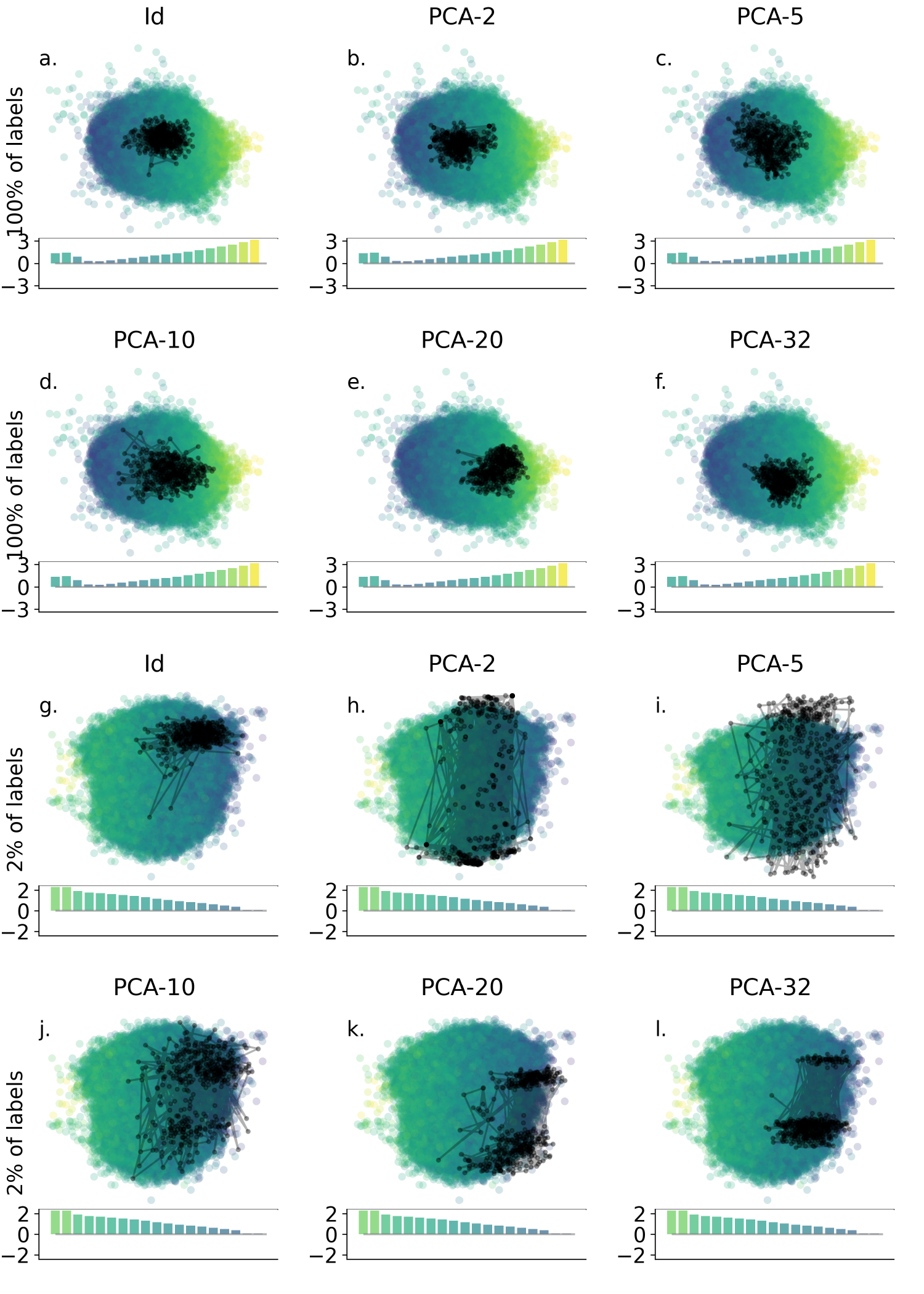}
    \caption{The BayesOpt run with the best objective score is plotted on a 2D visualization of oracle-organized latent spaces. (a-f) the oracle-organized VAE had access to 100\% of the property labels during training. (g-l) only 2\% of the property labels were available during training. Optimization was done in different projections of the underlying latent space: (a,g) directly in the 64-dimensional latent space, (b,h) in the top 2 PCs of a PCA projection, (c,i) in the top 5 PCs of a PCA projection, (d,j) in the top 10 PCs of a PCA projection, (e,k) in the top 20 PCs of a PCA projection, (f,l) in the top 32 PCs of a PCA projection.}
    \label{fig:si:bayesopt_run_in_oracle_latent_spaces}
\end{figure*}

In Fig.~\ref{fig:si:bayesopt_run_in_oracle_latent_spaces}, we visualize the best-scoring run of five for all oracle-organized latent spaces at 100\% and 2\% label percentage. (For more information cf.~also SI Figs ~\ref{fig:si:bayesopt_run_in_boman_latent_spaces_colouredByOracle}-\ref{fig:si:bayesopt_run_in_bch_latent_spaces_colouredByOracle}). As previously quantified, we visually observe that BayesOpt runs tend to cover more of the 2D visualizations when their associated search space was a PCA projection (\textit{e.g.}, compare Fig.~\ref{fig:si:bayesopt_run_in_oracle_latent_spaces}g with Fig.~\ref{fig:si:bayesopt_run_in_oracle_latent_spaces}h-l {}). Additionally, there is substantially greater spatial exploration when the percentage of labels is low, although as we have seen, BayesOpt is still able to converge to satisfactory optima in this case. In particular, when comparing the low-label regime (2\%) with the full label regime (100\%), we observe that the BayesOpt runs explore substantially more of the visualization; this is especially the case when BayesOpt runs search through the top two, five, or ten PCA dimensions. 

The previous observations are certainly due at least partly to the fact that the search procedure is more restricted to the visualized directions for the PCA projections than for the full search space. This leads us to note that interpretation of BayesOpt runs is easiest when the search space is similar to the visualization. We can also observe this in that when the organizing property is a physicochemical property the BayesOpt runs are centered, appearing to only explore a portion in the directions used to make the visualization, since the projection is not well-aligned with the directions of the search. However, when looking at the oracle-organized spaces, we can clearly see the BayesOpt runs identifying the side of the space that contains better objective function values (low value oracle predictions corresponds to higher objective score) after covering more of the visualized region, particularly when fewer labels are available.  

\subsection{Performance is correlated with sampling a range of objective values but not with exploration of input design space}

\begin{figure*}[h!]
    \centering
    \includegraphics[width=\linewidth]{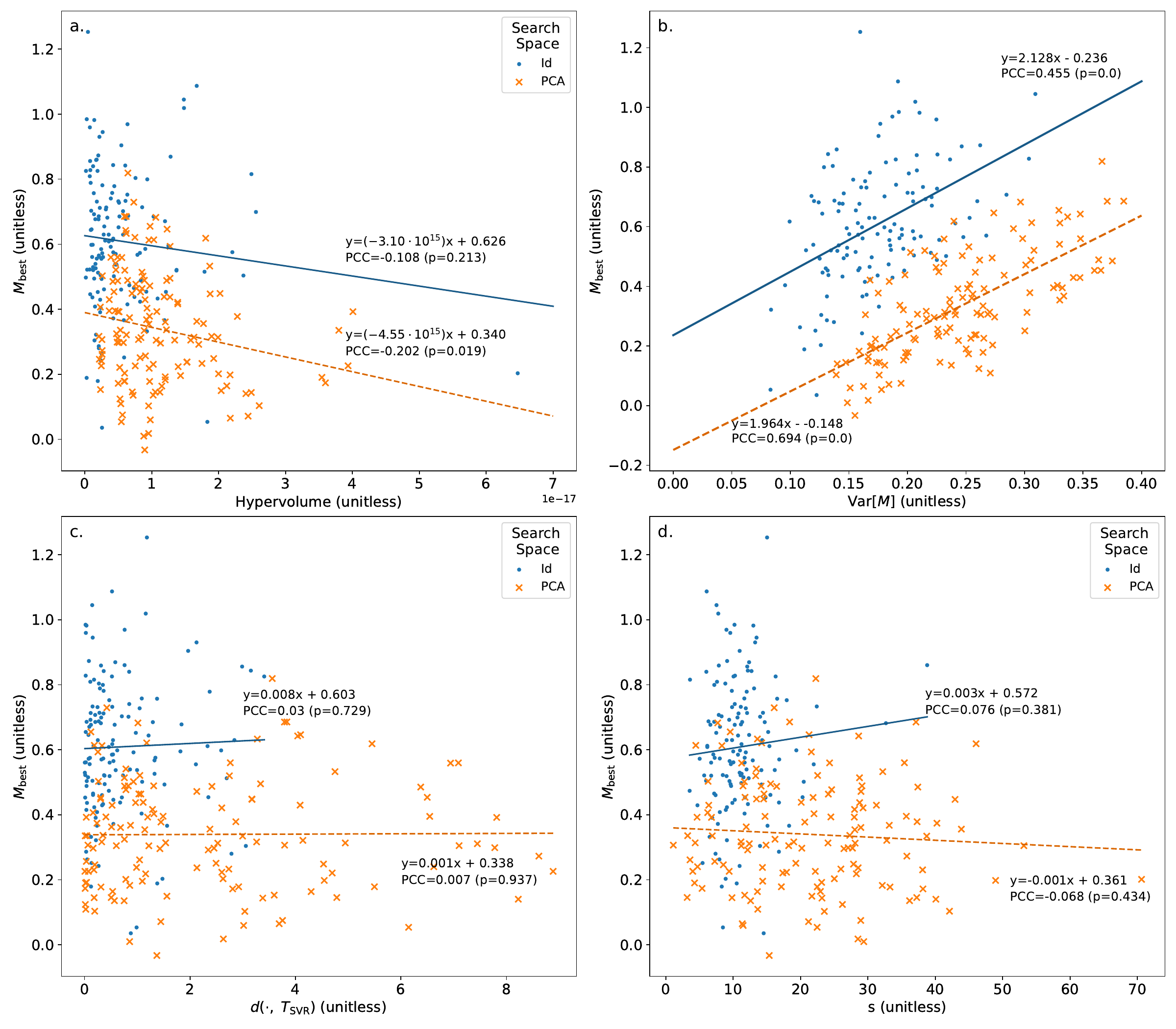}
    \caption{Relationship between $M_{\text{best}}$ and (a) the hypervolume explored in that run; (b) the variance in objective scores $\text{Var}[M]$ sampled throughout the run, for each BayesOpt run we performed; (c) the distance between the best sequence's latent representation and the SVR's training set; and, (d) the path length $s$ of the corresponding BayesOpt run. Each BayesOpt run searched either over ("Id", blue dots, solid regression line) a 64-dimensional latent space, or ("PCA", orange crosses, dashed regression line) the PCA projection of the latent space, with search box edges ranging from $-10$ to $10$ in each dimension. Solid (dashed) lines correspond to regressing on the dots (crosses). Pearson correlation coefficients (PCCs) are written next to their corresponding point group, with p-values in parentheses. (a) We observe a very weak negative correlation between the normalized hypervolume of latent space explored during a BayesOpt run, and the best score found during that run. (b) We observe a moderate (PCC=$0.498$ dots) to strong (PCC=$0.652$, crosses) correlation between variance in scores sampled throughout a run, and the best score found during that run. We observe minimal correlation between the best objective score sampled in a run, and (c) that run's final distance from the SVR's training set, and (d) that run's total path length. We compute a run's path length by taking the Euclidean distance between point $\vec{\mu}_k$ and $\vec{\mu}_{k-1}$ and summing over all $k=2,\dots,500$. For (c) and (d) the distances are computed in the two PCs most correlated with the oracle values.}
    \label{fig:boloops:exploration:scatterplot-bestScore-vs}
\end{figure*}

In this section, we assess whether there is a relationship between exploration as quantified in the previous section and BayesOpt performance as measured by the final best score of the run. We first compare the hypervolume of input points sampled during a run with the best objective score during that run (Fig. \ref{fig:boloops:exploration:scatterplot-bestScore-vs}a). For the BayesOpt runs done in PCA reduced spaces, we observe a weak negative PCC of -0.202 that is statistically robust ($p=0.019<0.05$). For BayesOpt runs done in the 64-dimensional latent spaces, we observe a very weak negative PCC of -0.108 that does not appear statistically robust.

 We then compare the variance of scores during a run with the best objective value encountered during that run (Fig. \ref{fig:boloops:exploration:scatterplot-bestScore-vs}b). For BayesOpt runs done in PCA reduced spaces, we observe a strong Pearson Correlation Coefficient (PCC) of 0.652 that is statistically robust with a p-value $<0.001$ (Fig. \ref{fig:boloops:exploration:scatterplot-bestScore-vs}b, crosses). For BayesOpt runs done in the 64-dimensional latent spaces, we observe a moderate PCC of 0.498 that is statistically robust with a p-value $<0.001$ (Fig. \ref{fig:boloops:exploration:scatterplot-bestScore-vs}b, dots). 

Additionally, we compare the best objective score found in each run to the distance of the corresponding point to the SVR's training set and to the total path length of a BayesOpt run in two dimensions (Fig. \ref{fig:boloops:exploration:scatterplot-bestScore-vs}c,d). We observe no statistically significant correlation in either of these values.

Overall, sampling a greater range of objective scores results in better final scores, while exploration of the input space as measured here is not strongly correlated with performance.

\section{Discussion and Conclusions}
In this article, we have trained a series of Transformer-based VAE models for peptide sequence generation, which we have used as continuous search spaces for latent Bayesian optimization. {We presented a version of latent Bayesian optimization with additional dimensionality reduction, and compared it with traditional latent BayesOpt and BayesOpt with Deep Kernel Learning}. Inspired by a desire to better align visualization/analysis techniques with algorithmic properties, we took PCA projections of the latent spaces--previously used solely for visualization--and performed BayesOpt directly in them. We assessed the properties of the latent spaces themselves, compared our low-dimensional BayesOpt search with direct Bayesian Optimization in the full 64-dimensional latent space of the trained generative model, and connected latent space properties with resultant properties of the BayesOpt search trajectories. 

In line with previous work \cite{gomez-bombarelli_automatic_2018,renaud_latent_2023} we found that jointly-training a property-predictor and a VAE can lead to an organized latent space. Expanding on that work, we investigated whether such organization persists even in a semi-supervised scenario, finding evidence that just $2\%$ of property labels suffices to induce organization along that property; additionally we showed that jointly-training with multiple properties can lead to organization along each property. 

We employed these organized latent spaces for projected latent BayesOpt and traditional latent BayesOpt. When we varied the number of PCA components to optimize in, we found that optimizing in a PCA projection could outperform optimizing directly in the high-dimensional latent space, but it is not obvious \textit{a priori} how to select the optimal number of PCA components to optimize over. In general, using projected latent BayesOpt may provide an early or late advantage and can show impressive performance; however, it is less reliable than performance in the full latent space, which demonstrates less variability in performance with latent space organization and other hyperparameters. 

To better understand the use of projected latent BayesOpt, we assessed the effects of different organizing properties of the latent space. We hypothesized that if the properties provided a weak but ample signal regarding the oracle, that it may improve the BayesOpt performance through either increasing the rapidity at which it found potent peptides, or through increasing the max value attained throughout an optimization run. We find that, among the single physicochemical property spaces, charge tends to give better BayesOpt performance in terms of final objective values, and how quickly the objective value increases. This occurs primarily when searching through the linear projection of the original 64-dimensional latent space, but also maintained strong performance when searching through the full high-dimensional space. As charge is the most informative of the physico-chemical properties we studied, this demonstrates that BayesOpt in PCA-projected spaces works best when the space is organized by relevant information. We also demonstrated that at low label percentage, we could obtain high performance from employing multiple organizing properties or using the oracle. Overall, in a real-world situation, we expect to see good performance from projected latent BayesOpt if the space is organized by a highly-relevant property; the more types of information available at low label percentage, the better.

Despite the fact that the performance of projected latent BayesOpt can be more variable and more sensitive to latent space organization than latent BayesOpt, projected latent BayesOpt has a significant advantage in terms of interpretability. We can more easily and accurately visualize and understand the trajectories of projected latent BayesOpt. In addition, we showed that projected latent BayesOpt explores more of the search space than latent BayesOpt, in terms of both full hypervolume and exploration along relevant dimensions. Furthermore, the variability in performance may also be framed as heightened exploration of output scores sampled, which we showed to be correlated with better performance. Although projected latent BayesOpt has a weakness, we suggest that it can be ameliorated by running multiple short trajectories in spaces of different dimensionalities, and it comes with the advantage of better understanding of the search procedure. We also note that this work further underscores an important point we have previously raised \cite{renaud_latent_2023}, which is that naive use of low-dimensional projection algorithms to visualize latent spaces of generative models can be misleading. The community should not over-rely on the assumption that a two-dimensional PCA or t-SNE projection is sufficient to reveal qualities of the latent space, at least not without further analysis.

In sum, latent Bayesian optimization can provide an approach to identifying highly-potent antimicrobial peptides leveraging the continuous-valued latent spaces of generative models. We found that VAEs based on transformers can have their latent spaces organized by multiple different physicochemical properties simultaneously through the joint-training of multiple property predictors with the VAE. We further compared whether using physicochemical properties to organize the latent space is better than using few oracle values, finding evidence {(Sect.} \ref{sect:oracle-organization-compared-to-bch}) that using few oracle evaluations can be better when restricting the search space to linear projection of the latent space; \textit{i.e.} having a more relevant organizing property with few labels may be a good use case for searching through PCA projection of the latent space. Using the linear projection comes with the additional benefit of to being more interpretable, and easier to visualize.

{Although we have primarily discussed our work in the context of AMP design, given that BayesOpt is an efficient, iterative algorithm for searching a design space to optimize a function under limited data, our results extend straightforwardly to any peptide design question. The only limitation is the definition of a relevant objective function. For example, one could optimize for cell-penetrating peptides through in silico measurements of maximum force experienced by a peptide as it is pulled through a lipid bilayer in constant-velocity steered molecular dynamics; alternatively, one could optimize for anticancer peptides through in vitro measurements of the minimum peptide concentration required to inhibit cancer cell growth, or through in silico approximations of binding against a target protein. More broadly, the application of BayesOpt to a particular peptide sequence design challenge is limited primarily by the availability of an oracle that can be regularly queried for estimates of the underlying objective.}

\section*{Author Contributions}
\textbf{JM}: Conceptualization; Methodology; Software; Validation; Formal Analysis; Investigation; Data curation; Writing — original draft; Writing — review \& editing; Visualization. \textbf{RAM}: Conceptualization; Methodology; Resources; Writing — original draft; Writing — review \& editing; Supervision; Project administration; Funding acquisition. 

\section*{Conflicts of interest}
There are no conflicts to declare.

\section*{Data Availability Statement}

{The code for model training and analysis can be found at} \url{https://github.com/Mansbach-Lab/compare-latent-spaces-amps/tree/main}. {We note that for certain model training we employed the publicly available GRAMPA dataset} \cite{witten_deep_2019}. {Trained model checkpoints and datasets are released through Zenodo} \url{https://zenodo.org/records/17872434} {with DOI} \url{https://doi.org/10.5281/zenodo.17872449}.

\section*{Acknowledgements}
This research was supported in part by Discovery Grant \#RGPIN-2021-03470 from the National Sciences and Engineering Research Council of Canada. This research was enabled in part by support provided by Calcul Quebec (\url{www.calculquebec.ca}) and the Digital Research Alliance of Canada (\url{https://alliancecan.ca}). This research was undertaken, in part, thanks to funding from the Canada Research Chairs Program under grant number CRC-2020-00225. JM acknowledges an NSERC CGS-D scholarship, as well as the motivating support of the Mansbach lab, particularly Mohammadreza Niknam Hamidabad and Adam Graves for expressing interest and wanting to see this manuscript finished.





\bibliography{references} 
\bibliographystyle{rsc} 

\appendix
\onecolumn

\setcounter{figure}{0} 
\setcounter{section}{0}
\setcounter{table}{0}
\renewcommand{\thefigure}{S\arabic{figure}}
\renewcommand{\thetable}{S\arabic{table}}
\renewcommand{\thesection}{S\arabic{section}}

\clearpage
\section{Supplementary information for \textit{Towards best practices in low-dimensional semi-
supervised latent Bayesian optimization for the design
of antimicrobial peptides}}

\clearpage
\section{Support Vector Regression Oracle}
To serve as an oracle to rapidly evaluate our ``true'' objective function, we trained a Support Vector Regression (SVR) model on data compiled by Witten \& Witten (2019)\cite{witten_deep_2019} by using the scikit-learn SVR implementation \cite{scikit-learn}. This dataset contains peptide sequences and their associated Minimum Inhibitory Concentration (MIC), which we split into a training (4145 sequences) and a validation (2615 sequences) set. This dataset primarily contains sequences of lengths $<50$ amino acids.
\begin{figure}[H]
    \centering
    \includegraphics[width=\linewidth]{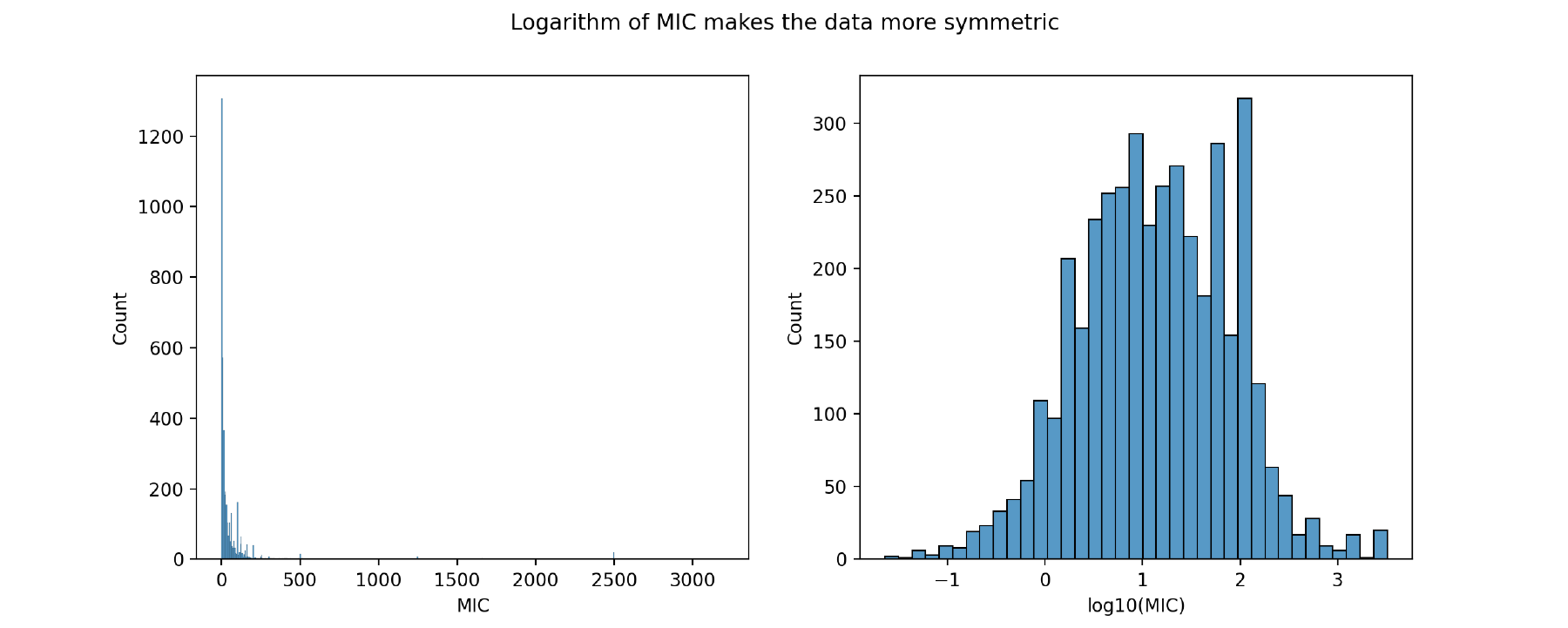}
    \caption{Distribution of MIC values in the oracle training dataset. Before applying a base-10 logarithm transformation the data is heavily skewed (left). After applying a base-10 logarithm transformation the data becomes more normally distributed (right).}
    \label{sifig:mic_dataset_before_after_log10}
\end{figure}
For the outputs of the oracle, we employed the base-10 logarithm of the MIC values $\log_{10}(\text{MIC})$ to ensure the values are distributed more symmetrically (Fig. \ref{sifig:mic_dataset_before_after_log10}). For the inputs to the oracle, we used easily-computed physico-chemical descriptors.  We used a previously-validated feature selection procedure\cite{lee_mapping_2016} to identify the most relevant subset from a larger set of possible descriptors. 

Computing all descriptors available in the ProPy3 Python package \cite{propy3_docs} resulted in 1529 features per sequence (Table \ref{sitbl:propy_descriptors}). We reduce the number of features necessary for the model via a two-step feature selection procedure. We perform an initial feature screening stage using the minimum Redundancy Maximum Relevance (mRMR) algorithm\cite{ding_minimum_2003}, in which one attempts to identify those features that have the least overlap that are the most relevant to the prediction of the output of interest. We defined \textit{redundancy} as the mutual information between pairs of physico-chemical features and \textit{relevancy} as the mutual information between features and predicted outputs (i.e. $\log_{10}(\text{MIC})$). 

\begin{table}[H]
\centering
\begin{tabular}{lll}
Abbreviation & Name                                                                               & Description                                                                                                                                                                                                                                              \\ \hline
AAC          & Amino Acid Composition                                                             & \begin{tabular}[c]{@{}l@{}}The fraction of a sequence composed of \\ a given amino acid.\end{tabular}                                                                                                                                                    \\
DPC          & Dipeptide Composition                                                              & \begin{tabular}[c]{@{}l@{}}The fraction of a sequence composed of \\ a given pair of amino acids.\end{tabular}                                                                                                                                           \\
MBauto       & \begin{tabular}[c]{@{}l@{}}Normalized Moreau-Broto \\ Autocorrelation\end{tabular} & \begin{tabular}[c]{@{}l@{}}Normalized autocorrelation of properties, \\ where property similarities are computed.\end{tabular}                                                                                                                           \\
Moranauto    & Moran autocorrelation                                                              & \begin{tabular}[c]{@{}l@{}}Autocorrelation of properties, where deviations \\ from the average are computed.\end{tabular}                                                                                                                                \\
Gearyauto    & Geary Autocorrelation                                                              & \begin{tabular}[c]{@{}l@{}}The squared-difference of a property at different\\ sequence locations is computed, and normalized \\ by the variance.\end{tabular}                                                                                           \\
CTD          & Composition, Transition, Distribution                                              & \begin{tabular}[c]{@{}l@{}}Amino acids are grouped based on 7 different\\ properties. Then for each property the \\ composition of the sequence, the number of \\ transition between groups, and the distribution \\ of groups is computed.\end{tabular} \\
SOCN         & Sequence Order Coupling Numbers                                                    & \begin{tabular}[c]{@{}l@{}}Quantifies the extent to which amino acids \\ are 'coupled' in a sequence based on amino \\ acid physicochemical distances.\end{tabular}                                                                                      \\
QSO          & Quasi Sequence Order                                                               & \begin{tabular}[c]{@{}l@{}}Composition and amino acid correlation, \\ computed using amino acid \\ physicochemical "distance" matrices.\end{tabular}                                                                                                     \\
PAAC         & Pseudo Amino Acid Composition                                                      & \begin{tabular}[c]{@{}l@{}}Computes the amino acid composition of a \\ sequence while maintaining a notion of how \\ correlated amino acids' hydrophilicity, \\ hydrophobicity, and mass are \\ throughout the sequence.\end{tabular}                   
\end{tabular}
\caption{Physicochemical descriptors computed using ProPy3 python package\cite{propy3_docs}. }
\label{sitbl:propy_descriptors}
\end{table}

After mRMR, we use LASSO \cite{tibshirani_regression_1996} regression to select a sparse subset of high-performing features, in a procedure analogous to that outlined in Bi \textit{et al.} (2003)\cite{bi_dimensionality_2003}\footnote[5]{Bi \textit{et al.} (2003) use L1-regularized Linear SVCs to perform feature selection.}, and applied in Lee \textit{et al.} (2016)\cite{lee_mapping_2016}. We first augment our relevant, non-redundant training data features with $10$ `dummy features', each sampled from a Normal distribution of mean zero and variance one. Then, we perform $T=10$ random $80$-$20$ splits of the entire training data into training and validation subsets for an initial hyperparameter search. For each split, we fit a LASSO regression model, performing a grid-search over the regularization hyperparameter $\alpha$ over values $[10^{-4},\; 10^{-3},\; 10^{-2},\; 10^{-1},\; 0.5,\; 1.0,\; 1.5,\; 2.5,\; 5.0,\; 7.5,\; 10,\; 50,\; 10^2]$ for the $\alpha$ value with lowest mean squared error (MSE) on the validation set. We keep the coefficient vector of the best model of each split. We average the magnitude of the coefficient of every dummy feature to determine a threshold coefficient magnitude. Any coefficient with magnitude less than or equal to the dummy feature threshold contributes to the prediction less than a random variable does, so its coefficient is set to zero (\textit{i.e.} that feature is removed). SI Table \ref{sitable:number_features_oracle} provides a summary of the number of features after each stage. 

\begin{table}[H]
\centering
\begin{tabular}{llrl}
Rank & Name                       & \multicolumn{1}{l}{\begin{tabular}[c]{@{}l@{}}LASSO\\ Weight\end{tabular}} & Brief Description                                                                                                                                        \\ \hline
1    & tausw2                     & -0.19                                                                      & Quasi Sequence Order coupling number                                                                                                               \\
2    & PolarizabilityC3           & -0.14                                                                      & Polarizability Composition                                                                                                                         \\
3    & ChargeT12                  & -0.13                                                                      & Number of charge transition between charged and neutral AAs                                                                                        \\
4    & NormalizedVDWVC2           & -0.09                                                                      & van der waals volume                                                                                                                               \\
5    & PAAC20                     & 0.08                                                                       & \begin{tabular}[c]{@{}l@{}}Type I Pseudo amino acid composition descriptors; \\ it uses hydrophobicity, hydrophilicity, residue mass.\end{tabular} \\
6    & QSOSW39                    & -0.07                                                                      & Quasi Sequence Order using Schneider-Wrede distance matrix                                                                                         \\
7    & PAAC21                     & 0.06                                                                       & Type I Pseudo amino acid composition descriptors                                                                                                   \\
8    & MoreauBrotoAuto\_Steric2   & 0.06                                                                       & Moreau-Broto Autocorrelation                                                                                                                       \\
9    & ChargeD2100                & 0.05                                                                       & Charge distribution group 2 100\%                                                                                                                  \\
10   & MoranAuto\_Hydrophobicity5 & 0.05                                                                       & Moran Autocorrelation                                                                                                                             
\end{tabular}
\label{sitbl:top-10-features}
\caption{The top-10 features selected after mRMR and LASSO. Note that this is a subset of the 149 features we kept after our feature selection procedure.}
\end{table}

Next, we determined hyperparameters for a nonlinear SVR with Radial Basis Function kernel. Using 5-fold cross-validation, we performed a grid search for hyperparameters $C=10^\beta$ and $\epsilon=10^\beta$ over $\beta$ values running from $-2$ to $2$ in 45 equidistant steps. Once hyperparameters were selected, we fit a final nonlinear SVR on the full training set, yielding a mean-squared error of $0.328$ on the test set while a simpler linear regressor on the same features yielded a less-optimal, higher error of $0.407$ and a predictor using only the mean of the training data set yielded a mean-squared error of $0.602$. This final SVR was used to predict $\log_{10}(\text{MIC})$ to obtain our `true' MIC value; throughout we use this as our 'oracle'.

\begin{table}[H]
\centering
\begin{tabular}{ll}
Stage      & No. of Features \\
\hline
Initial    & 1529            \\
Post-mRMR  & 248             \\
Post-LASSO & 149            
\end{tabular}
\caption{Number of features at each stage of feature selection when building the SVR oracle.}
\label{sitable:number_features_oracle}
\end{table}

\clearpage
\section{Additional TransVAE information and analysis}

\clearpage
\subsection{Additional model details}
\begin{figure}[H]
    \centering
    \includegraphics[width=\linewidth]{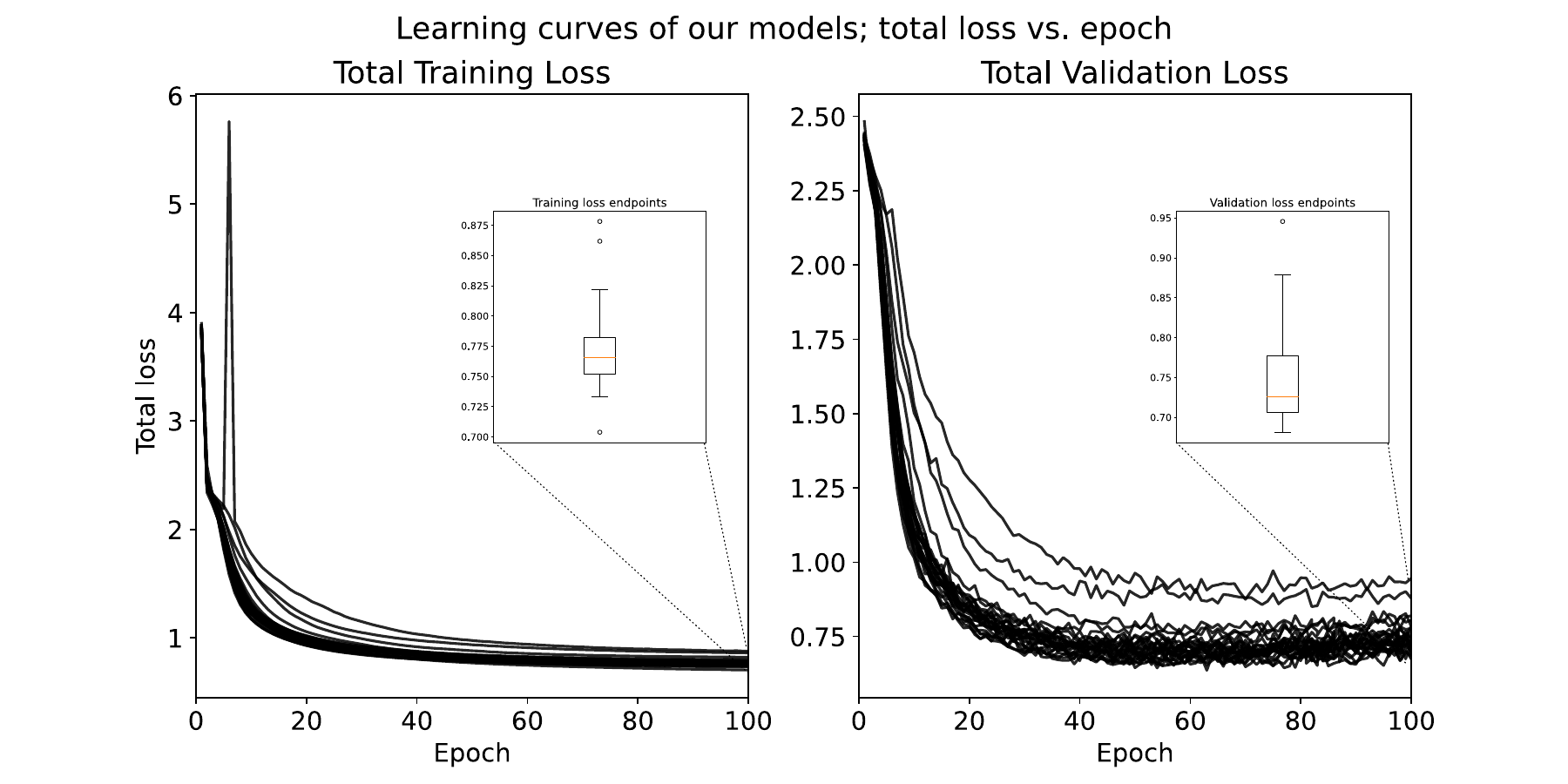}
    \caption{Learning curves for all models trained. Inset boxplots of total loss values at epoch 100. Total loss is the sum of reconstruction loss, KL Divergence, and property prediction loss. At epoch 100, all models have total training loss $<0.9$ and total validation loss $<0.95$. The majority of models have even lower loss values that are all relatively near each other. The validation loss is frequently lower than the training loss due to dropout layers being turned off during inference on the validation set.}
    \label{fig:si:learning_curves}
\end{figure}

We trained variational autoencoders (VAEs) with transformer architectures based on \citet{renaud_latent_2023}. Model architecture was chosen based on work done in \citet{renaud_latent_2023}, with specific hyper-parameter values and descriptions listed in Table \ref{tbl:hyperparameters}. To optimize the weights of the neural networks, we used the Noam optimizer\cite{vaswani_attention_2023}, a variant of the Adam optimizer \cite{kingma_adam_2017} with a varying learning rate. The output dimensions of the property predictor module varied over none, 1, 2, 3, depending on the number of organizing properties. All models were trained to 100 epochs, with the total training loss $<0.9$ and validation losses $<0.95$, with most models have a total validation loss $<0.8$ (Fig. \ref{fig:si:learning_curves}).
\begin{table}[H]
\centering
\begin{tabular}{l|l|l}
Hyperparameter& Value & Description \\ 
\hline
$d_{\text{latent}}$ & 64 & Dimensionality of latent space \\ 
$d_{\text{model}}$ & 512 & Dimensionality of model \\ 
$N_{\text{block}}$ & 3 & Number of Encoder/Decoder blocks \\ 
$d_{\text{ff}}$ & 256 & Dimensionality of feedforward layers \\ 
$h$ & 4 & Number of attention heads per multi-head attention layer \\
$d_{\text{pp}}$ & 64 & Dimensionality of property predictor layers \\ 
$\text{depth}_{\text{pp}}$ & 2 & Number of property predictor layers \\ 
$d_{\text{pp,out}}$ & 1,2,3 & Output dimensionality of property predictor \\ 
$\beta_{\text{pp},0}$ & 0 & Initial value for property predictor annealing coefficient \\ 
$\beta_{\text{pp}}$ & 1 & Final value for property predictor annealing coefficient \\ 
$\beta_{\text{KL},0}$ & 0 & Initial value for KL annealing coefficient \\ 
$\beta_{\text{KL}}$ & 0.05 & Final value for KL annealing coefficient \\ 
$N_\text{batch}$ & 256 & Batch size during training\\ 
$\beta_1$ & 0.9 & Adam optimizer's $\beta_1$ \\ 
$\beta_2$ & 0.98 & Adam optimizer's $\beta_2$ \\ 
lr & $(d_{model})^{-0.5} \cdot \min\Big(\text{step}^{-0.5}, \text{step}\cdot (n_{\text{warmup}})^{-1.5}\Big)$ & Learning rate set adaptively using the Noam optimizer  \\  
$n_{\text{warmup}}$ & 10000 & Number of warmup steps for the Noam optimizer \\
$p_\text{dropout}$ & 0.1 & Dropout rate \\ \hline
\end{tabular}
\caption{Hyper-parameters for the model architecture and training. The TransVAE consists of an encoder and decoder module, each consisting of $N_{\text{block}}$ blocks of multi-head self-attention feeding into feedforward layers. }
\label{tbl:hyperparameters}
\end{table}

\begin{figure}[H]
    \centering
    \includegraphics[width=\linewidth]{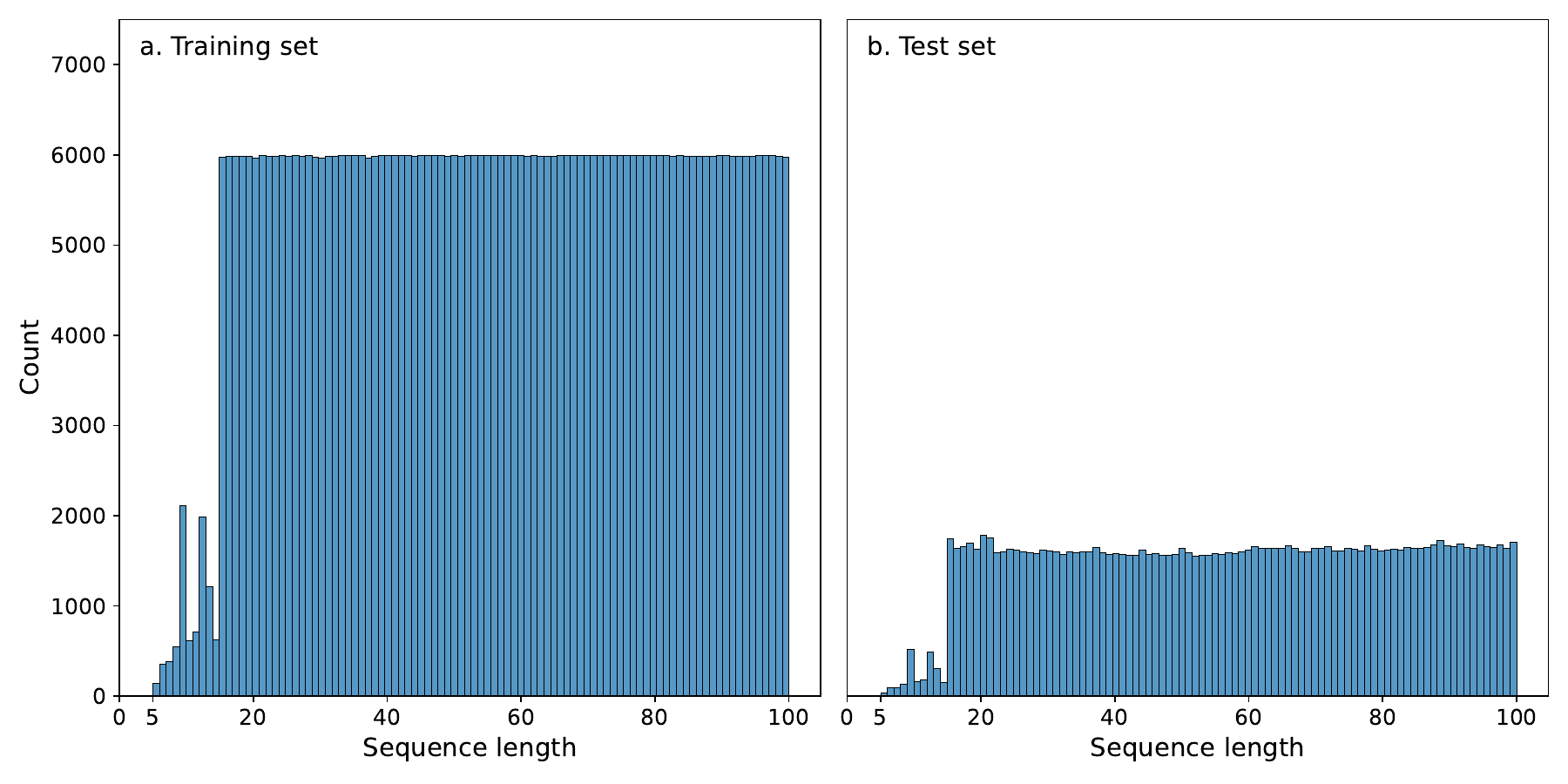}
    \caption{Distribution of peptide sequence lengths in our dataset.}
\label{sifig:histogram_of_dataset_sequence_lengths}
\end{figure}

\subsection{Summary of models}

\clearpage
\subsection{Additional PCA projection analysis}
\begin{figure*}[h!]
    \centering
    \includegraphics[width=\linewidth]{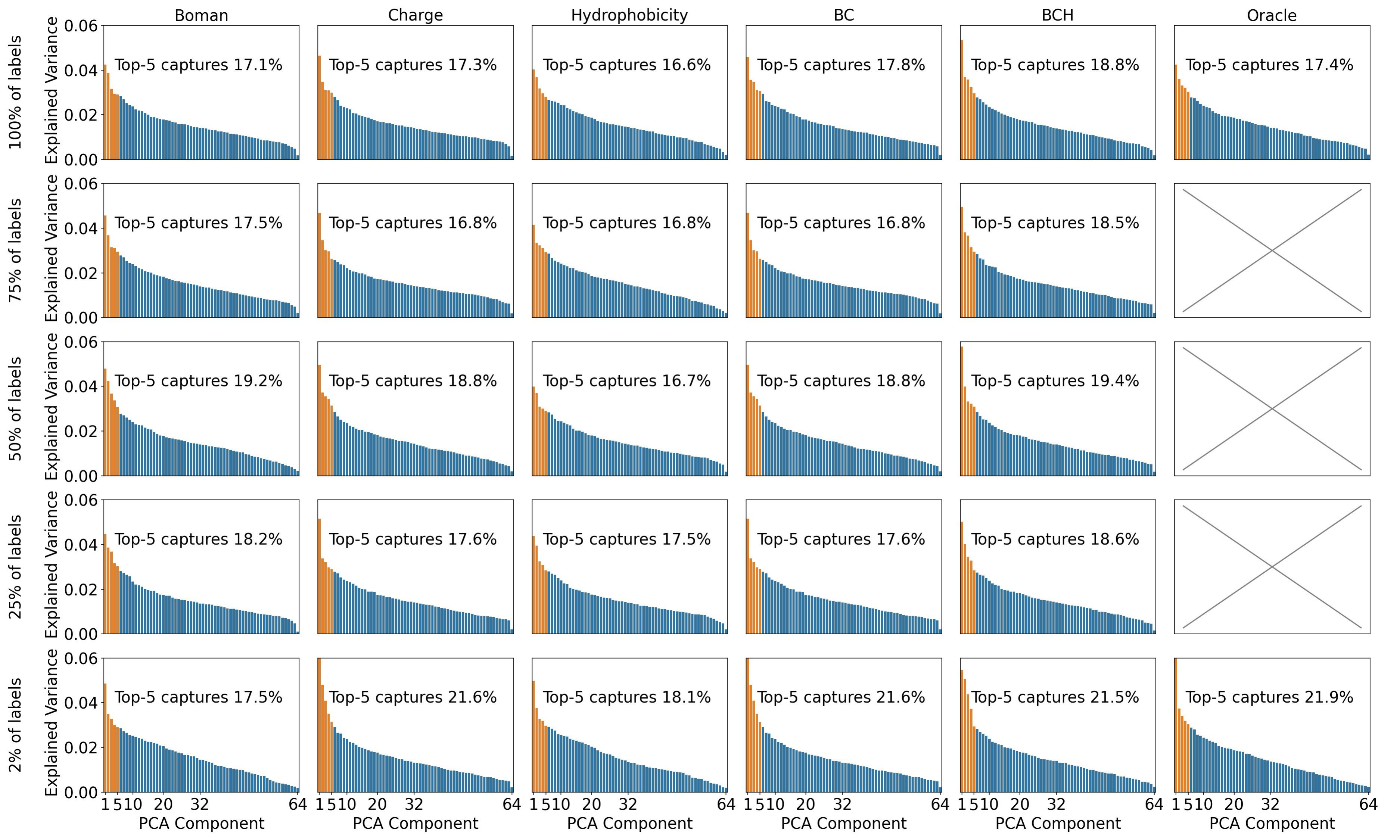}
    \caption{Principal components explained variance ratios for each TransVAE model. The cumulative explained variance, as a percentage, of the top 5 PCs is reported in its corresponding panel. Top 5 PCs are coloured orange.}
    \label{fig::si:explained_var_all_PCs}
\end{figure*}

Principal Component Analysis (PCA) determines a set of ordered basis vectors for a set of data points, where each basis vector in the ordering captures a decreasing amount of variance in the original cloud of data points. In the main text, we frequently use only the top five principal components as a lower dimensional representation of a full latent space. Here we provide the full distribution of principal components (Fig. \ref{fig::si:explained_var_all_PCs}). We observe that the top five principal components capture between 17\% and 22\% of the variance in the VAE training set. 

Along with explained variance, we calculate the correlation between each of the top five principal components, and the organizing property (or each individual property in the case of multi-property organization), allowing us to best visualize the property-induced organization of the latent space. In Fig. \ref{fig:pca_bch_train}, we visualized the extent to which the latent space is organized by Boman index, charge, and hydrophobicity, when the underlying TransVAE model was being organized by all three simultaneously. Table \ref{tbl:si:PCs_in5x3} provides the PC pair used for visualization. Charge is frequently most organized along (most correlated with) PCs separate from Boman and Hydrophobicity. Boman and Hydrophobicity frequently have overlapping, but not always the same, PC pair; while Boman and hydrophobicity have similar trends, they are not quite the same, this can suggest that the 2D projection, and the linear correlation used to select PCs, is hiding differences in organization. 
\begin{table}[H]
\centering
\begin{tabular}{rlll}
\multicolumn{1}{l}{Label Percentage} & Boman & Charge & Hydrophobicity \\ \hline
100\%                                & (2,3) & (5,2)  & (2,3)          \\
75\%                                 & (2,3) & (5,3)  & (2,4)          \\
50\%                                 & (2,3) & (4,5)  & (2,1)          \\
25\%                                 & (2,1) & (3,4)  & (2,1)          \\
2\%                                  & (1,3) & (3,1)  & (1,3)         
\end{tabular}
\caption{Principal component pairs used for visualization in Fig. \ref{fig:pca_bch_train}. The location of each pair corresponds to their subpanel in Fig. \ref{fig:pca_bch_train}.}
\label{tbl:si:PCs_in5x3}
\end{table}

\clearpage
\subsection{Relation between organizing properties and predictive oracle}
\label{sect:relationship-organizing-props-oracle}

\begin{figure*}[h!]
    \centering
    \includegraphics[width=\linewidth]{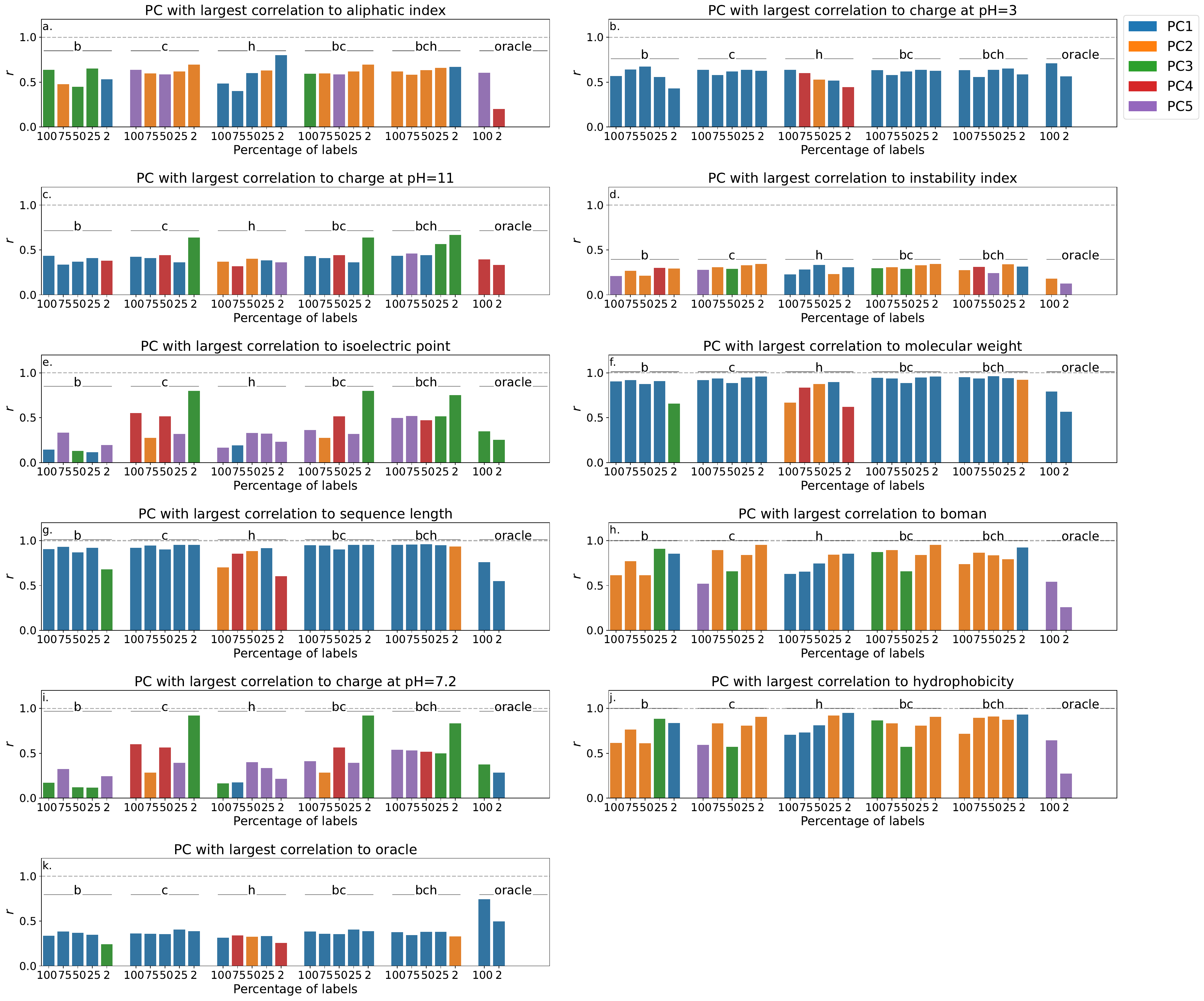}
    \caption{Top PCs and their correlation scores with physico-chemical properties. (a-k) Pearson correlation coefficients ($r$) between PCs and (a) aliphatic index, (b) charge at pH$=3$, (c) charge at pH$=11$, (d) instability index, (e) isoelectric point, (f) molecular weight, (g) sequence length, (h) Boman index, (i) charge at pH$=7.2$, (j) hydrophobicity, (k) predicted $\log_{10}(\text{MIC})$ values from the SVR oracle. Bars grouped by model, and shown in order of decreasing percentage of property labels available during training. Note: Boman index and hydrophobicity are length-normalized quantities, while charge is not.}
    \label{fig:si:bridge_variables}
\end{figure*}

We investigated in some detail the correlations between the principal components and the ``bridge variables,'' that is different physicochemical properties, including organizing variables. We observed that sequence length was most correlated with the first principal component across 21/27 of the models (SI Fig \ref{fig:si:bridge_variables}g), with the second or third principal component being correlated with the property used during training to organize the latent space (SI Fig. \ref{fig:si:bridge_variables}h for Boman index; SI Fig. \ref{fig:si:bridge_variables}i for charge; SI Fig. \ref{fig:si:bridge_variables}j for hydrophobicity; SI Fig. \ref{fig:si:bridge_variables}k for oracle). Although we may desire the first principal component to be most correlated with the organizing property in most cases, we note that the top five principal components have similar amounts of explained variance, and that they capture on the order of $20\%$ of the total variance (SI Fig. \ref{fig::si:explained_var_all_PCs}). Additionally, previous work found sequence length can be an easy-to-learn pattern\cite{renaud_latent_2023}. Patterns that are straightforward to recognize may become unexpectedly strongly correlated with a PC direction.

To check whether searches through spaces organized by different properties performed differently due to certain properties providing stronger signal about the objective value (the output of the SVR), we computed two different metrics. First, we computed the correlation between organizing properties (Boman Index, Charge, and Hydrophobicity) and the top-10 features of the SVR. Because correlation is biased towards monotonic trends, we also estimate the mutual information between the organizing properties and the features of the SVR. 

\begin{table}[H]
\centering
\footnotesize
\tabcolsep=0.11cm
\begin{tabular}{llll}
\textbf{Input Feature} & \textbf{Boman Index} & \textbf{Charge} & \textbf{Hydrophobicity} \\ \hline
tausw2           & 0.27 (0.0075)  & 0.022 (0.011)  & -0.27 (0.0058)  \\
PolarizabilityC3 & 0.34 (0.0099)  & 0.50 (0.0038)  & -0.34 (0.0082) \\
ChargeT12        & 0.51 (0.0062)  & 0.60 (0.0038)  & -0.45 (0.0091) \\
NormalizedVDWVC2 & -0.28 (0.0020)  & -0.29 (0.0012)  & 0.33 (0.014) \\
PAAC20           & -0.34 (0.0069) & -0.091 (0.0041) & 0.43 (0.013)   \\
QSOSW39          & 0.037 (0.0084)  & -0.0041 (0.0063) & -0.027 (0.019) \\
PAAC21           & 0.44 (0.009)   & 0.21 (0.0036)  & -0.36 (0.0098)\\
MoreauBrotoAuto\_Steric2 & -0.0021 (0.0043) & 0.014 (0.0069) & -0.025 (0.013) \\
ChargeD2100      & -0.20 (0.017) & -0.038 (0.011) & 0.17 (0.012) \\
MoranAuto\_Hydrophobicity5 & -0.047 (0.015) & 0.015 (0.012) & 0.045 (0.0074) \\ \hline
Num. with magnitude $<0.2$ & 3 & 6 & 3
\end{tabular}
\caption{Average (standard deviation) Pearson correlation coefficient between the organizing property (column) and each of the top-10 input features (row) to the SVR. Properties and features are computed on the training set. For each property, the Pearson correlation was computed across five independent subsamples of 10,000 points; the reported values are the mean and standard deviation across these subsamples. Threshold of $0.2$ corresponds to an correlation magnitude of very weak.}
\label{tbl:PCC-props-with-svr-inputs}
\end{table}

Compared to Boman index and hydrophobicity, charge had very weak correlations with the most features (6/10, Table \ref{tbl:PCC-props-with-svr-inputs}), but strong mutual information scores. Compared to hydrophobicity, Boman index had similar correlation values (frequently with switched signs), but weaker mutual information scores. Hydrophobicity tended to be most informative about the top-10 features inputted into the SVR. While with the outputted predicted log10(MIC) values, charge is both more strongly correlated and has a higher mutual information score than Boman Index — the caveat being that charge has a weak correlation, and a somewhat low mutual information score (Table \ref{tbl:props-with-svr-outputs}).

\begin{table}[H]
\centering
\small
\tabcolsep=0.11cm
\begin{tabular}{lrrr}
\textbf{Input Feature}     & \textbf{Boman Index} & \textbf{Charge} & \textbf{Hydrophobicity} \\ \hline
tausw2                     & 0.13  & 0.22 & 0.12 \\
PolarizabilityC3           & 0.11  & 0.37 & 0.44 \\
ChargeT12                  & 0.22  & 0.59 & 0.35 \\
NormalizedVDWVC2           & 0.083 & 0.19 & 0.37 \\
PAAC20                     & 0.088 & 0.065& 0.15 \\
QSOSW39                    & 0.043 & 0.15 & 0.11 \\
PAAC21                     & 0.15  & 0.076& 0.11 \\
MoreauBrotoAuto\_Steric2   & 0.029 & 0.025& 0.025\\
ChargeD2100                & 0.033 & 0.087& 0.13 \\
MoranAuto\_Hydrophobicity5 & 0.018 & 0.028& 0.014\\ \hline
Num. with highest value    & 2 & 4 & 4           \\ 
Average of MI scores       & 0.26 & 0.53 & 0.53
\end{tabular}
\caption{Mutual information between organizing properties and the top-10 features of the SVR.}
\label{tbl:MI-props-with-svr-inputs}
\end{table}

\begin{table}[H]
\centering
\begin{tabular}{lrrr}
 & \textbf{Boman Index} & \textbf{Charge} & \textbf{Hydrophobicity} \\\hline
PCC & -0.15 & -0.30 & -0.14 \\
Mutual Information & 0.027 & 0.096 & 0.021                                      
\end{tabular}
\caption{Pearson correlation coefficient and Mutual Information between organizing properties and predicted log10(MIC) over the training set. The Pearson correlation was computed across five independent subsamples of 10,000 points; the reported values are the mean and standard deviation across these subsamples. Charge is calculated at pH$=7.2$.}
\label{tbl:props-with-svr-outputs}
\end{table}

\clearpage
\subsection{Energy consumption}
We estimate the total energy consumption of training the deep learning models to 100 epochs. The system we performed training on was a NVIDIA V100 Volta (32G HBM2 memory) through the Digital Research Alliance of Canada's Cedar cluster. From comet-ml, we observed an average power usage of $\approx 275 W$ throughout a given training job \cite{comet_ml_tool}. To train a given model to 100 epochs, approximately $24$hrs were required, yielding $275\cdot24/1000 = 6.6$ kWh per 100 epochs. For the main text, we trained 27 models to 100 epochs, giving an approximate total energy usage of $178.2$kWh.

The Hyundai Ioniq 6 is a 2022 battery electric sedan. Its long-range battery capacity is $77.4$kWh, corresponding to an estimated range of $614$km \cite{hyundai_ioniq_6_specs}, giving $\approx 0.126$kWh/km. Comparing to our total energy usage, we obtain an approximate car range using energy from 100 epochs as $\approx52.36$km. Therefore, the total approximate range of a Hyundai Ioniq 6 using the energy from training our main text models is $1413.63$km, or we could fully recharge $2.3$ Hyundai Ioniq 6 sedans.   

\begin{table}[H]
\centering
\begin{tabular}{l|l}
System & NVIDIA V100 Volta \\ 
\hline
Average GPU energy utilization (Watts) & 275 \\ 
Approx. hours for 100 epochs & 24 \\ 
kWh per 100 epochs & 6.6 \\ 
Hyundai Ioniq 6 battery capacity & 77.4 kWh \\ 
Hyundai Ioniq 6 estimated range & 614 km \\ 
Energy per km & 0.1261 kWh/km \\ 
Approx. car range from 100 epochs & 52.36 km \\ 
Number of 100 epoch runs & 27 \\ 
Approximate total energy used & 178.2 kWh \\ 
Approximate equivalent car range & 1413.63 km \\ \hline
\end{tabular}
\label{sitbl:energy-usage}
\caption{Energy usage of deep learning training on Cedar and its equivalent in electric car distance (Hyundai Ioniq 6).}
\end{table}

\clearpage
\section{Additional BayesOpt Results}

\subsection{At low numbers, the number of initial data points has little effect.}
\label{sect:vary-number-initial-points}
In the case of antimicrobial peptides, and peptides with hemolytic effect, the number associated with high-quality verification of the mechanism of action — through experimental or simulation methods — is rather low. To establish whether this will limit optimization in our models, we test the effect of varying the number of points with which we initialize the BayesOpt Gaussian Process Regressor. We ensure that each smaller initialization set is contained in the previous, larger initialization set, and that the peptide with strongest predicted activity is maintained throughout all of the sets to ensure that different initializations begin at the same objective value. That is, the best peptide from the 100 data point initialization set is held throughout each subset of it, and we randomly select the remaining points; \textit{e.g.} to construct the 10 data point subset, we begin with the best peptide from the 100, then without replacement randomly select the other 9 peptides from the remaining 99. 

\begin{figure}[H]
    \centering
    \includegraphics[width=\linewidth]{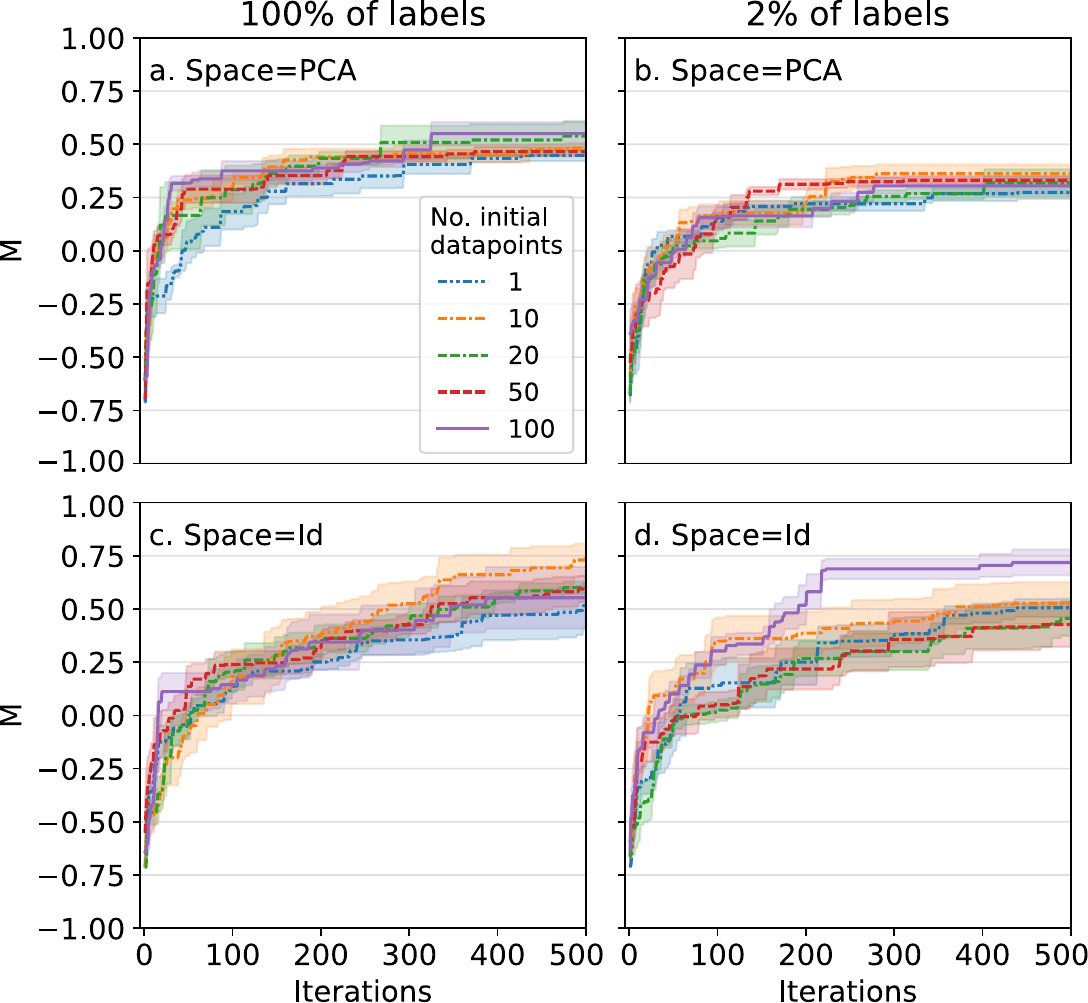}
    \caption{Bayesian optimization performance as the number of initialization points is varied. Trajectories correspond to the best value encountered up to a given iteration. Error bars correspond to the standard error over five different random initializations. We tested different numbers of initial datapoints in the initial dataset: 1 (blue, dash-dot-dotted), 10 (orange, dash-dotted), 20 (green, dash-dash-dotted), 50 (red, dashed), and 100 (purple, solid). Each initialization is run in a space trained with 100\% access to property labels (a,c) and 2\% (b,d), and searching through the full 64-dimensional latent space (c,d) and 5-dimensional PCA reduced space (a,b). The organizing property is predicted-log10mic, which is also the (negative) of the objective.}
    \label{fig:boloops:exploitation:plog10mic-vary-init-points}
\end{figure}

When varying the number of initial data points in this way, we observe very little change in the optimization results. In the predicted-$\log_{10}$(MIC)–organized VAE with access to 100\% of the property labels during training, varying the number of initial data points does not significantly impact the optimization performance, regardless of whether optimization is done in the VAE's latent space or in a PCA'ed projection of it (Fig \ref{fig:boloops:exploitation:plog10mic-vary-init-points}ac). While it appears that one initial data point results in a slight slowdown (Figure \ref{fig:boloops:exploitation:plog10mic-vary-init-points}a) within the first 100 iterations, by the end of 500 iterations it reaches the same objective value as the other optimization runs. In the case of having access to only 2\% of the property labels during VAE training, having 100 initial data points can provide a boost if we optimize in the 64-dimensional latent space (Figure \ref{fig:boloops:exploitation:plog10mic-vary-init-points}d). Ultimately, we observe little change to optimization performance when varying the number of initial data points below $100$, implying that in the case when limited data is extant and generating initial data is expensive, few data points may be used without significant loss of optimization performance.

\begin{figure}[H]
    \centering
    \includegraphics[width=\linewidth]{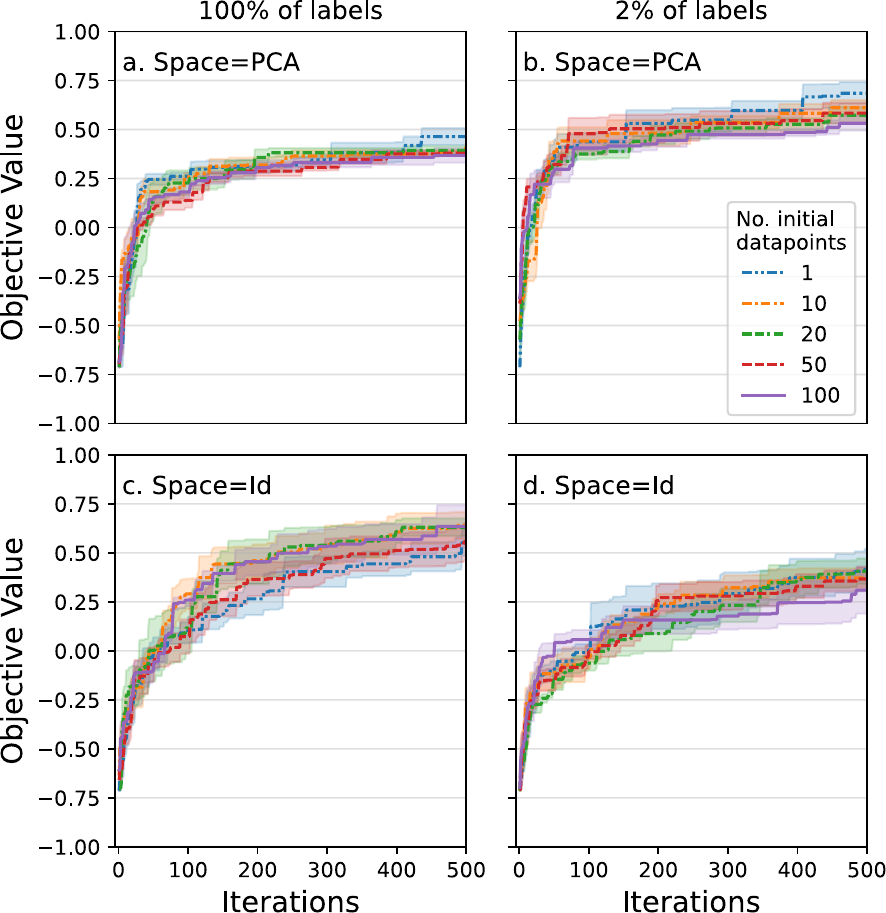}
    \caption{BayesOpt optimization performance over BCH-organized latent space while varying the number of initial points used to initialize the GPR.}
    \label{sifig:boloops:exploitation:bch-vary-init-points}
\end{figure}

We further perform BayesOpt in the BCH-organized latent spaces, both in the full 64-dimensional space and in the PCA reduced space, with the same sweep in number of initial data points. When optimizing in this differently-organized VAE, we observe little effect on the optimization performance due to the different number of initial points (SI Fig. \ref{sifig:boloops:exploitation:bch-vary-init-points}).

\clearpage
\subsection{Vary percentage of labels for given organizing properties}
\begin{figure}[h!]
    \centering
    \includegraphics[width=\linewidth]{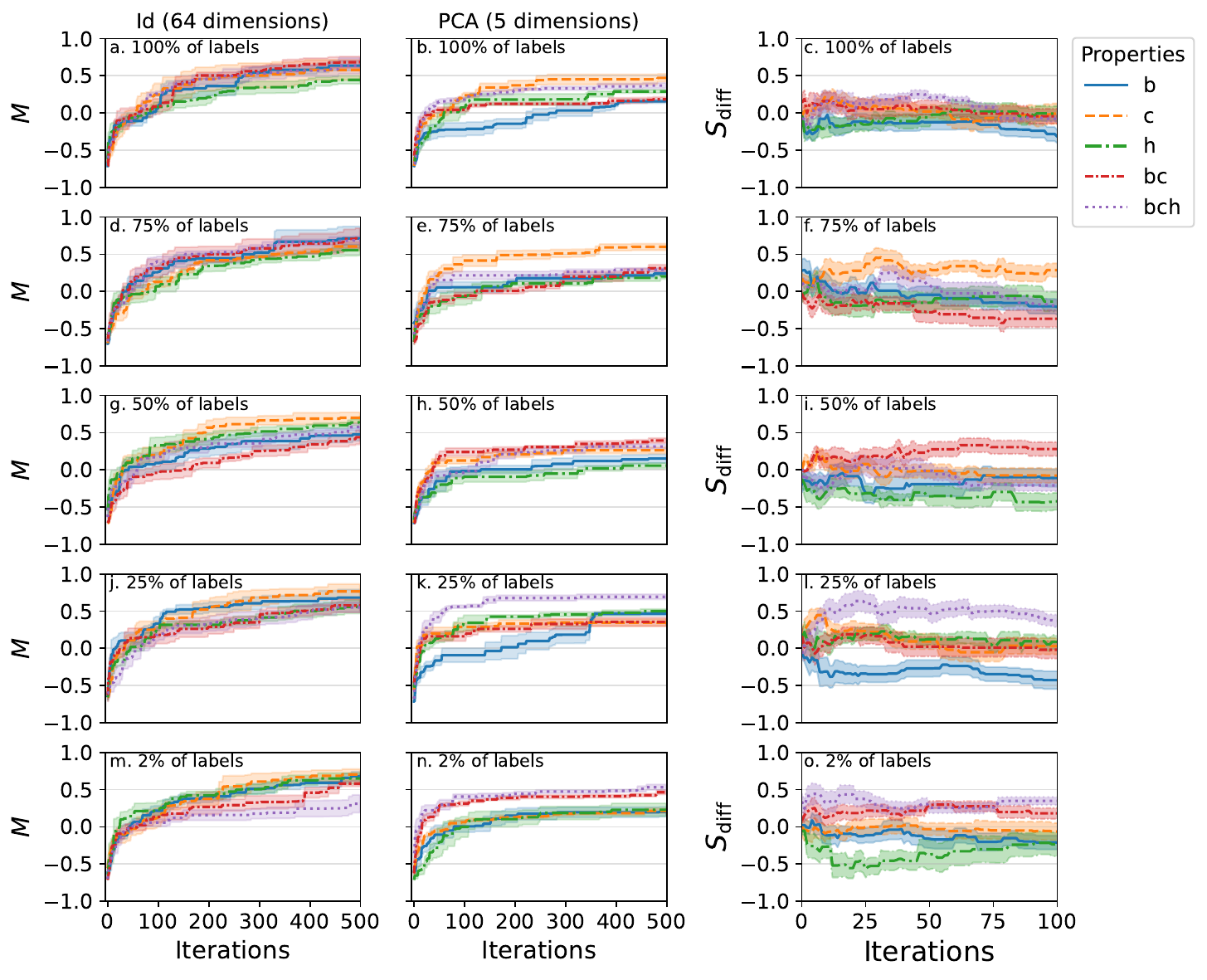}
    \caption{\textbf{Bayesian optimization performance in different percentage of labels available while varying the VAE-organizing properties.} Optimization was initialized with 100 data points. Error bars correspond to standard error over 5 different initializations. (third column, cfili) Difference in Bayesian optimization results $S_{\text{diff}}$, the difference in best score at each iteration. {Error bars correspond to the standard error in the difference of scores, with the variance computed through $\sigma^2(S_{\text{diff}}) = \sigma^2( M_{\text{Id}}
    ) + \sigma^2( M_{\text{PCA}} ) - 2\sigma(M_{\text{Id}}, M_{\text{PCA}} ) $}. Note that the the x-axis range is only up to iteration 100 to more clearly see early optimization behaviour. We observe that searching through the PCA'ed latent space offers an advantage in earlier iterations compared to searching through the 64-dimensional latent space. }
    \label{fig:boloops:exploitation:foreach-perc-vary-prop}
\end{figure}

\subsection{Vary organizing properties for a given percentage of labels}
\begin{figure}[H]
    \centering
    \includegraphics[width=0.75\linewidth]{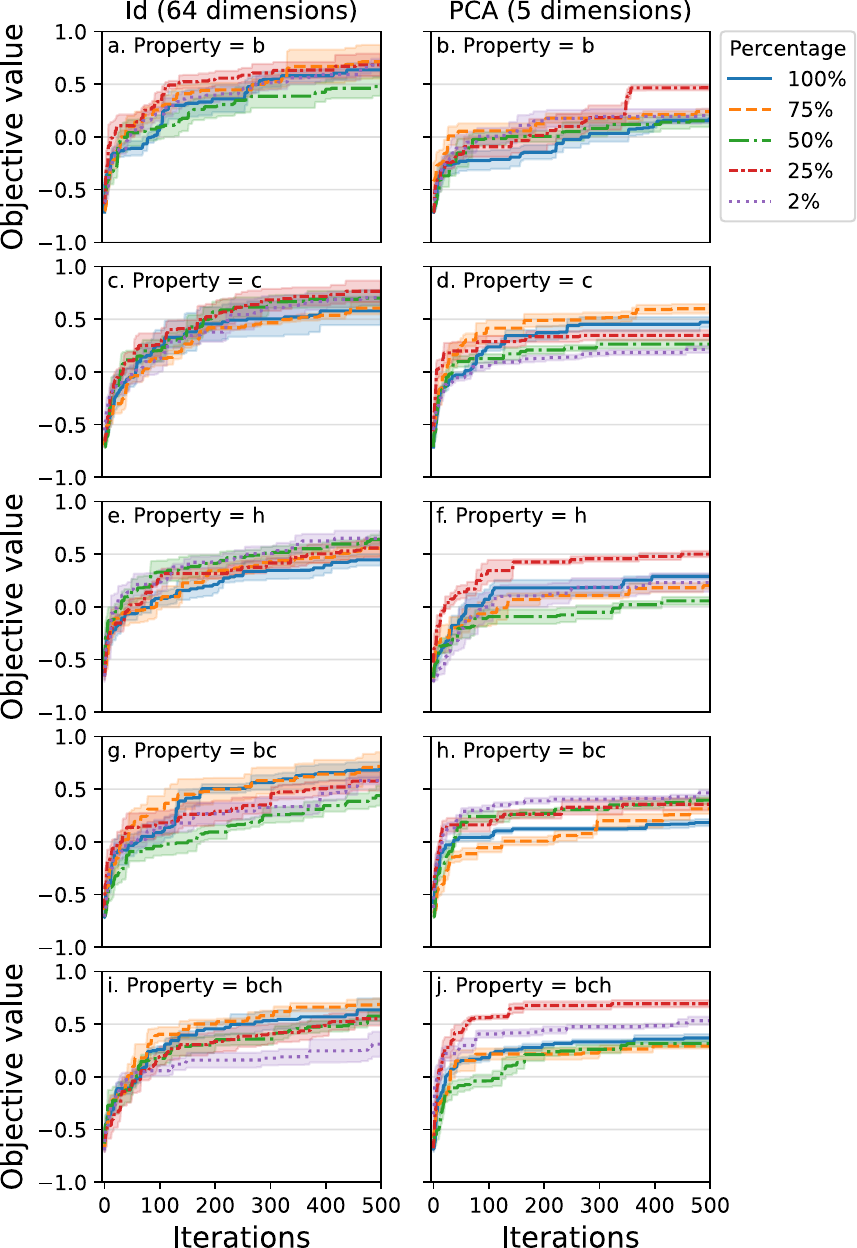}
    \caption{\textbf{Bayesian optimization performance in differently-organized spaces with varying percentage of property labels.}}
    \label{fig:placeholder}
\end{figure}

\subsection{Vary number of initial data points}

\clearpage
\subsection{Final values at iterations 50, 100, 500}
\begin{figure*}[h!]
    \centering
    \includegraphics[scale=0.3]{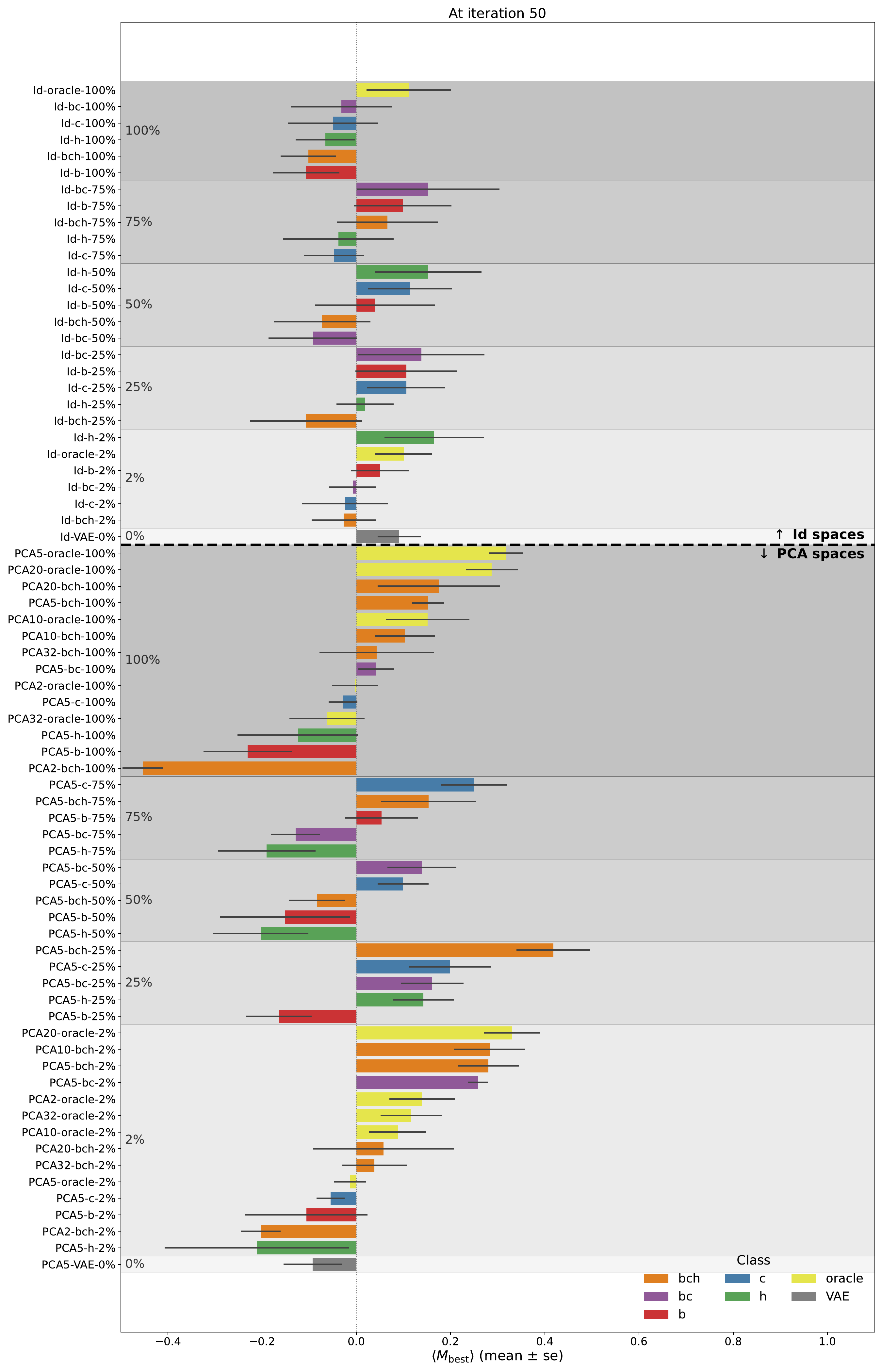}
    \caption{Average best objective scores found at iteration $50$. Error bars correspond to the standard error of $M_{\text{best}}$ over five BayesOpt runs. Black dashed line separates those runs performed in the high-dimensional latent space (above the dashed line), and those runs performed in a PCA projection of the latent space (below the dashed line). Shaded regions denote the label percentage available at training time, with darker regions corresponding to a higher label percentage. Values are sorted within groups from largest to smallest. The full BayesOpt trajectories are depicted in SI Fig. \ref{fig:boloops:exploitation:foreach-perc-vary-prop}.}
    \label{sifig:final_values_barplot_it50}
\end{figure*}

\begin{figure*}[h!]
    \centering
    \includegraphics[width=0.8\linewidth]{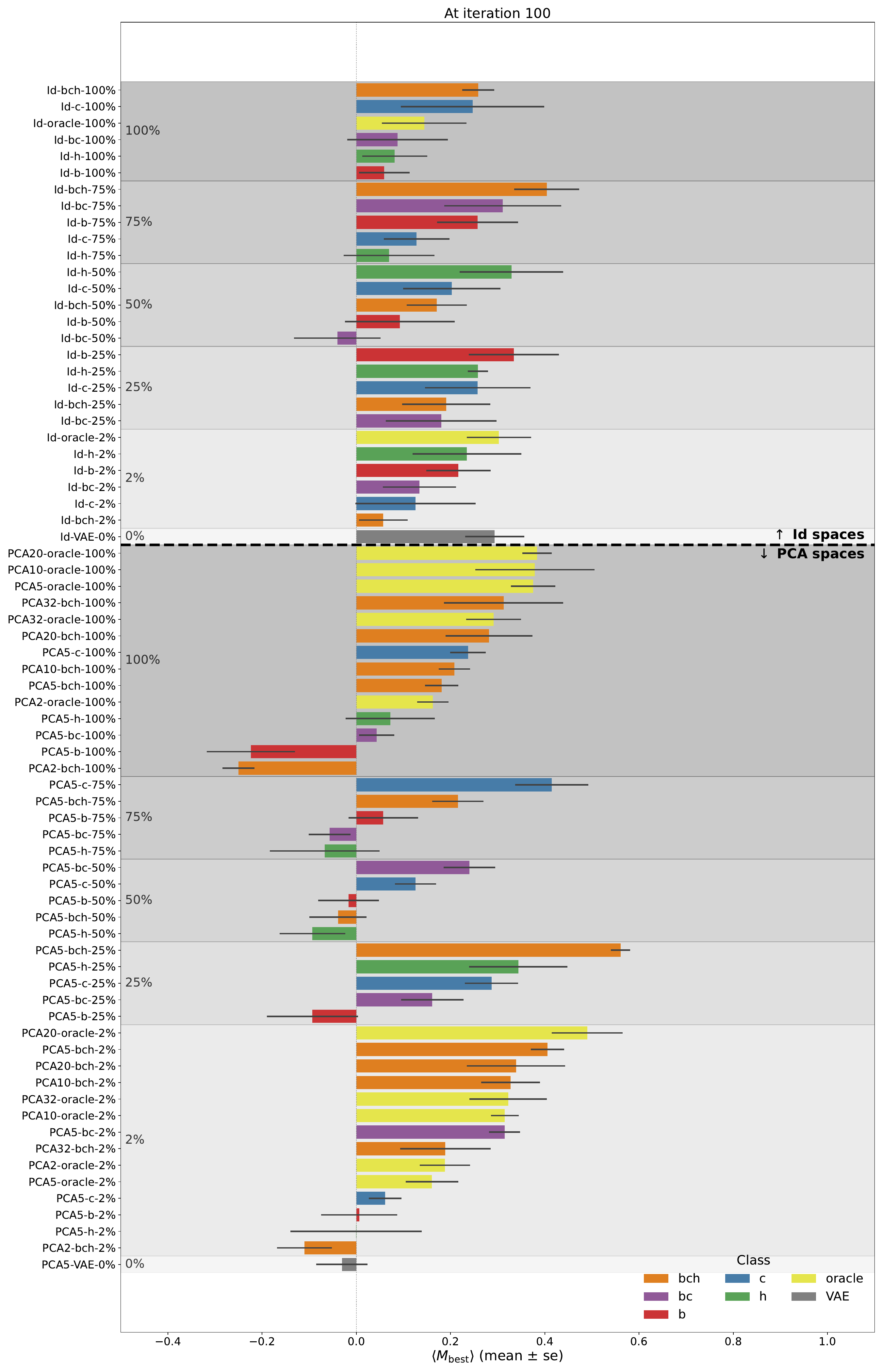}
    \caption{Average best objective scores found at iteration $100$. Error bars correspond to the standard error of $M_{\text{best}}$ over five BayesOpt runs. Black dashed line separates those runs performed in the high-dimensional latent space (above the dashed line), and those runs performed in a PCA projection of the latent space (below the dashed line). Shaded regions denote the label percentage available at training time, with darker regions corresponding to a higher label percentage. Values are sorted within groups from largest to smallest. The full BayesOpt trajectories are depicted in SI Fig. \ref{fig:boloops:exploitation:foreach-perc-vary-prop}.}
    \label{fig:final_values_barplot_it100}
\end{figure*}

\begin{figure*}[h!]
    \centering
    \includegraphics[width=0.8\linewidth]{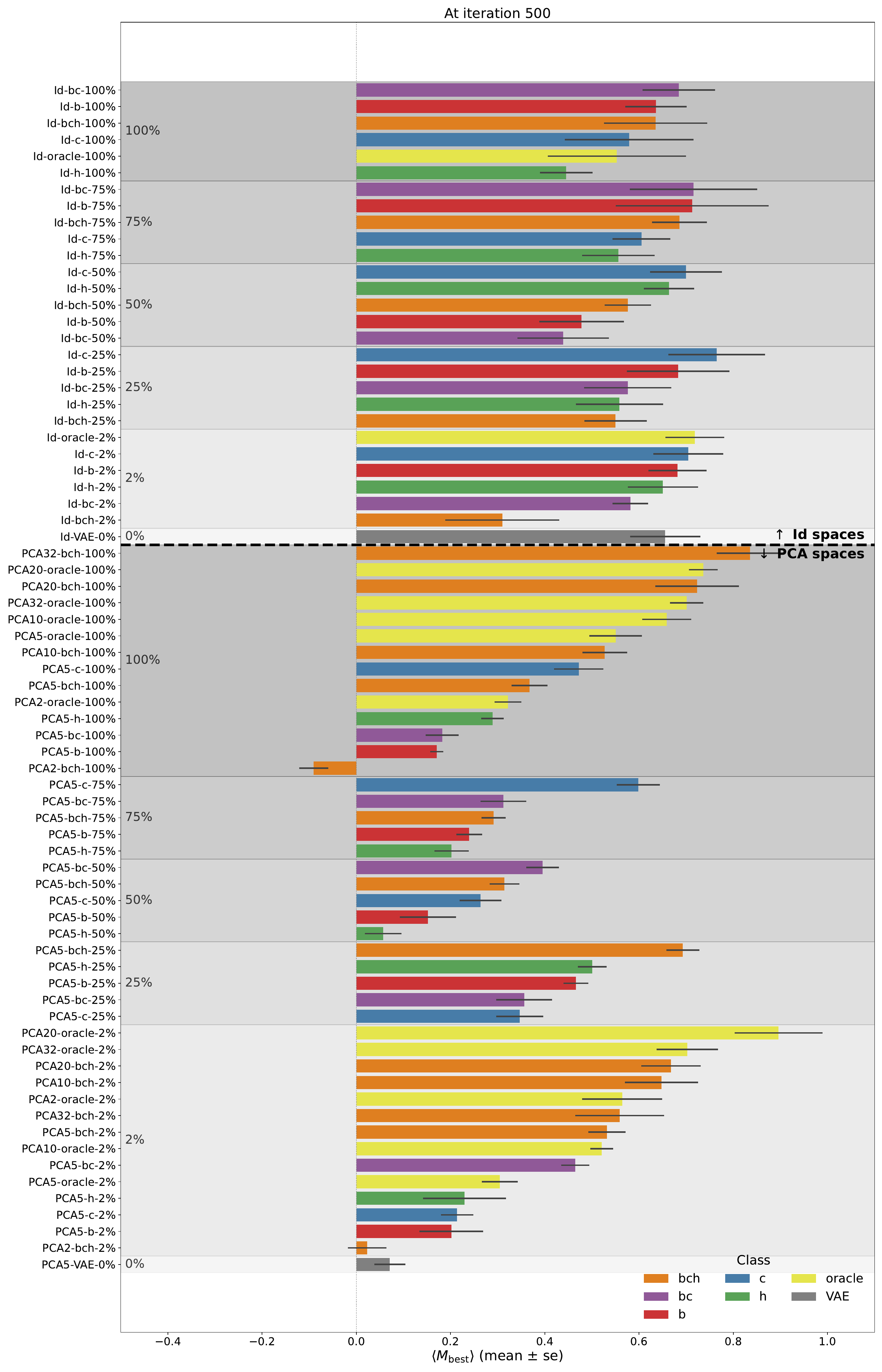}
    \caption{Average best objective scores found at iteration $500$. Error bars correspond to the standard error of $M_{\text{best}}$ over five BayesOpt runs. Black dashed line separates those runs performed in the high-dimensional latent space (above the dashed line), and those runs performed in a PCA projection of the latent space (below the dashed line). Shaded regions denote the label percentage available at training time, with darker regions corresponding to a higher label percentage. Values are sorted within groups from largest to smallest. The full BayesOpt trajectories are depicted in SI Fig. \ref{fig:boloops:exploitation:foreach-perc-vary-prop}.}
    \label{fig:final_values_barplot_it500}
\end{figure*}

\clearpage
\subsection{Deep Kernel Learning}
\label{sect:si:deep-kernel-learning}
Deep Kernel Learning \cite{wilson_deep_2016_dkl} augments the kernel $k_\phi$ with parameters $\phi$ of a Gaussian Process with a neural network $g_\theta$ with weights $\theta$ non-linearly transforming design points into a lower-dimensional representation. In the GP-DKL approach the neural network weights $\theta$ are fit simultaneously with the kernel parameters $\phi$. To do this simultaneous fit, at each iteration of BayesOpt we performed 400 epochs of backpropagation using the Adam optimizer. 

\begin{figure}[H]
    \centering
    \includegraphics[width=\linewidth]{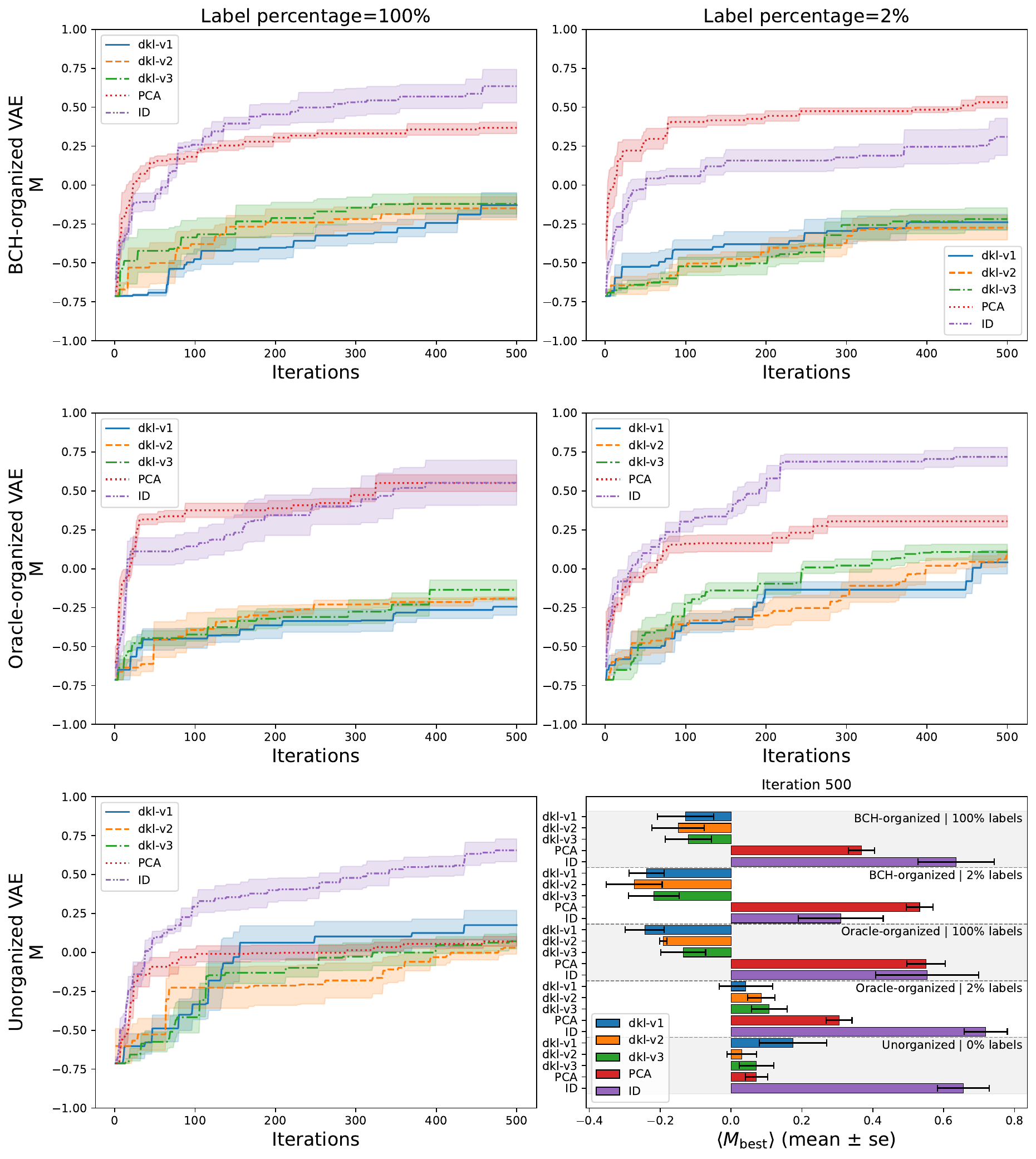}
    \caption{BayesOpt results comparing DKL to LBO and LBO in a PCA reduced latent space. }
    \label{sifig:dkl-2x3}
\end{figure}

We summarize the neural network architectures we tested in Table \ref{sitbl:dkl-architectures}. We took architecture inspiration from the original DKL publication \cite{wilson_deep_2016_dkl}, and shrunk each architecture as the amount of data they are exposed to is on the order of a couple of hundred points. In dkl-v1, we used two hidden layers with dimensions ${32}$ and $16$, outputting into $\mathbb{R}^{5}$. In dkl-v2, we used three hidden layers of dimensions $32$, $16$, and $12$, outputting into $\mathbb{R}^{5}$. In dkl-v3 we used three hidden layers of dimensions $48$, $24$, and $12$, outputting into $\mathbb{R}^{5}$. Each architecture used simple feed-forward layers and leaky ReLU activate functions. The outputs of the neural networks were then fed into a standard radial basis function kernel. Lastly, 64-dimensional latent space points are inputs to each architecture, rather than raw sequences.
\begin{table}[H]

\centering
\begin{tabular}{lllll}
\multicolumn{1}{c}{\multirow{2}{*}{\textbf{Abbreviation}}} 
& \multicolumn{3}{c}{\textbf{Architecture}}
& \textbf{Base kernel} \\
\multicolumn{1}{c}{} & \multicolumn{1}{c}{\begin{tabular}[c]{@{}c@{}}Input  \\ dimension\end{tabular}} 
                     & \multicolumn{1}{c}{\begin{tabular}[c]{@{}c@{}}Hidden \\ layers\end{tabular}} 
                     & \multicolumn{1}{c}{\begin{tabular}[c]{@{}c@{}}Output \\ dimension\end{tabular}} 
                     & \\ \hline
dkl-v1                                                     & 64                                                                             & 32-\textgreater{}16                                                          & 5 & RBFKernel                                                                               \\
dkl-v2                                                     & 64                                                                             & 32-\textgreater{}16-\textgreater{}12                                         & 5 & RBFKernel                                                                               \\
dkl-v3                                                     & 64                                                                             & 48-\textgreater{}24-\textgreater{}12                                         & 5 & RBFKernel                                                                              
\end{tabular}
\caption{Neural network architectures used for our Deep Kernel Learning experiments. We used a simple feed-forward layer for each hidden layer, leaky ReLU activation functions, and initialized weights with Xavier normal, and initialized each layer's bias as a vector of zeros. Note that we used relatively small architectures compared to the original DKL publication as the neural network is exposed to much less data.}
\label{sitbl:dkl-architectures}
\end{table}
Results for GP-DKL compared to LBO and LBO in a PCA reduce space are presented in SI Fig. \ref{sifig:dkl-2x3}. We performed GP-DKL BayesOpt over the latent space of five different transVAE models: BCH-organized transVAE with access to 100\% of the property labels, BCH-organized transVAE with access to 2\% of labels, oracle-organized with access to 100\% and 2\% of property labels, and an unorganized transVAE. Across all four organized transVAE models, DKL underperforms relative to LBO and LBO in a PCA reduced latent space. Over an unorganized transVAE, each of the approaches perform similarly, with traditional LBO pulling ahead after $500$ iterations. 

\clearpage
\subsection{Sequence space exploration}
\label{sisect:sequence-space-exploration}
Similar to Sect. \ref{sect:search-in-pca-more-exploration}, we examined the extent to which sequences are different between searching in the full latent space, and searching in a PCA projection of the latent space. Specifically, in response to the question: is there a loss of novelty when searching through the PCA projection compared to the full latent space? 

\begin{figure}[H]
    \centering
    \includegraphics[width=\linewidth]{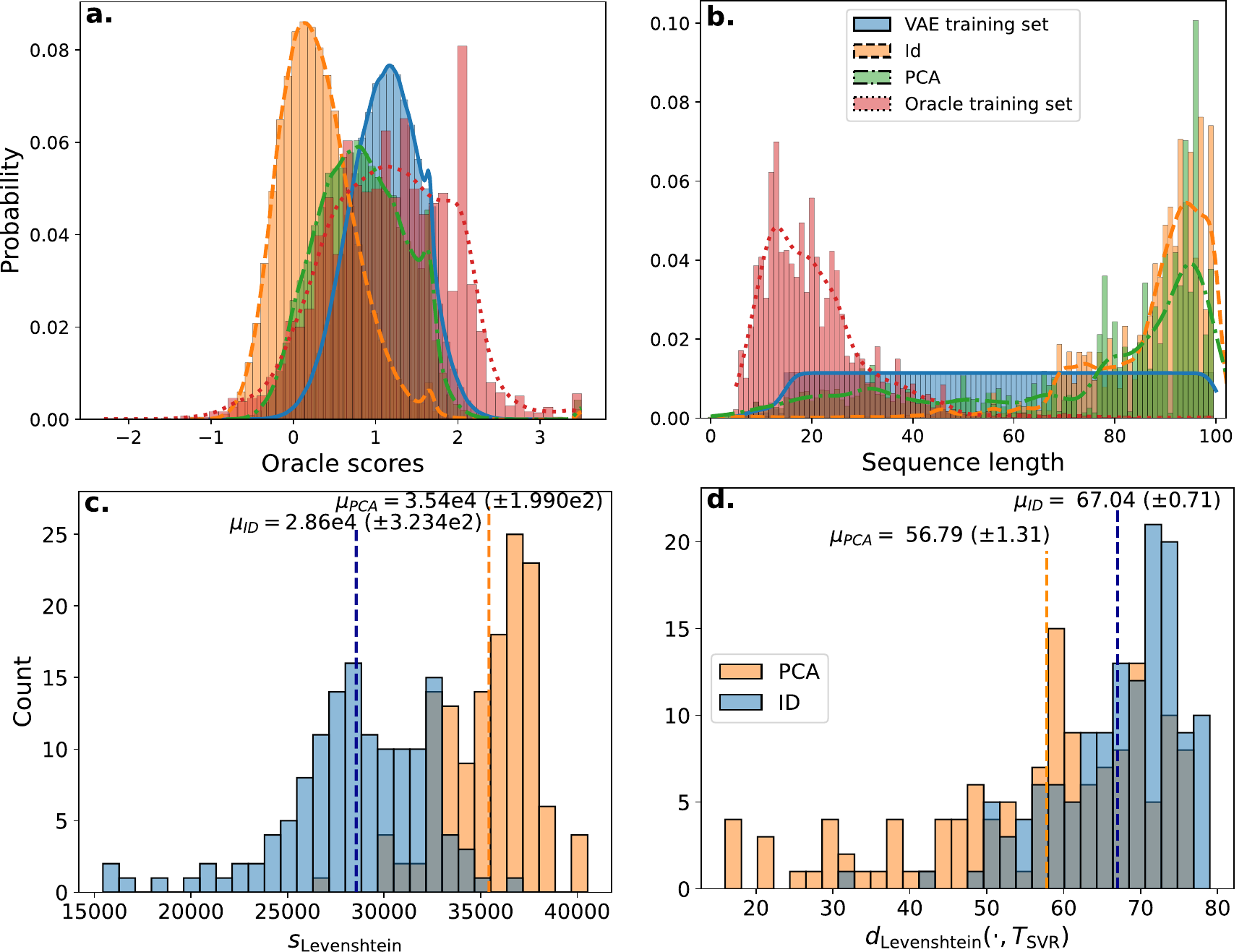}
    \caption{Exploring sequence space. (a) Distribution of oracle scores for the VAE training set (blue, solid line), the oracle's training set (red, dotted line), the sequences sampled when searching through the full latent space (orange, dashed line) or the PCA projection (green, dash-dotted line). (b) The lengths of sequences from the VAE training set (blue, solid line), oracle training set (red, dotted), or sequences sampled from the full latent space (orange, dashed) or PCA projection (green dash-dotted). (c) The Levenshtein path lengths of sequences sampled in a given BayesOpt run searching through either the full latent space (blue) or PCA projection (orange). (d) The Levenshtein distance of the best sequence identified in each BayesOpt when searching through the full latent space (blue) or PCA projection (orange) to the oracle's training set.}
    \label{sifig:sequence-space-exploration}
\end{figure}

In Fig. \ref{sifig:sequence-space-exploration}a we have compared the oracle scores of sampled sequences compared to both the VAE training set, and the oracle’s training set, finding that BayesOpt done in the full latent space on average sampled better (i.e. lower) oracle scores compared to searching in the PCA-reduced space. In Fig. \ref{sifig:sequence-space-exploration}b we plotted the lengths of the sequences sampled throughout BayesOpt runs, finding that both BayesOpt in PCA reduced spaces and in the full latent space sample similar distributions. However, we note that the PCA reduced spaces sample more short peptides, thus a distribution closer to that of both training sets. On the one hand, these observations indicate a lower novelty traversed by the algorithm in the PCA space; on the other hand, it is possible that this partially protects the algorithm in this space against venturing into territory that is less well-predicted by the oracle. In Fig. \ref{sifig:sequence-space-exploration}c we observe that the Levenshtein path lengths of each BayesOpt run were smaller in the full latent space compared to the PCA-reduced space, suggesting that sequences sampled throughout a BayesOpt run in the full latent space were more similar to each other than those sampled when searching through the PCA-reduced space. In Fig. \ref{sifig:sequence-space-exploration}d, we check the sequence distance of the best sequence found in each run to the oracle’s training set; we observe that sequences sampled from the PCA-reduced space were closer to the oracle’s training set compared to the sequences sampled from the full latent space. Figures (a) and (d) together suggest that using a PCA-reduced space can lead to sequences more similar to the oracle’s training set compared to using the full latent space. Figure (b) and (c) together suggest that searching through a PCA-reduced space can have a larger breadth of sequences sampled (larger variety of lengths, larger path lengths in sequence space) compared to searching through the full latent space. 

\clearpage
\subsection{BayesOpt trajectories in latent spaces}

\subsubsection{Coloured by Oracle values}

\begin{figure}[H]
    \centering
    \includegraphics[width=\linewidth]{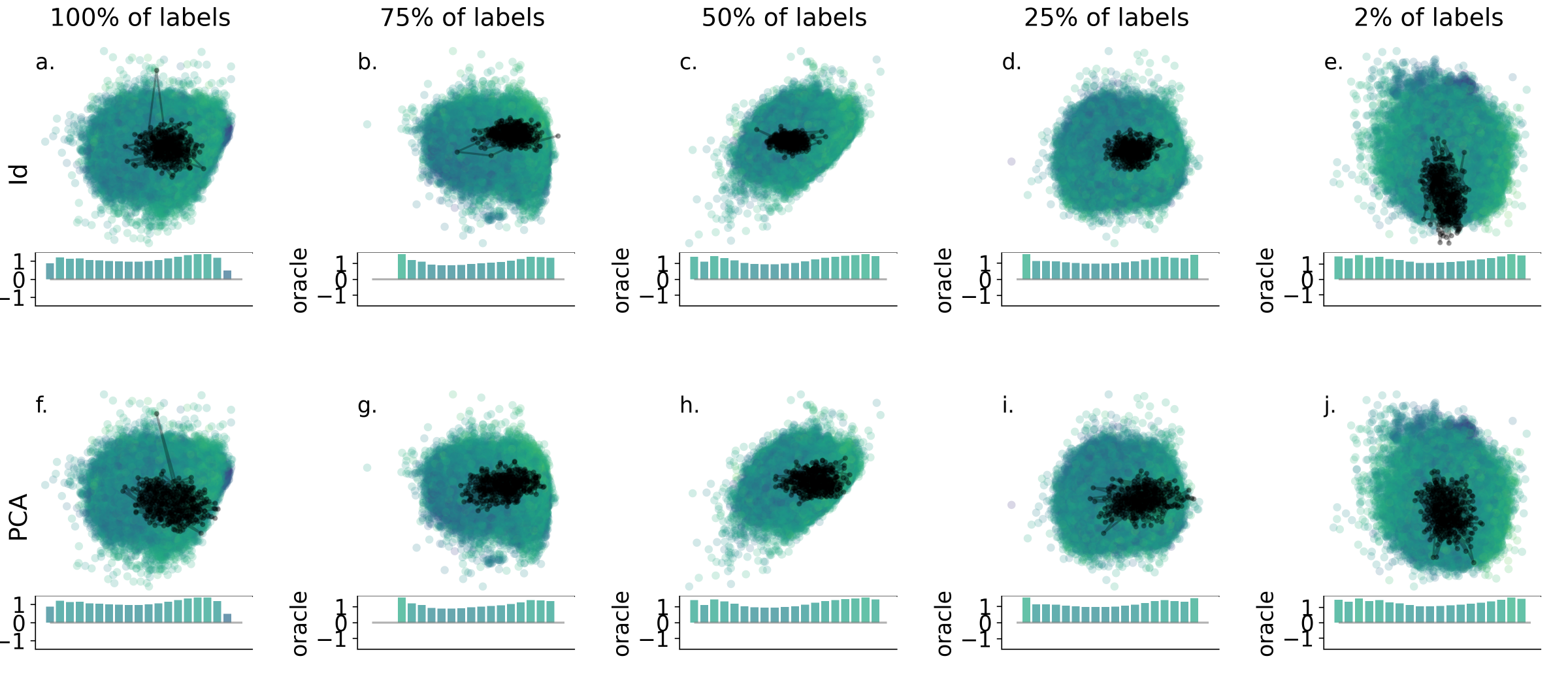}
    \caption{Boman-organized latent spaces, coloured by peptide training set oracle predictions.}
    \label{fig:si:bayesopt_run_in_boman_latent_spaces_colouredByOracle}
\end{figure}

\begin{figure}[H]
    \centering
    \includegraphics[width=\linewidth]{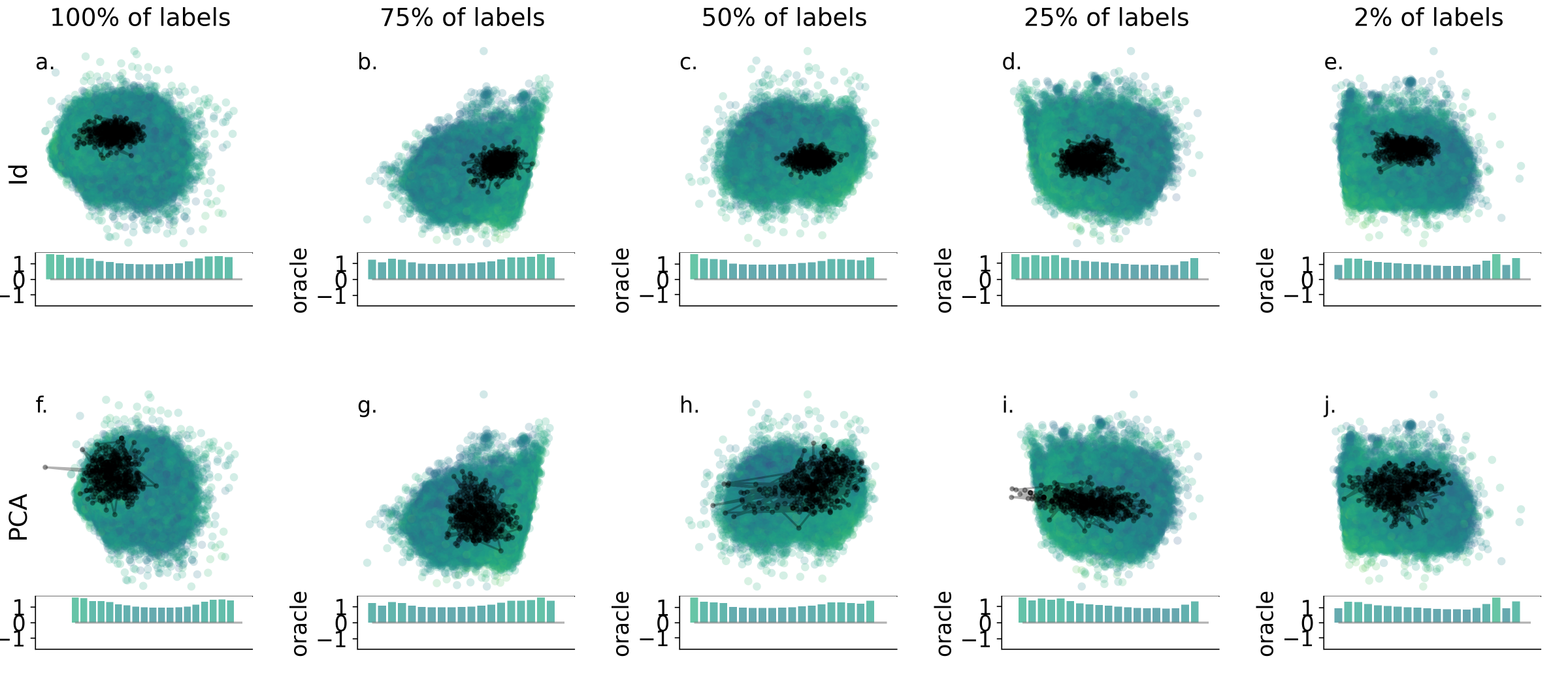}
    \caption{Charge-organized latent spaces, coloured by peptide training set oracle predictions.}
    \label{fig:si:bayesopt_run_in_charge_latent_spaces_colouredByOracle}
\end{figure}

\begin{figure}[H]
    \centering
    \includegraphics[width=\linewidth]{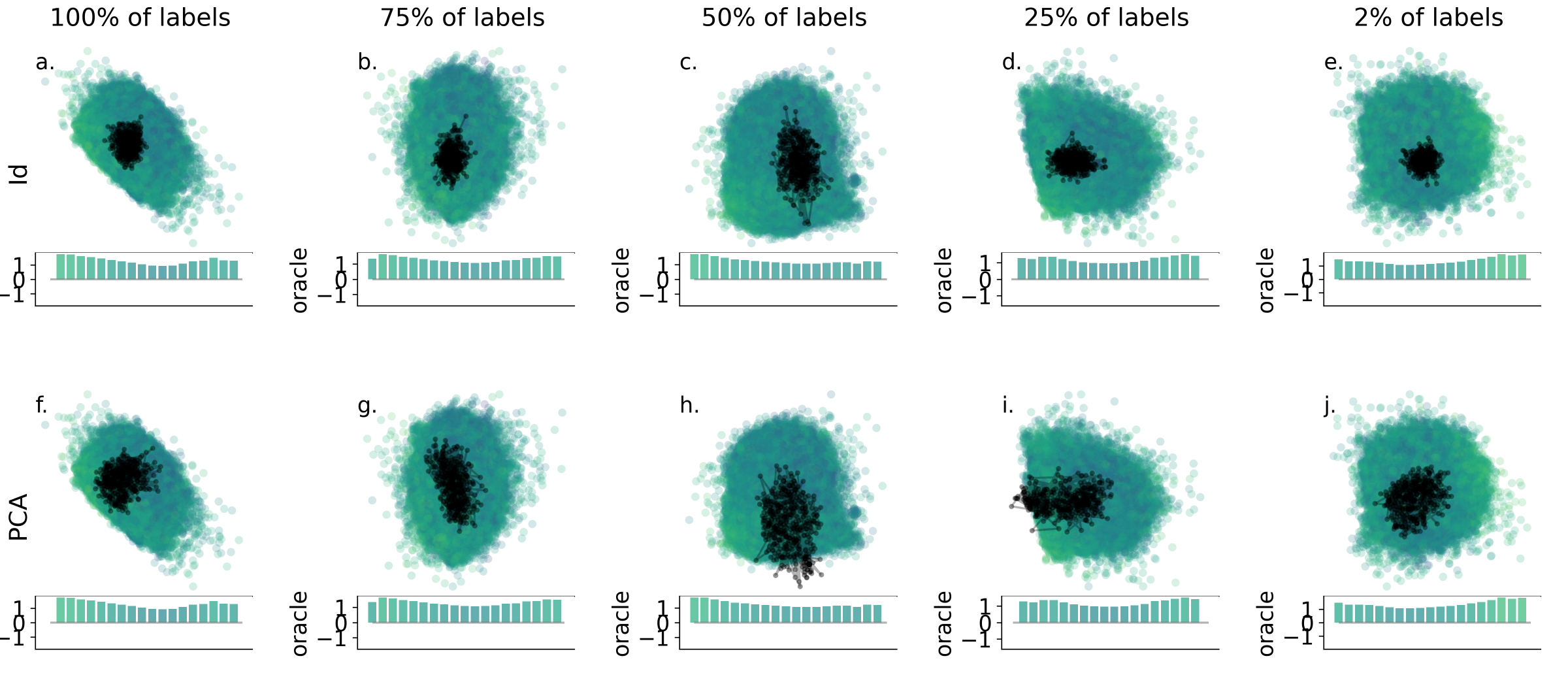}
    \caption{Hydrophobicity-organized latent spaces, coloured by peptide training set oracle predictions.}
    \label{fig:si:bayesopt_run_in_hydrophobicity_latent_spaces_colouredByOracle}
\end{figure}

\begin{figure}[H]
    \centering
    \includegraphics[width=\linewidth]{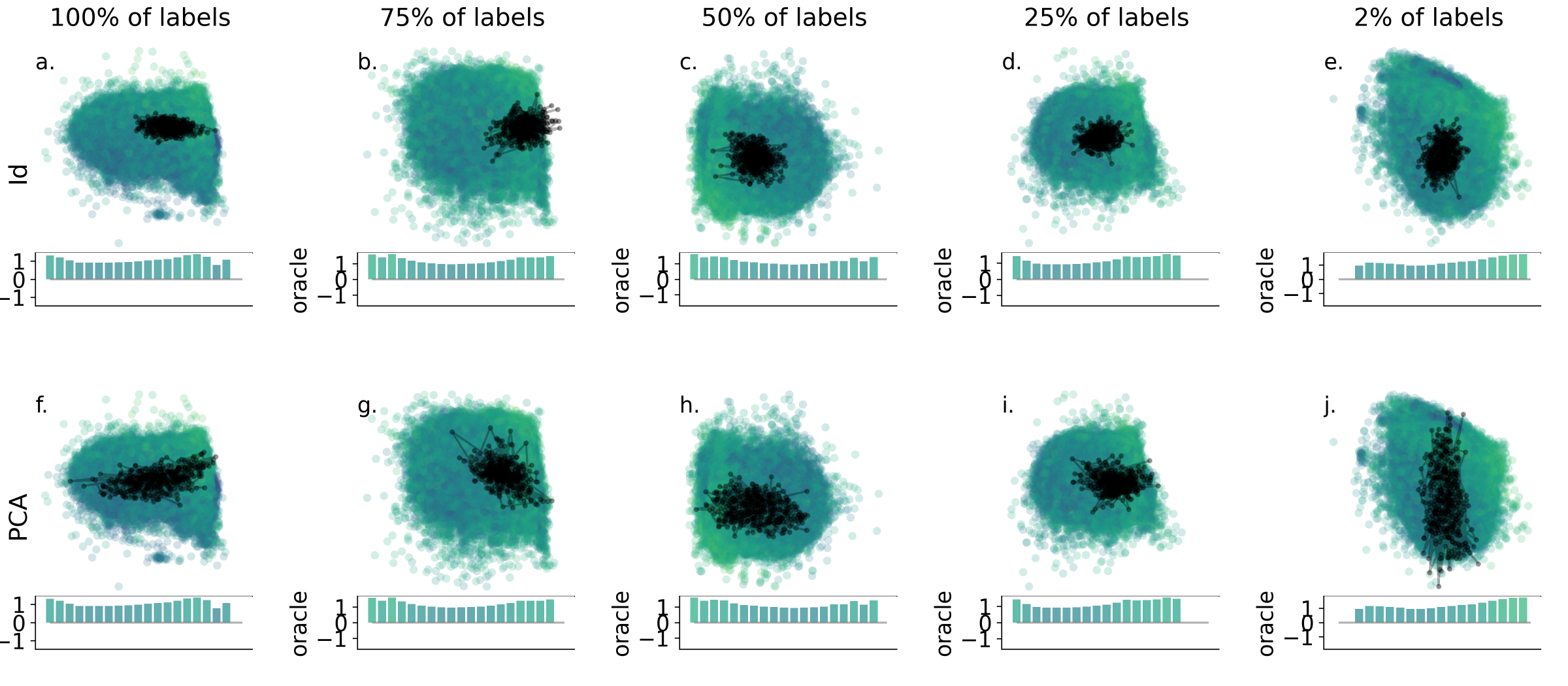}
    \caption{BCH-organized latent spaces, coloured by peptide training set oracle predictions.}
    \label{fig:si:bayesopt_run_in_bch_latent_spaces_colouredByOracle}
\end{figure}

\clearpage
\subsubsection{Coloured by organizing property}

\begin{figure}[H]
    \centering
    \includegraphics[width=\linewidth]{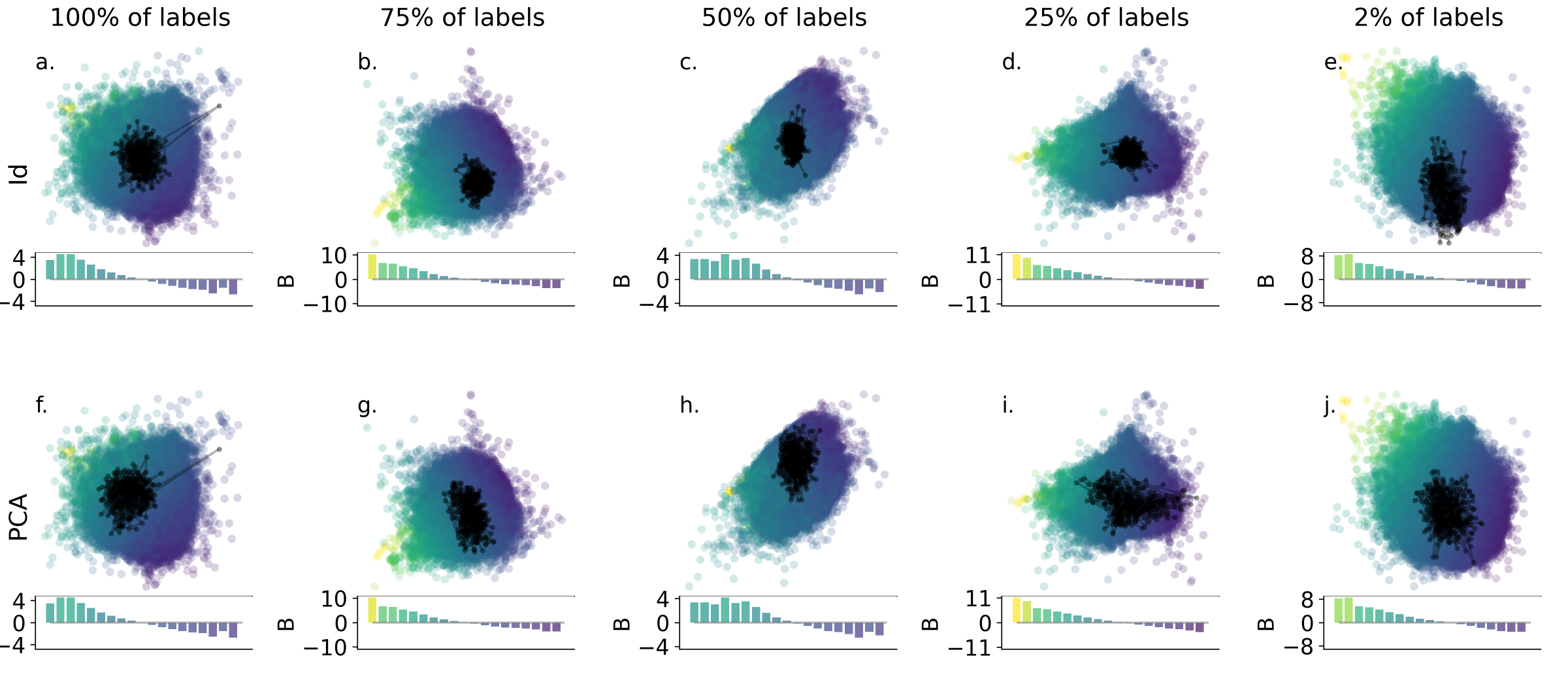}
    \caption{Boman-organized latent spaces.}
    \label{fig:si:bayesopt_run_in_boman_latent_spaces}
\end{figure}

\begin{figure}[H]
    \centering
    \includegraphics[width=\linewidth]{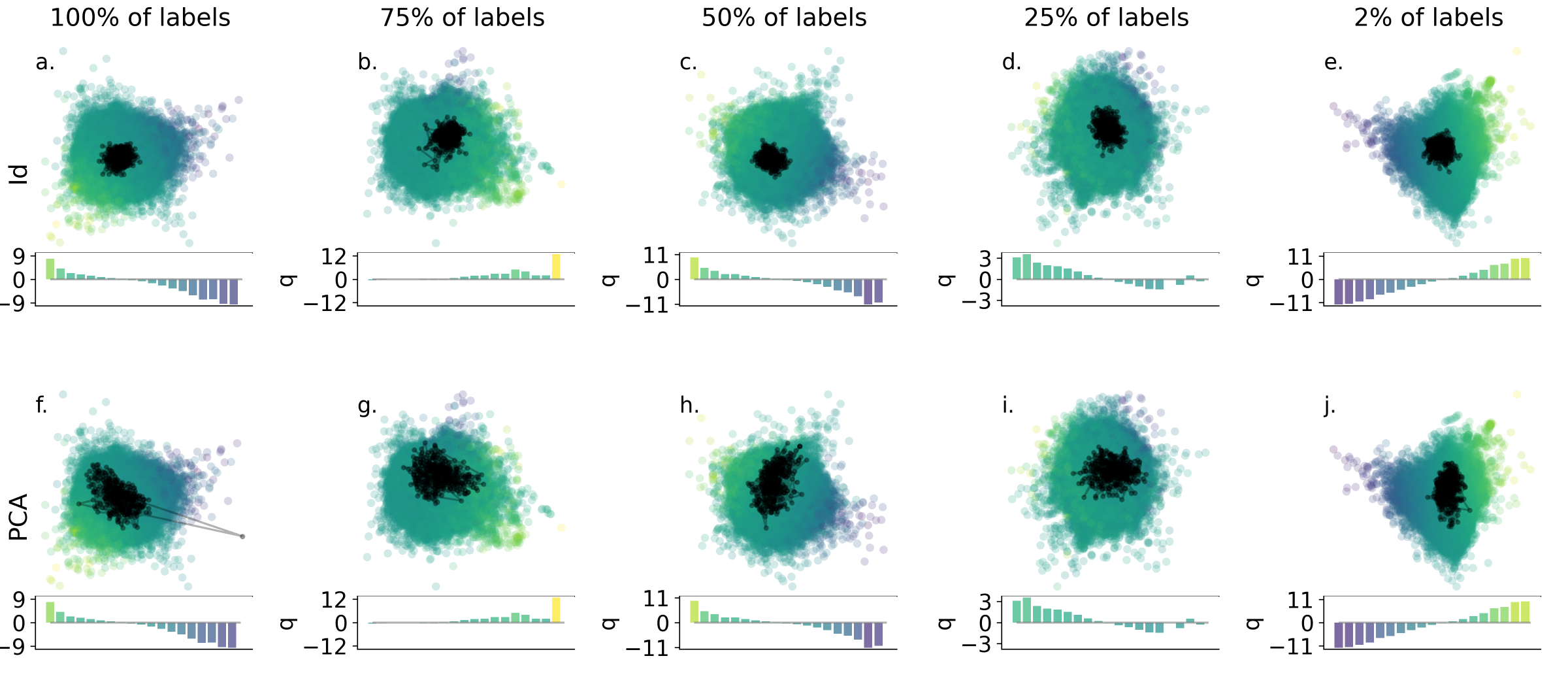}
    \caption{Charge-organized latent spaces.}
    \label{fig:si:bayesopt_run_in_charge_latent_spaces}
\end{figure}

\begin{figure}[H]
    \centering
    \includegraphics[width=\linewidth]{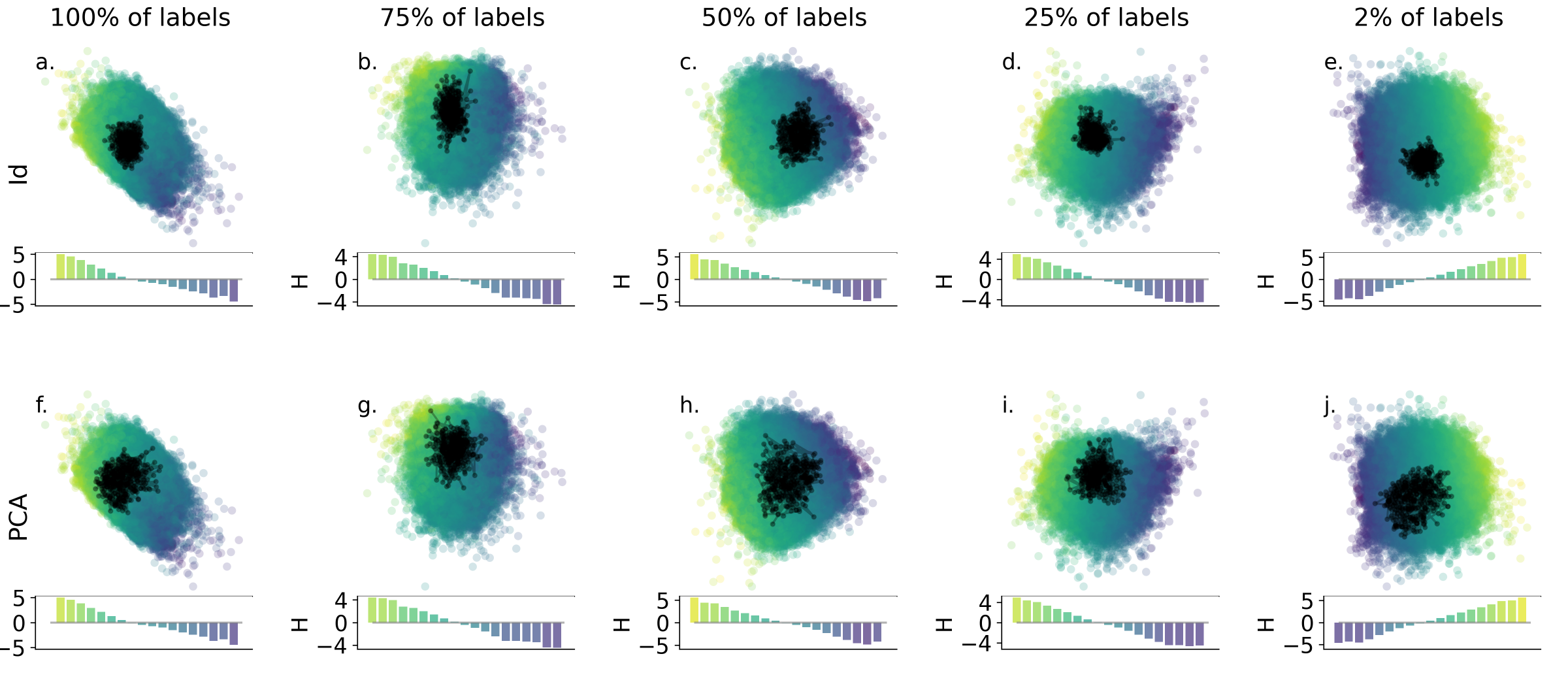}
    \caption{Hydrophobicity-organized latent spaces.}
    \label{fig:si:bayesopt_run_in_hydrophobicity_latent_spaces}
\end{figure}

\begin{figure}[H]
    \centering
    \includegraphics[width=\linewidth]{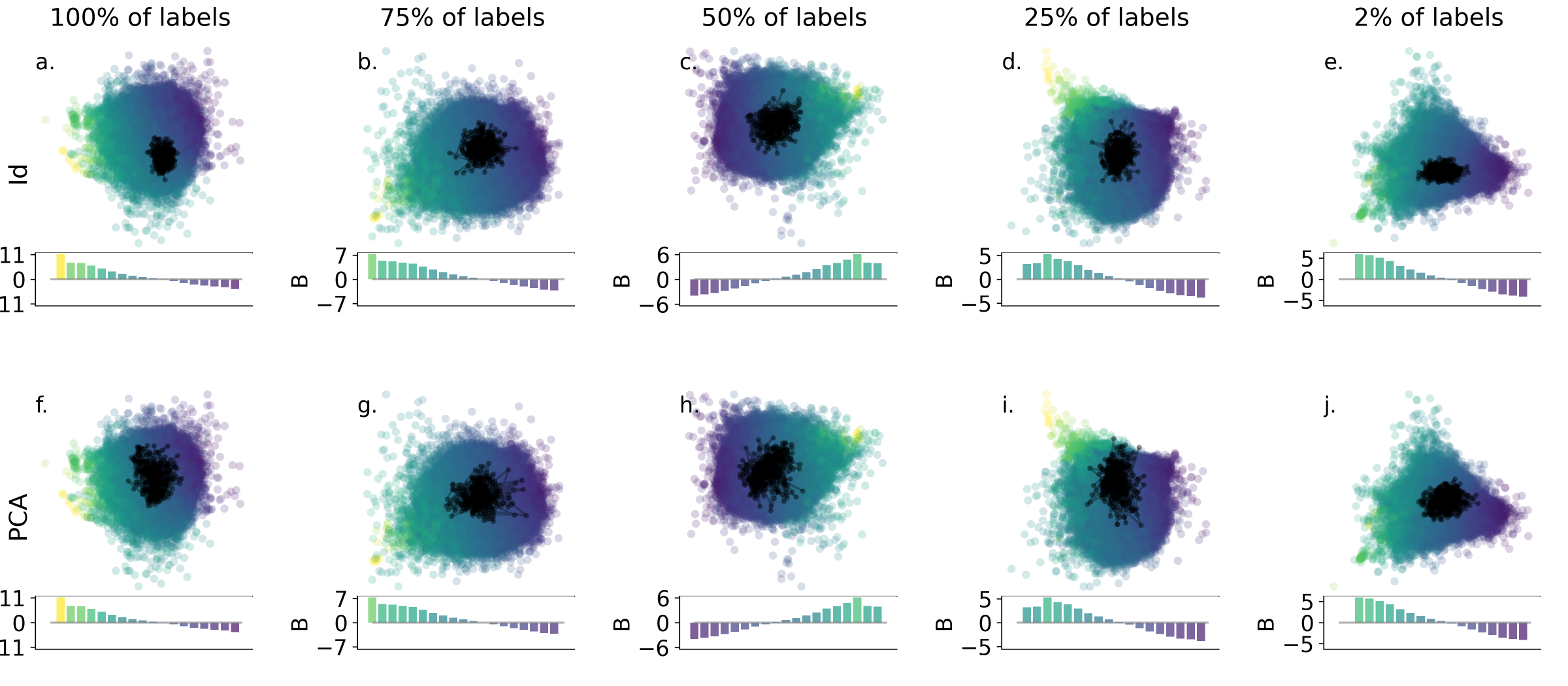}
    \caption{BCH-organized latent spaces coloured by Boman index.}
    \label{fig:si:bayesopt_run_in_bch_latent_spaces_boman}
\end{figure}

\begin{figure}[H]
    \centering
    \includegraphics[width=\linewidth]{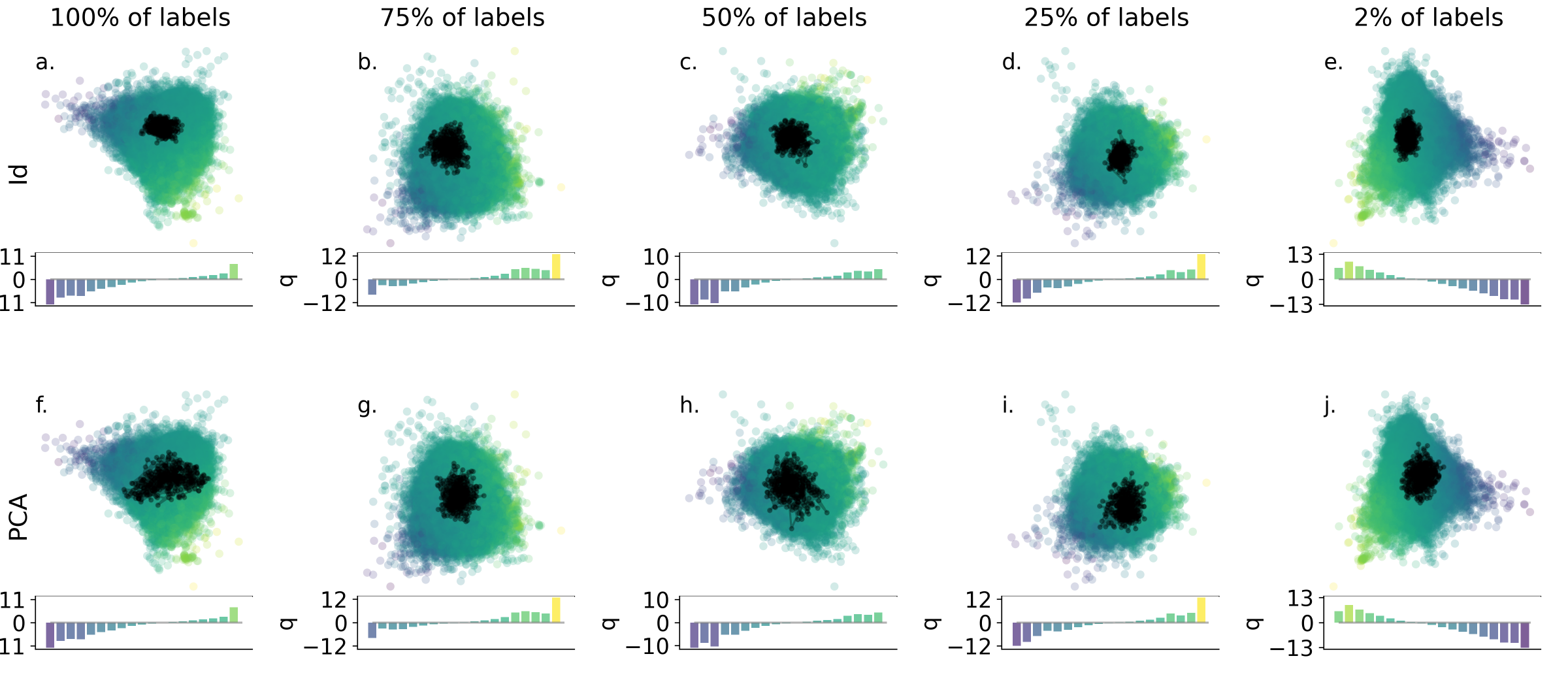}
    \caption{BCH-organized latent spaces coloured by charge.}
    \label{fig:si:bayesopt_run_in_bch_latent_spaces_charge}
\end{figure}

\begin{figure}[H]
    \centering
    \includegraphics[width=\linewidth]{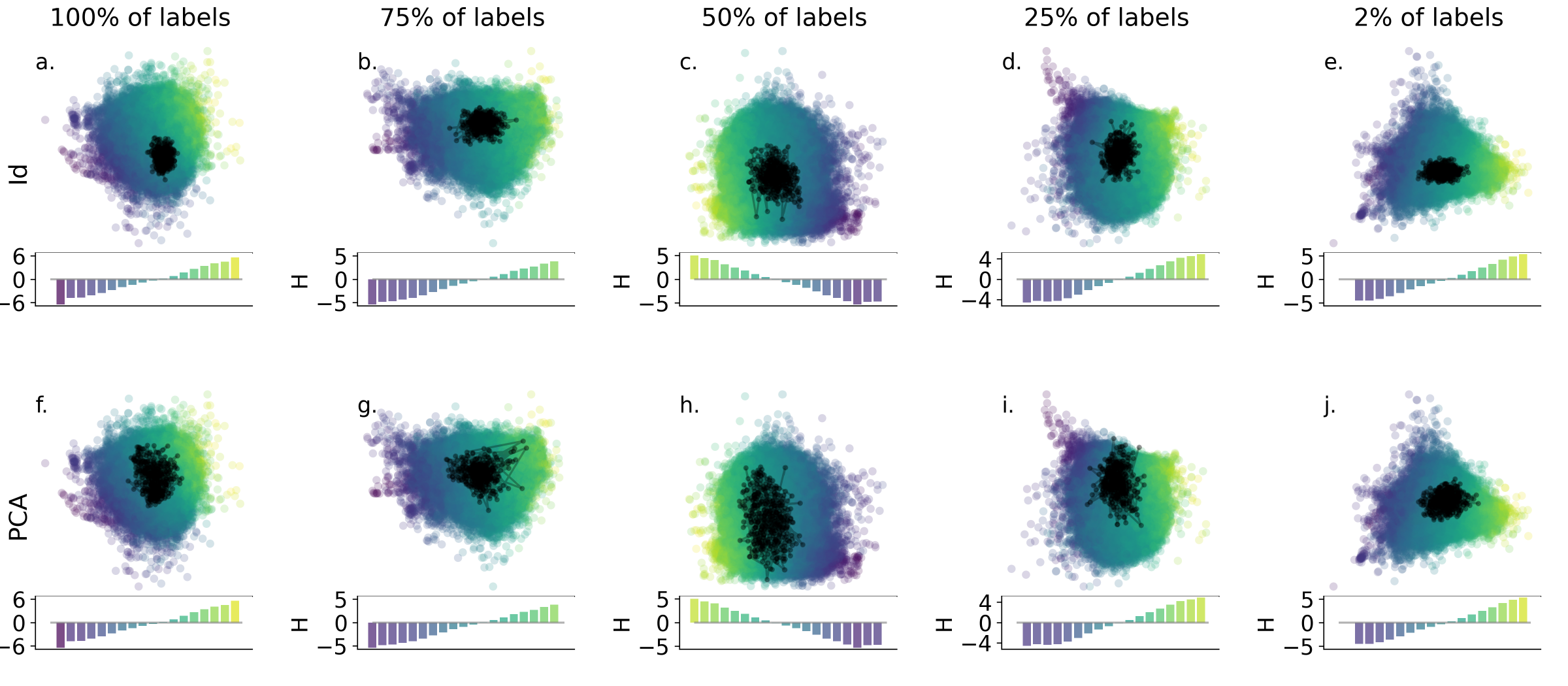}
    \caption{BCH-organized latent spaces coloured by hydrophobicity.}
    \label{fig:si:bayesopt_run_in_bch_latent_spaces_hydro}
\end{figure}


\end{document}